\documentclass[twoside,11pt]{article}
\usepackage{jmlr2e}

\usepackage{subcaption}

\usepackage{graphics,latexsym,
bm,enumerate,setspace,float,placeins,amsmath,color,amsfonts,dsfont}

\newtheorem{thm}{Theorem}
\newtheorem{lem}{Lemma}
\newtheorem{cor}{Corollary}
\newtheorem{pro}{Proposition}
\newtheorem{claim}{Claim}

\newcommand{\comment}[1]{}

\def\arg{\mathop{\rm arg}}

\ShortHeadings{An algorithmic view of $\ell_2$ regularization and some path-following algorithms}{Zhu and Liu}
\firstpageno{1}

\usepackage{lastpage}
\jmlrheading{22}{2021}{1-\pageref{LastPage}}{6/19; Revised
    12/20}{6/21}{19-477}{Yunzhang Zhu and Renxiong Liu}
\begin{document}

\title{An algorithmic view of $\ell_2$ regularization and some path-following algorithms}

\author{\name Yunzhang Zhu \email zhu.219@osu.edu \\
       \addr Department of Statistics\\
       The Ohio State University\\
       Columbus, OH 43210, USA
       \AND
       \name Renxiong Liu \email liu.6732@buckeyemail.osu.edu \\
       \addr Department of Statistics\\
       The Ohio State University\\
       Columbus, OH 43210, USA}

\editor{Rina Foygel Barber}

\maketitle

\begin{abstract}
  We establish an equivalence between the $\ell_2$-regularized
  solution path for a convex loss function, and the solution of an ordinary
  differentiable equation (ODE).
  Importantly, this equivalence reveals that
  the solution path can be viewed as the flow of
  a hybrid of gradient descent and Newton method applying
   to the empirical loss, which is similar to
  a widely used optimization technique called trust region method.
   This provides an interesting algorithmic view of
   $\ell_2$ regularization, and is in contrast
  to the conventional view that the $\ell_2$ regularization
   solution path is similar to the gradient flow of the empirical loss.
New path-following algorithms based on homotopy methods and
  numerical ODE solvers
   are proposed to numerically approximate
   the solution path. In particular, we consider respectively
  Newton method and gradient descent method as the basis algorithm
  for the homotopy method, and establish their
  approximation error rates over the solution path.
Importantly, our theory suggests novel schemes to choose grid points
that guarantee an arbitrarily small suboptimality for the solution path.
In terms of computational cost, we prove that in order to achieve an $\epsilon$-suboptimality
for the entire solution path, the number of Newton steps required for the Newton method is
$\mathcal O(\epsilon^{-1/2})$,
while the number of gradient steps required for the gradient descent method
is $\mathcal O\left(\epsilon^{-1} \ln(\epsilon^{-1})\right)$.
Finally, we use $\ell_2$-regularized logistic
regression as an illustrating example to demonstrate
the effectiveness of the proposed path-following algorithms.
\end{abstract}


\section{Introduction} 
\label{sec:intro}
It is of great interest to study statistical
procedures from a computational perspective.
Many regularization techniques
can be understood as iterative algorithmic procedures, providing an interesting
algorithmic view of regularization. For instance,
\citet{Friedman2004} studied variants of gradient descent and showed that
they closely correspond to those induced by commonly used regularization methods.
Building on the works by
\citet{Efron2004,hastie2007forward},
\citet{Freund:2017} showed that
the classic boosting algorithm in linear regression
can be viewed as the iterates generated by applying
 subgradient descent algorithm to
the loss function defined as the
maximum absolute correlation between the
features and residuals.

Tikhonov (or $\ell_2$) regularization \citep{Tikhonov1977}
is ubiquitously used in many modeling procedures, and
in the statistical literature it traces back to 
\citet{hoerl1970ridge},
 where it is often referred to as ridge regression.
It is natural to seek an algorithmic view of $\ell_2$ regularization, that is,
what algorithm would
produce a sequence of iterates that is identical to
the $\ell_2$-regularized solutions.
Surprisingly, this has not been formally
established for a general convex loss function, with most related works
 focusing on least squares error loss.
For example, \citet{fleming1990equivalence} showed
an equivalence between $\ell_2$ regularization and
the iterates of certain optimization algorithms.
More recently,
\citet{Suggala2018} formally studied the connection between
$\ell_2$ regularization path and the iterates of gradient descent
algorithm. They established a pointwise bound between these
two paths and use this to establish the risk bound of the
iterates of gradient descent algorithm.
\citet{Neu2018IterateAA}
proposed a weighting scheme for
gradient descent iterates so that it is exactly equal to
some $\ell_2$-regularized solution.
Another related work is by \citet{ali2019continuous}, which
compares the risk of gradient flow to that of $\ell_2$-regularized
solutions in the context of least squares regression.


Another line of work has focused on the similarity between
 algorithmic approaches
 and explicit regularization approaches in terms of their
 statistical performance.
Earlier work includes \citep{Frank1993}, who pointed out
a similarity between ridge regression and partial least squares regression, where
the latter has been shown to be equivalent to
conjugate gradient descent with squared-error loss \citep{Wold:1984}.
More recently,
\citet{Yao:2007aa} considered nonparametric regression
in a reproducing kernel Hilbert space (RKHS)
and provided some theoretical justification for early stopping of
gradient descent algorithm.
\citet{raskutti2014early}
proposed a data-dependent and easily computable stopping rule for gradient descent,
and showed that it can achieve similar risk bounds as that of the ridge regression.

In this article, we
 establish an algorithmic view of ridge regression
for a general convex loss function.
We first establish an equivalence between $\ell_2$-regularized
solution path for a convex loss function, and the solution of an ODE.
This reveals a formal
equivalence between $\ell_2$ regularization solution path
 and the iterates produced by
 a hybrid of gradient descent and Newton algorithm when the step
 size tends to $0$.
This equivalence has been previously discovered by
\citet{Suggala2018} (see proof of Theorem 1 in \cite{Suggala2018}).
However, a rigorous proof was not provided by \citet{Suggala2018}.

More formally, denote by
$L_n(\theta)$ some convex empirical loss function,
 where $\theta \in \mathbb R^p$ is the
parameter. Let $C(\cdot)$ be a differentiable increasing
 function with $C(0) = 0$ and $\lim_{t \rightarrow \infty}
C(t) = \infty$.
We consider the solution path of
an $\ell_2$-regularized convex optimization problem:
\begin{equation}
  \label{eq:path}
  \theta(t) = \arg \min_{\theta \in \mathbb R^p} \left( C(t) \cdot L_n(\theta) +
  \frac{1}{2}\|\theta\|_2^2 \right) \, .
\end{equation}
Note that as $t$ varies from $0$ to $\infty$, the solution $\theta(t)$
 varies from $\bm 0$ to a minimizer of $L_n(\theta)$.
Throughout this article, we focus on $\ell_2$ regularization, although
some of the results in this article can be easily extended to
the case where the regularizer is a general quadratic function.

Our first main result is that under some smoothness condition on $L_n(\theta)$,
the solution path defined by \eqref{eq:path}
coincides with the \textit{global solution} to the following
 \textit{ordinary differential equation (ODE)},
\begin{equation}
  \label{eq:ode}
  \theta^\prime(t)
  = - C^\prime(t) \left[ C(t) \cdot \nabla^2 L_n(\theta(t)) +
  I \right]^{-1}
  \nabla L_n(\theta(t)), t \geq 0 \, ,
\end{equation}
with an initial condition $\theta(0) = \bm 0$.
More specifically, under the assumption that $L_n(\theta)$
is convex and has continuous Hessian,
we show that $\theta(t)$ is differentiable
in $t$ and the solution to the above ODE is also a
solution path to the original optimization
problem \eqref{eq:path}.

To better interpret the ODE formulation in \eqref{eq:ode},
we consider a special
choice of $C(t) = \exp(t) - 1$ throughout this article.
In fact, based on the aforementioned equivalence,
it is easy to see that
the choice of $C(t)$ is not essential,
because the
 solution to the ODE can always be viewed as the solution path to
  \eqref{eq:path} regardless of the choice of $C(t)$.
In another words, different choices of $C(t)$ produce
 the same path $\theta(t)$ in $\mathbb R^p$---they just ``travel''
 at different speeds to the
 minimum $\ell_2$ norm minimizer of $L_n(\theta)$
 as $t$ varies from $0$ to $\infty$.
 Another rationale for choosing $C(t) = \exp(t) - 1$ is that this agrees
 with the common practice of picking grid points on a log scale
for tuning parameter selection.

Plugging $C(t) = \exp(t) - 1$ into \eqref{eq:ode},
the ODE becomes
\begin{equation}
  \label{eq:ode_special}
  \theta^\prime(t)
  = - \left[ (1 - e^{-t}) \cdot \nabla^2 L_n(\theta(t)) +
  e^{-t} \cdot I \right]^{-1}
  \nabla L_n(\theta(t)), t \geq 0 \text{ with } \theta(0) = \bm 0 \, .
\end{equation}
The left hand side $\theta^\prime(t)$ can be viewed as the local direction of
the solution path at time $t$.
Interestingly, we can see from \eqref{eq:ode_special} that
the search direction can be thought of as
certain hybrid of gradient descent search direction
$- \nabla L_n(\theta(t))$ and Newton direction
$- \left[\nabla^2 L_n(\theta(t)) \right]^{-1}
\nabla L_n(\theta(t))$. Moreover, the search direction is
closer to gradient search direction when $t$ is small,
and closer to Newton direction when $t$ is large. This provides
an interesting algorithmic perspective of $\ell_2$ regularization,
and
partially confirms previous belief that
the $\ell_2$ regularization path is closely related to the solution path
generated by the gradient descent method.
In particular,
the ODE update direction \eqref{eq:ode_special} resembles to that of
the trust region algorithm or its precursor the
Levenberg–Marquardt algorithm
\citep{Levenberg1944, conn2000trust}.
Both algorithms produce similar types of
 hybrid of gradient descent and Newton direction:
\begin{equation}
  \label{eq:trust-region}
  \theta_{k+1} = \theta_k -\left[ \nabla^2 L_n(\theta_k) +
  \lambda I \right]^{-1}
  \nabla L_n(\theta_k) \, ,
\end{equation}
where $\lambda > 0$ is often adaptively chosen or determined by the size
of the trust region.
Although these optimization algorithms
have very similar
update directions, they are designed
with the goal of finding
a minimizer of the unregularized loss $L_n(\theta)$ reliably and efficiently.
By contrast, our focus here is to provide an algorithmic
 interpretation of the $\ell_2$-regularized solution path,
 and to design numerical procedures to
 approximate the entire solution path for the regularized problem.
A more detailed discussion of this connection
 is provided in Section \ref{sec:newton_approx}.


Aside from providing
a conceptual connection between the $\ell_2$-regularized solution path
and solutions to an ODE,
the ODE formulation also opens up avenues for designing algorithms
to approximate the entire $\ell_2$-regularized solution path or the
minimum norm minimizer of $L_n(\theta)$.
In particular, the ODE formulation
 \eqref{eq:ode} is known as the \textit{initial-value problems}
 in the numerical ODE literature
 \citep[see, e.g.,][]{butcher2016numerical}. Many effective
numerical ODE solvers such as the
Euler's method and Runge-Kutta method
\citep[see Chapter 2 and 3 of ][]{butcher2016numerical}
can be used to approximately solve
the ODE over a discrete set of grid points.

In addition to ODE solvers, we also propose
 two new path-following (homotopy) methods based on
Newton method and gradient descent method as
their ``working horse'' algorithms to approximate
the solution path $\theta(t)$ over a given
region $[0, t_{\max})$, where $0 < t_{\max} \leq \infty$.
An approximate solution path $\tilde \theta(t)$
is constructed through linearly interpolating
the approximate solutions at the selected grid points
(see Section \ref{subsec:linear_interpolation} for its formal definition).
Theoretically, we bound
the global approximation error of the entire solution path in terms of
$\sup_{0 \leq t \leq t_{\max}} \left\{ f_t(\tilde \theta(t)) - f_t(\theta(t)) \right\}$
for both the Newton method and gradient descent
method (c.f. Theorem \ref{thm:global_bd_newton} and \ref{thm:uniform_grad_bd}),
where $f_t(\theta) = (1-e^{-t})L_n(\theta)
+ ( e^{-t} / 2 ) \cdot \|\theta\|_2^2$ is a scaled
version of the regularized objective function.
These bounds reveal an important interplay between
the choice of grid points and accuracy of the solutions
at the selected grid points.
In particular, they
allow us to design novel schemes to
select grid points $t_1, \ldots, t_N$ so that
the overall computations required to achieve a prespecified
suboptimality is minimized.

Using the newly proposed grid point selection schemes,
we further derive upper bounds on the total number of steps required
to achieve an $\epsilon$-suboptimality, i.e.,
\begin{equation}
  \sup_{0 \leq t \leq t_{\max}} \left\{ f_t(\tilde \theta(t)) - f_t(\theta(t)) \right\}
  \lesssim \epsilon \, ,
\end{equation}
where $\epsilon > 0$. In particular, the number of Newton steps and gradient steps
required to achieve $\epsilon$ suboptimality are at most $\mathcal O(\epsilon^{-1/2})$
and $\mathcal O\left(\epsilon^{-1} \ln(\epsilon^{-1})\right)$, respectively.
To the best of our knowledge, these complexity results are new, and
parallel to existing complexity results for the Newton method and gradient descent method
when applied to solving a single optimization problem
(i.e., the problem corresponding to $t = t_{\max}$).
Moreover, the new complexity results
also suggest that Newton method,
being more expensive at each iteration, requires less number of iterations
as compared with the gradient descent method.
Numerical experiments on a $\ell_2$-regularized logistic regression
corroborate with the theoretical results in that the Newton method
tends to perform better than the gradient descent method for small to medium scale problems,
while the gradient descent method is more efficient for large-scale problems.


In optimization, homotopy techniques have been used
in many algorithms including the interior point algorithm
\citep{nesterov1993}. For example, the solution to a
constrained convex optimization problem can be viewed as the limit
of the solutions to a family of
unconstrained optimization by introducing a barrier (or penalty) function.
However, the focus of these methods is the recovery of the limit of the path,
rather than the entire solution path.
That said, the idea of the warm-start strategy has been well developed, which
consists of the so-called ``working horse'' algorithm and the policy
for updating the penalty parameter
 \citep[see, e.g., Chapter 1.3 of ][]{nesterov1993}.
 Typically, Newton method is used as the ``working horse'' for
the modern path-following interior point methods.
 In statistical learning literature,
\citet{Osborne1992} and \citet{homo_osborne2000} applied the homotopy
technique to generate piecewise linear trajectories in quantile
regression and LASSO, respectively. Later \citet{Efron2004},
\citet{Hastie2004}, and \citet{Rosset2007} exploited the homotopy
path-following methods to generate an entire solution path for a
family of regularization problems.
Subsequent developments include
\citet{friedman2007, hoefling2010, Arnold2016},
among others. These works often leverage the piecewise
linearity of the solution path so that an exact path-following
algorithm can be explicitly derived.
For situations where the solution paths are not
piecewise linear, approaches based on ODE solvers were considered in
\citet{wu2011, zhou2014} and a path-following algorithm based on Newton method
was considered in \citet{rosset2004tracking}.
In particular, \citet{rosset2004tracking} also proposed to use
one-step Newton update to generate the solution path, and is the most relevant to our work.
However, it used a constant step size scheme and
only established the pointwise closeness to the solution path.
To the best of our knowledge,
our work is the first to theoretically analyze the global approximation error
of the entire solution path.

To summarize, our key contributions are that
\begin{itemize}
  \item we provide an algorithmic view
   of $\ell_2$ regularization
  through establishing a formal equivalence to
  the solution of an ODE,
  which further reveals an interesting connection to the
  trust region algorithm and
  Levenberg-Marquardt algorithm;
  \item we propose
  two path-following algorithms based
  on Newton update and gradient descent update, and
  establish global approximation-error bounds for
  the solution paths generated
  by both algorithms;
  \item we also consider various numerical ODE solvers
   to approximate the $\ell_2$-regularized solution path.
\end{itemize}

The rest of the paper is organized as
follows. Section~\ref{sec:properties} discusses the properties
of the solution path, and provides a proof of the equivalence to
the ODE solution.
Section~\ref{sec:newton_approx} introduces the linear interpolation scheme and
discusses various approaches to approximate the regularized solution
path. In Section~\ref{sec:newton_approx_theory},
global approximation-error bounds for two path-following algorithms
are established. New grid point selection schemes and the associated computational
complexities are derived.
In Section \ref{sec:sims}, we compare the
proposed methods with some competing methods through
a simulated study using $\ell_2$-regularized logistic regression.
We close with some remarks in Section \ref{sec:discussion}.

\section{Properties of the solution path}
\label{sec:properties}
In this section, we first start with an informal derivation of the
ODE \eqref{eq:ode}
using the optimality condition of the $\ell_2$-regularized solutions.
We then rigorously establish the
differentiability of $\theta(t)$, which turns out to be the key ingredient
in establishing the equivalence between
\eqref{eq:path} and \eqref{eq:ode}.
Note that the optimality condition of \eqref{eq:path}
at time $t$ is
\begin{equation}
 C(t) \nabla L_n(\theta(t)) + \theta(t)  = 0 \, .
\end{equation}
If we assume for now
that $\theta(t)$ is differentiable in $t$, by taking derivative
with respect to $t$, we obtain that
\begin{equation*}
  C^\prime(t) \nabla L_n(\theta(t)) +
  C(t) \nabla^2 L_n(\theta(t)) \theta^\prime(t)
  + \theta^\prime(t) = 0 \, ,
\end{equation*}
which implies that
\begin{equation*}
  \theta^\prime(t)
   = - C^\prime(t)   \left( C(t) \nabla^2 L_n(\theta(t))
  + I \right)^{-1}   \nabla L_n(\theta(t)) \, .
\end{equation*}
It is easy to see that $\theta(0) = \bm 0$. Thus, it follows that
the $\ell_2$-regularized solution path must be a solution to the ODE \eqref{eq:ode}.

Next, we make the above argument rigorous. The missing piece
of the above argument is
the differentiability of the solution
 path $\theta(t)$ in $t$.
To formally establish this,
we impose convexity and smoothness conditions on $L_n(\theta)$,
and show that
the solution path $\theta(t)$ is differentiable in $t$.
The assumptions needed
on the loss function $L_n(\theta)$ are described below in
Assumption (A0).

\vskip 0.1in
\noindent
\textbf{Assumption (A0). } Suppose that $L_n(\theta)$ is
convex and has continuous second derivative,
with $\bm 0 \in \textbf{dom}\, L_n$, where
$\textbf{dom}\, L_n$ denotes the domain of $L_n(\theta)$.

\begin{thm}
  \label{thm:equiv_to_ode}
  Assume that $C(t)$ is a strictly increasing and
   differentiable function, with $C(0) = 0$ and
    $\lim_{t \rightarrow \infty} C(t) = \infty$.
Under Assumption (A0),
  the solution path $\theta(t)$ defined by \eqref{eq:path}
  is differentiable in $t$, and it
  is the unique solution to the ODE \eqref{eq:ode}.
\end{thm}
Some remarks are in order.
The above result has been informally presented in
\cite{Suggala2018} without rigorously proving the differentiability of
the regularization path $\theta(t)$.
The connection to the ODE
suggests that the choice of $C(t)$ is not essential.
If we choose $C(t) = \exp(t) - 1$, then it follows from Theorem
\ref{thm:equiv_to_ode} that the search direction
at time $t$ is
\begin{equation*}
  - \left( (1 -e^{-t}) \nabla^2 L_n(\theta(t))
  + e^{-t} I \right)^{-1} \nabla L_n(\theta(t)) \, ,
\end{equation*}
which is similar to the search direction of
Levenberg-Marquardt algorithm \citep{Levenberg1944}.
 Interestingly, the direction is
a hybrid of Newton and gradient direction, and it is
close to gradient direction when $t$ is small and close
to the Newton direction when $t$ is large.

It is also worth pointing out that
\citet{Efron2004} shows that
$\ell_1$-regularized solution path is piecewise linear.
By contrast, here we show that
the $\ell_2$-regularized solution path is more smooth in the sense that
it is differentiable everywhere.
Moreover, without the smoothness
assumption, the differentiability of the solution path can not be
established in general. Examples include $\ell_2$-regularized quantile regression
and support vector machine, both of
 which have nonsmooth loss functions and their
solution paths were shown to be nondifferentiable in $t$ by
\citet{Osborne1992} and \citet{Hastie2004}, respectively.
In this sense, the smoothness assumption for the loss function is necessary.

Next we present some properties of the $\ell_2$-regularized solution path, which may be interesting
on their own.  In particular, it shows that the $\ell_2$ norm of the
 solutions $\|\theta(t)\|_2$ is nondecreasing, while
 $\|\theta(t)\|_2 / C(t)$ is nonincreasing.
 Moreover, it is shown that the solution $\theta(t)$
 converges to the minimum
$\ell_2$ norm minimizer of $L_n(\theta)$ as $t$ goes to infinity if it is finite.
\begin{cor}
  \label{cor:solution_prop}
  Assume that $C(t)$ is a strictly increasing and
  differentiable function, with $C(0) = 0$ and
$\lim_{t \rightarrow \infty} C(t) = \infty$.
Then
\begin{itemize}
  \item[(i)] $\|\theta(t)\|_2$ is nondecreasing in $t$ and
   $L_n(\theta(t))$ is nonincreasing in $t$;
\item[(ii)] $ \|\theta(t)\|_2 / C(t)$ is nonincreasing in $t$;
\item[(iii)] if $L_n(\theta)$ is a continuous, closed proper convex function and
the minimum $\ell_2$ norm minimizer of $L_n(\theta)$,
 denoted as $\theta^\star$, is finite,
then $\lim_{t \rightarrow \infty} \theta(t) = \theta^\star$.
\end{itemize}
\end{cor}
We remark that the convergence of $\theta(t)$ to the
minimum $\ell_2$ norm minimizer has already been established in
Theorem 8 of \cite{Suggala2018}. Moreover, the monotonicity property
of the solution path and the loss function is also probably well-known as
folklore. We include them here to make the paper largely self-contained.
Also, as pointed in \cite{Suggala2018},
results of similar flavor have also been obtained recently
for various types of optimization algorithms
\citep[see, e.g., ][]{soudry2017implicit,Gunasekar2017,gunasekar18a}.
Moreover, it is noted that the smoothness assumption on the loss function
is not necessary for establishing monotonicity or convergence
to the minimum $\ell_2$ norm
solution. As such, this result is applicable to
nonsmooth loss functions such as
support vector machine and quantile regression.


Both Theorem \ref{thm:equiv_to_ode}
and Corollary \ref{cor:solution_prop} can be extended to
handle general quadratic regularizers.
More specifically, it can be shown that
Theorem \ref{thm:equiv_to_ode}
and Corollary \ref{cor:solution_prop} continue to hold if
a general quadratic regularization function
 $\frac{1}{2} (\theta - \theta_0)^\top Q (\theta - \theta_0)$
 is used, where $Q$ is a positive definite matrix
  and $\theta_0$ is some starting point.
The corresponding ODE becomes
\begin{equation*}
  \theta^\prime(t)
  = - C^\prime(t) \left[ C(t) \cdot \nabla^2 L_n(\theta(t)) +
  Q)\right]^{-1}
  \nabla L_n(\theta(t)), t \geq 0 \,
  \text{ with } \theta(0) = \theta_0 \, .
\end{equation*}
For Corollary \ref{cor:solution_prop},
the limit of $\theta(t)$ would be the minimizer
of $L_n(\theta)$ that is closest to $\theta_0$ with
distance induced by $\|\cdot\|_Q$-norm.

\section{Approximation of the solution path}
\label{sec:newton_approx}
When the solution path $\theta(t)$ is not piecewise linear, typically
only an approximate solution path
can be obtained.
There are in general two types of approaches to
obtain an approximate solution path.
One is based on the idea of homotopy method
\citep{Osborne1992,nesterov1993,rosset2004tracking},
and the other one is based on numerical ODE methods
\citep[see, e.g., ][]{wu2011, zhou2014}.
In this section, we study
these two types of approximation schemes.
Specifically,
for homotopy methods,
we use Newton update and gradient descent update
as the basis, and derive the corresponding path-following algorithms.
We also consider numerical ODE solvers
 based on the explicit forward Euler method and
the (second-order) Runge-Kutta method
\citep{butcher2016numerical}.

Note that although the focus of
typical homotopy algorithms is to find a single solution at the limit,
here we use the idea of homotopy algorithm
with the goal of approximating the entire solution path
\citep[see, e.g.,][]{friedman2007} through linear interpolation.
More specifically,
given the approximate solutions $\{\theta_k\}_{k=1}^N$
at a set of prespecified grid points $0 < t_1 < \cdots < t_N < \infty$,
we propose an approximate solution path through linearly interpolating
these solutions. This produces a continuous approximate solution path for
$\theta(t)$. Throughout this section,
we assume that $C(t) = \exp(t) - 1$ and
consider \eqref{eq:ode_special} instead of \eqref{eq:ode},
because they generate the same solution path.


\subsection{Approximate solution path through linear interpolation}
\label{subsec:linear_interpolation}
Suppose that the goal is to approximate the solution path
$\theta(t)$ over a given interval $[0, t_{\max})$ for some $t_{\max} \in (0, \infty]$,
where we allow $t_{\max} = \infty$.
Given a set of grid points $0 < t_1 < \cdots < t_N < \infty$, and
the approximate solutions $\{\theta_k\}_{k=1}^N$ at these grid points,
a natural way to produce an approximate solution path over $[0, t_{\max})$
is by linear interpolation.
In particular, we define a piecewise linear
function $\tilde \theta(t)$ as the
approximate solution path
through linearly interpolating the solutions at each grid point:
\begin{align*}
  \tilde \theta(t) & =  \frac{t_{k+1} - t}{t_{k+1} - t_k}\theta_k +
  \frac{t - t_{k}}{t_{k+1} - t_k} \theta_{k+1}
   \text{ for any } t \in [t_k, t_{k+1}],
   k = 0, 1, \ldots, N-1 \, ,   \\
   \tilde \theta(t) & =  \theta_{N} \text{ for any } t_N < t \leq t_{\max}
   \text{ if } t_N < t_{\max} \, ,
\end{align*}
where $t_0 = 0$ and $\theta_0 = \bm 0$. This defines an approximate solution path
$\tilde \theta(t)$ for any $t \in [0, t_{\max})$.
In view of this definition, we may also assume that $t_{N-1} \leq t_{\max}$, because
we do not need $\tilde \theta(t)$ over $t \in [t_{N-1}, t_N]$ if $t_{N-1} > t_{\max}$.
We also remark that the above interpolation scheme allows two possible approaches to
approximating the solution path around $t_{\max}$.
The first approach is to
specify all grid points from $[0, t_{\max})$
and use a constant path $\tilde \theta(t) = \theta_N$ to approximate $\theta(t)$
when $t_N < t \leq t_{\max}$. The other approach is to allow
 $t_N > t_{\max}$ but $t_{N-1} < t_{\max}$ when $t_{\max} < \infty$, and
use a linear interpolation of $\theta_{N-1}$ and $\theta_N$ to approximate $\theta(t)$
when $t_{N-1} < t < t_{\max}$.

To obtain the approximate solution path $\tilde \theta(t)$ as constructed above,
one also needs to choose the grid points $t_1, \ldots, t_N$ and a numerical algorithm to
generate the approximate solutions at these grid points, both of which will likely have
an impact on how well the solution path approximates the true path $\theta(t)$.
For the rest of this section, we first discuss
some path following algorithms that can produce solutions at a given set of grid points.
Given a path following algorithm, the issue of how to optimally design its grid points
to minimize the overall computations will be investigated later in Section 4.

\subsection{Path following algorithm: Newton}
In this subsection, we propose a path following algorithm
based on Newton update over a set of grid points.
A special version of this algorithm was considered in
\citet{rosset2004tracking} with $C(t) = 1 / t$.
The Newton method is constructed based on
taking one-step Newton steps at each grid point to obtain
an approximate solution at the next grid point.
More specifically, we consider an one-step Newton update at $t_{k+1}$ using
 $\theta_k$ as the initial solution, which can be shown to have the following form
\begin{equation}
  \theta_{k+1} = \theta_{k} -
  \left((1 - e^{-t_{k+1}})
   \nabla^2 L_n(\theta_{k})+ e^{-t_{k+1}} I \right)^{-1}
   \left( (1 - e^{-t_{k+1}}) \nabla L_n(\theta_{k}) +
  e^{-t_{k+1}} \theta_k \right) \, .
   \label{eq:Newton}
\end{equation}
To facilitate a comparison to the update of the Euler's method to be presented later
in \eqref{eq:euler_approx_iterates}, we present an alternative updating formula.
Let $g_k = (1 - e^{-t_k}) \nabla L_n(\theta_k) + e^{-t_k} \theta_k$
denote the scaled gradient at $\theta_k$.
By substituting $\theta_k$ with $e^{t_k} g_k - (e^{t_k} - 1)\nabla L_n(\theta_k)$
in \eqref{eq:Newton}, we obtain an alternative expression for the Newton update,
\begin{equation}
  \theta_{k+1} =
  \theta_{k} -
  \left( (1 - e^{-t_{k+1}}) \nabla^2 L_n(\theta_{k})+
   e^{-t_{k+1}} I\right)^{-1}
  \left((1 - e^{-\alpha_{k+1}})
  \nabla L_n(\theta_{k}) + e^{-\alpha_{k+1}}g_k \right) \, ,
   \label{eq:Newton_with_grad}
\end{equation}
where $\alpha_{k+1} = t_{k+1} -t_k$.
It will be shown later that the iterates generated by the Newton method
are all ``close'' to the true solution path $\theta(t)$ in some sense.
(c.f. Theorem \ref{thm:newton_global_bd}).
Moreover, it will be theoretically justified later that
only one Newton step is needed at each grid point as the overall
approximation error would not improve further if more Newton steps are taken
(c.f. Theorem \ref{thm:global_bd_newton}).
We also establish that the linearly interpolated solution path based on
Newton algorithm can achieve
$\epsilon$-suboptimality after taking at most $\mathcal O(\epsilon^{-1/2})$
 Newton iterations
(c.f. Theorem \ref{thm:complexity_newton}).

\subsection{Path following algorithm: gradient descent}
In this subsection,
we consider the gradient descent algorithm as the basis algorithm
for the path following scheme.
More specifically, at time $t_{k+1}$,
we perform $n_{k+1}$ gradient descent steps
to minimize $f_{t_{k+1}}(\theta) =
(1-e^{-t_{k+1}}) L_n(\theta) + (e^{-t_{k+1}}/2) \cdot \|\theta\|_2^2$
starting from $\theta_k$.
The update can be written down explicitly as
\begin{equation}
  \label{eq:grad_iterates}
  \theta_{k+1} = \circ^{n_{k+1}}
   (I - \eta_{k+1} \nabla f_{t_{k+1}}) \theta_k \, ,
\end{equation}
where $\eta_{k+1}$ is the gradient step size chosen at step $k+1$,
and $\circ^{l} h$ denotes $l$ function compositions of $h$.
In practice, a varying gradient step size can be implemented using
a line search.
As suggested by subsequent theoretical analysis
(see Theorem \ref{thm:gradient_convergence}),
multiple gradient descent steps are needed to ensure
a small approximation error and convergence.
This is in contrast to the Newton method, for
which one step is sufficient to achieve
 good approximation and convergence.
Moreover, the search direction at each step can be thought of
as a ``damped'' gradient descent search direction as we have that
\begin{equation}
  \nabla f_{t_k}(\theta) = (1 - e^{-t_k})
  \nabla L_n(\theta) + e^{-t_{k}} \theta \, ,
\end{equation}
which becomes closer and closer to the gradient search direction
$\nabla L_n(\theta)$ as $t_k$ increases.

In practice, the gradient descent method has the
advantage that it is typically cheaper to compute
as compared to the Newton method, although multiple steps need
to be taken in order for it to enjoy a good approximation-error bound
(c.f. Theorem \ref{thm:gradient_convergence} and \ref{thm:uniform_grad_bd}).
We also establish that the linearly interpolated solution path using the
gradient descent iterates can achieve
$\epsilon$-suboptimality after taking at most
$\mathcal O(\epsilon^{-1} \ln(\epsilon^{-1}))$ gradient descent iterations
(c.f. Theorem \ref{thm:complexity_gd}).


\subsection{Numerical ODE methods}
\label{numerical_ode}
In view of  Theorem \ref{thm:equiv_to_ode},
the solution path of \eqref{eq:path} is also the unique solution
of the ODE \eqref{eq:ode_special}.
Hence, any numerical methods that approximately solve
\eqref{eq:ode_special} with initial condition $\theta(0) = \bm 0$
would also produce an approximate solution path for \eqref{eq:path}.
In the numerical ODE literature,
the ODE \eqref{eq:ode_special} is often referred to as the \textit{initial value problem}
and standard solvers are available to find an approximate solution.
In this subsection, we consider two popular approaches:
the explicit forward Euler method and the second-order Runge-Kutta method \citep{butcher2016numerical}.

The explicit forward Euler method leads to
the following updating scheme:
\begin{equation}
  \label{eq:euler_approx_iterates}
  \theta_{k+1} = \theta_k  - \alpha_{k+1} \left[ ( 1- e^{-t_k}) \cdot
  \nabla^2 L_n(\theta_k) +
  e^{-t_k} I  \right]^{-1}\nabla L_n(\theta_k) \, .
\end{equation}
Note that if we choose a constant step size $\alpha_k = \alpha$, then
$t_k = k \alpha$.
Again this update is similar to the
Levenberg–Marquardt algorithm.
The difference here is
that the iterates are close to the true path as $\alpha \rightarrow 0$,
while in the Levenberg-Marquardt algorithm,
the goal is to recover the unregularized solution
as $k \rightarrow \infty$. Euler's method has been known to have
bad approximation error, and is referred to as first-order method
as the approximation error
$\|\theta_k - \theta(t_k)\|$ is typically of order $\mathcal O(\alpha)$ when
$\alpha_k = \alpha$ for all $k \geq 1$.

Higher order approximation can be achieved using
more sophisticated approximation schemes.
Runge-Kutta method is such a scheme whose
global approximation error is
$\|\theta_k - \theta(t_k)\| = \mathcal O(\alpha^m)$
with $m \geq 2$ when $\alpha_k = \alpha$ for all $k \geq 1$
\citep[see Chapter 3 of ][]{butcher2016numerical}.
Although it can achieve higher-order
 approximation accuracy compared to the
 Euler's method, it does require higher
 computational cost at each step.
For example,
the second-order Runge-Kutta method considers the following update
\begin{equation}
  \label{eq:Runge-Kutta}
  \theta_{k+1} = \theta_k  + \frac{\alpha_{k+1}}{2}
  \left( J(\theta_k, t_k) + J\left(\theta_k + \alpha_{k+1} J(\theta_k, t_k),
  t_{k+1}\right)
  \right) \, ,
\end{equation}
where $J(\theta, t) = -
\left( (1 -e^{-t}) \nabla^2 L_n(\theta(t))
  + e^{-t} I \right)^{-1} \nabla L_n(\theta(t))$. It can be immediately seen that, compared to
the Euler's method and the Newton method, it requires solving
two linear systems as opposed to just
one for the Euler method and Newton method.
Therefore, there is an apparent
 trade-off between
approximation error and per-iteration cost here.
Another popular choice is the
fourth-order Runge-Kutta method, which achieves a fourth-order
approximation accuracy,
but again requires solving four linear
systems at each iteration. Empirically, it will be demonstrated in Section \ref{sec:sims}
that the first-order ODE method
generally performs much worse than the Newton method, while the second-order
ODE method performs slightly worse than the Newton method.

\subsection{Discussion and connections}
The two types of updates are derived from two different perspectives.
The numerical ODE approach
tries to approximate the solutions to the corresponding ODE, while the
homotopy methods are based on applying path-following optimization algorithms with
warm-start.
Moreover, it is worth pointing out that
the updating formulas of the Euler's method and Newton method,
are very similar. In fact, the only difference is
the presence of an extra gradient term in the Newton update
 \eqref{eq:Newton_with_grad}. If we ignore the gradient term
 in the Newton update \eqref{eq:Newton_with_grad}, we have that
 \begin{eqnarray*}
   \theta_{k+1} &= & \theta_{k} - (1 - e^{-\alpha_{k+1}})
   \left( (1 - e^{-t_{k+1}}) \nabla^2 L_n(\theta_{k})
   + e^{-t_{k+1}} I \right)^{-1}
    \nabla L_n(\theta_{k}) \\
   & \approx &
   \theta_k  - \alpha_{k+1} \left[ ( 1- e^{-t_k}) \cdot
   \nabla^2 L_n(\theta_k) +
   e^{-t_k} I  \right]^{-1}\nabla L_n(\theta_k) \, ,
 \end{eqnarray*}
 where the right hand side is the Euler's update
  \eqref{eq:euler_approx_iterates}. In practice, however,
  we will show that the Newton method work much better than the
  Euler method in terms of approximation accuracy.

In terms of computational cost and ease of implementation,
the gradient descent update has the smallest per-iteration
cost, but it requires running more steps at each grid point,
especially when $t_k$ is large (c.f. Theorem \ref{thm:gd_bd}).
By contrast, the Newton method and the ODE solver
have higher per-iteration cost, but only requires one update at each grid point.
We also remark that other optimization
algorithms could also be used in the path following algorithm.
For example, glmnet \citep{friedman2010regularization}
uses coordinate descent algorithm in the path following
algorithm to get an approximate solution path.
Other viable choices include accelerated gradient descent or
 conjugate gradient descent algorithm. Hybrid approaches
that mix two types of algorithms can also be considered.
  We shall investigate these alternative approaches in the future.


\section{Solution path approximation-error bounds}
\label{sec:newton_approx_theory}
In this section, we derive approximation-error
bounds for the solution path over $[0, t_{\max})$
generated by the Newton method and gradient descent method.
The bounds for the ODE solvers have been extensively studied in the numerical
ODE literature, but are less satisfactory in that most results are proved for
generic ODE problems. We present one such version in Appendix \ref{sec:appendix_ode}.

We aim to bound the function-value suboptimality of an approximate solution path $\tilde \theta(t)$
measured by $\sup_{0 \leq t \leq t_{\max}} \{ f_t(\tilde \theta(t)) - f_t(\theta(t)) \}$,
where $f_t(\theta) := (1 - e^{-t})
  L_n(\theta) + (e^{-t} / 2) \|\theta\|_2^2$ is a scaled
version of the objective function.
Given the definition of $\theta(t)$, this is a natural performance
metric that captures the accuracy of the approximate solution path.
In what follows, we call $\sup_{0 \leq t \leq t_{\max}} \{ f_t(\tilde \theta(t)) - f_t(\theta(t)) \}$
the global approximation error for $\tilde \theta(t)$.
Our analysis proceeds in two steps: (i) we first
relate the global approximation error to
 approximation errors at the selected
grid points measured by the
size of the gradients $\|g_k\|_2$,
where $g_k := \nabla f_{t_k}(\theta_k) =
  (1 - e^{-t_{k}}) \nabla L_n(\theta_k) + e^{-t_{k}} \theta_k$;
(ii) we then bound $\|g_k\|_2$ for the Newton method and gradient descent method
proposed in Section \ref{sec:newton_approx}.

For step (i), we have the following result.
\begin{thm}
\label{thm:uniform_bd_general}
For any $0 < t_1 < t_2 <  \cdots < t_N < \infty$, we have that
\begin{align}
	 \sup_{t \in [0, t_{1}]}
	 \left\{ f_t(\tilde \theta(t)) - f_t(\theta(t)) \right\} & \leq
	 \max\left( e^{t_1} \|g_1\|_2^2 , \,
    \|\theta_1\|_2^2 \right) +
   \frac{e^{t_1} (1-e^{-t_1})^2}{2}
  \|\nabla L_n(\bm 0)\|_2^2 \, , \label{eq:thm_fn_bound1} \\
   \sup_{t \in [t_k, t_{k+1}]}
   \left\{ f_t(\tilde \theta(t)) - f_t(\theta(t)) \right\}  & \leq
   e^{t_{k+1}}
   \max \left\{\left( \frac{1-e^{-t_{k+1}}}{1 - e^{-t_k}} \right)^2  \|g_k\|_2^2, \,
  \|g_{k+1}\|_2^2 \right\} \nonumber \\
   &
   +  (e^{-t_k} - e^{-t_{k+1}})^2 \max \left\{
   \frac{e^{t_{k+1}} \|\theta_k\|_2^2 }{(1 - e^{-t_{k}})^2} , \,
   \frac{e^{t_k} \|\theta_{k+1}\|_2^2 }{(1 - e^{-t_{k+1}})^2} \right\}
    \label{eq:thm_fn_bound2}
\end{align}
for any $k = 1,\ldots,N-1$.
If we further assume that $\|\theta(t_{\max})\|_2 < \infty$, then
we have that
\begin{align}
	 \sup_{t_N < t \leq t_{\max}}  \left\{ f_t(\tilde \theta(t)) - f_t(\theta(t)) \right\}
   & \leq
  \frac{e^{t_N}(1-e^{-t_{\max}})}{1 - e^{-t_N}} \|g_N\|_2^2  +
\frac{3}{2(e^{t_N}-1)} \|\theta(t_{\max})\|_2^2  \label{eq:thm_fn_bound3}
\end{align}
when $t_N < t_{\max}$.
\end{thm}
We can see that the upper bounds consist of two parts, with
the first part (depending on $g_k$) being algorithm-specific and
 the other part stemming from interpolation over the selected grid points.
We call them \textit{optimization error} and \textit{interpolation error},
respectively.  Note that the optimization error depends
on the size of the gradient at time $t_k$ and is roughly of order
$e^{t_k} \|g_k\|_2^2$, while
the interpolation error is essentially independent of the choice of optimization algorithm as it
only depends on how finely we
choose the grid points and the norm of the solutions
along the solution path (typically $\|\theta_k\|_2 =
\mathcal O(\|\theta(t_k)\|_2)$, c.f., Lemma \ref{lm:thetak_theta_true_bd}).
In other words, given a specific set of grid points,
the interpolation error is irreducible for any optimization algorithms.
The optimization error, however, does depend on the optimization algorithms,
and can be pushed to be arbitrarily small if we run the algorithm long enough
at the selected grid points.
In this sense, if the goal is to approximate the solution path,
then both the grid points and the optimization algorithm
should be designed carefully to strike a balance between these two types of errors
to save the overall computation.
For instance, it would be wasteful to have
the optimization error much smaller than
the interpolation error, because the additional computations
would not improve the overall approximation error in terms of order.

We next derive bounds on $\|g_k\|_2$ for the Newton method and gradient descent method
to obtain an overall approximation-error bound for
$\sup_{0 \leq t \leq t_{\max}} \{ f_t(\tilde \theta(t)) - f_t(\theta(t)) \}$.
Using the bounds on $\|g_k\|_2$,
we then investigate how many Newton steps or gradient steps
are needed so that the optimization error can be dominated by the interpolation error.
Moreover, novel grid point schemes will be constructed to control
both the overall approximation error
and amount of computation.
This allows us to derive upper bounds on the total number of iterations
to achieve a prespecified suboptimality over the entire path for both methods.

\subsection{Newton method}
In this subsection, we show that
by taking only one Newton step at each grid point,
the optimization error is comparable to the interpolation error,
under some conditions on the grid points.
To show this, we first bound $\|g_k\|_2$ for the Newton method.
The following local
Lipschitz Hessian condition on $L_n(\theta)$ is assumed.

\vskip 0.1in
\noindent
\textbf{Assumption (A1).}
Assume that $L_n(\theta)$ is a proper, closed, convex function, and
there exists
constants $\beta > 0$,
$0 \leq \gamma_1 < 2$, and $0 \leq \gamma_2< 2$ such that
the second-order derivative of $L_n(\theta)$
 exists and satisfies a local Lipschitz
condition
\begin{equation}
\label{eq:hessian_cond}
  \|\nabla L_n(\theta + \delta) - \nabla L_n(\theta)
   - \nabla^2 L_n(\theta) \delta\|_2
  \leq \beta \delta^\top \left[\nabla^2
  L_n(\theta) \right]^{\gamma_1} \delta
\end{equation}
for any $\theta \in \textbf{dom}\, L_n$ and
$\delta$ satisfying $\theta + \delta \in \textbf{dom}\, L_n$ and
\begin{equation}
  \label{eq:Lipschitz_local_neighbor}
  \delta^\top [\nabla^2 L_n(\theta)]^{\gamma_2} \delta
  \leq \beta^{-2}  \, .
\end{equation}
Assumption (A1) can be thought of as a
local version of Lipschitz Hessian condition, and
is similar to the (generalized) self-concordant
condition imposed for the
convergence analysis of second-order method
\citep[see, e.g., ][]{nesterov1993, sun2017generalized}.
This avoids making the assumption that $L_n(\theta)$ is
strongly convex.
It will later be verified that many commonly used
loss functions satisfy Assumption (A1)
(see Table \ref{tab:function_examples}).
The following result provides bound on $\|g_k\|_2$ under Assumption (A1) and
some conditions on the step sizes $\{\alpha_k\}_{k =1}^\infty$.
\begin{thm}
  \label{thm:newton}
  Suppose that Assumption (A1) holds for some constants $\beta > 0$,
$0 \leq \gamma_1 < 2$,
and $0 \leq \gamma_2 < 2$.
We further assume that the step sizes $\{\alpha_k\}_{k = 1}^\infty$ satisfy
\begin{subequations}
\label{eq:step_size_conditions}
\begin{align}
   & C_1 \beta (e^{\alpha_1} - 1)^{\min(2-\gamma_1, 1-\gamma_2 / 2)}
 \|\nabla L_n(\bm 0)\|_2 \leq 1, \,
  \alpha_k < \ln(2),
  \, 2^{-1} \alpha_k \leq \alpha_{k+1} \leq 2 \alpha_k;
   \label{eq:step_size_condition_A} \\
  &
  C_2 \beta e^{t_{k+1}}
  \max\left((e^{t_{k+1}} - 1)^{-\gamma_1},
  (e^{t_{k+1}} - 1)^{-1-\gamma_2/2}\right)  (e^{\alpha_{k+1}} - 1) \|\theta_k\|_2
  \leq 1  \label{eq:step_size_condition_B}
\end{align}
 \end{subequations}
for any $k \geq 1$, where $C_1 = 15 \mathbb I(\gamma_1 \leq 1)
+ 15\min \left( \nu^{\gamma_1 - 1}, \,
 \nu^{\gamma_1} e^{-\alpha_1}(1- e^{-\alpha_1})
 \right)  \mathbb I(\gamma_1 > 1)$,
$C_2 = 442$, and $\nu$ denotes the maximum eigenvalue of
$\nabla^2 L_n(\bm 0)$.
Then, the scaled gradients $g_k$ evaluated at the
iterates $\theta_k$ generated by the Newton method in \eqref{eq:Newton}
satisfy
\begin{equation}
\label{eq:grad_bound_unified}
\|g_{k}\|_2  \leq  \frac{\|\theta(t_k)\|_2}{2(e^{t_k}-1)} (1 - e^{-\alpha_{k}})
\text{ for every } k \geq 1 \, .
\end{equation}
\end{thm}
Some remarks are in order.
First, fixing $t_k$, the upper bound for $\|g_k\|_2$
decreases as the step size $\alpha_k$ decreases.
In other words, smaller step size generally leads to a small upper bound.
Moreover, the first term in
the upper bound decreases as $k$ increases,
because
$\|\theta(t_k)\|_2 / (e^{t_k} - 1)$ is a nonincreasing function
of $t_k$ (c.f. part (ii) of Corollary \ref{cor:solution_prop}).
Second, the existence of step sizes that satisfy
\eqref{eq:step_size_conditions} is not obvious.
A novel step size scheme will be proposed later so that
it satisfies \eqref{eq:step_size_conditions} and at the same time
leads to fast exploration of the solution path.
Finally, we remark that the dependence of $C_1$ on
the largest eigenvalue of $\nabla^2 L_n(\bm 0)$ is to ensure that the bound
\eqref{eq:grad_bound_unified} holds for
$\|g_1\|_2$, and such dependence can be eliminated
if multiple Newton steps are taken at $t = t_1$ to ensure
\eqref{eq:grad_bound_unified} for $\|g_1\|_2$.

To facilitate a comparison to the theoretical analysis
of \citet{rosset2004tracking} and second-order
Runge-Kutta method,
an alternative bound on $\|g_k\|_2$ is presented below, which can be derived
using some partial results obtained in the proof of Theorem \ref{thm:newton}.
\begin{cor}
\label{cor:newton}
Under the assumptions in Theorem \ref{thm:newton}, we have that
the gradients $g_k$ evaluated at the
iterates $\theta_k$ generated by the Newton method in \eqref{eq:Newton}
satisfy
  \begin{subequations}
    \label{eq:grad_bound_unified_cor}
    \begin{align}
      \|g_1\|_2  & \leq  \frac{C_1}{15} \beta
\|\nabla L_n(\bm 0)\|_2^2
(e^{\alpha_1} - 1)^{\max(2,3-\gamma_1)} \, , \\
\|g_{k}\|_2  & \leq
   30\beta \left(\|\theta(t_{k-1})\|_2 + \|\nabla L_n(\bm 0)\|_2\right)^2
    \frac{e^{-\gamma_1 t_{k}} (e^{\alpha_{k}} - 1)^2}{(1 - e^{-t_{k}})^{\gamma_1-1}}
    \end{align}
  \end{subequations}
  for any $k \geq 2$.
\end{cor}
The above corollary can be viewed as an extension of Theorem 1
in \citet{rosset2004tracking},
which established that $\|\theta_k- \theta(t_k)\|_2 \lesssim \alpha^2$
when $t_k = t_0 + k\alpha$ is equally spaced
over a bounded interval $[t_0, t_{\max}]$ with $C(t) = 1/t$.
In particular,
we can see from \eqref{eq:grad_bound_unified_cor} that
$\|g_k\|_2 \lesssim \alpha_k^2$ when $\gamma_1 \leq 1$ or $t_k = \mathcal O(1)$.
In other words, when $\gamma_1 \leq 1$, we have $\|g_k\|_2 \lesssim \alpha_k^2$
for all $k \geq 1$; and when $\gamma_1 > 1$, we have
$\|g_k\|_2 \lesssim \alpha_k^2$ when $t_k$ is large enough.
This suggests that
the precision at the selected grid points for
the Newton method is often of order $\mathcal O(\alpha_k^2)$. This rate is comparable to
that derived in \citet{rosset2004tracking} and
that of the second-order Runge-Kutta method \citep[see Chapter 3 of][]{butcher2016numerical}
if a constant step size scheme is taken $\alpha_k = \alpha$.

Combining the bounds for $\|g_k\|_2$ in Theorem \ref{thm:newton} with
 Theorem \ref{thm:uniform_bd_general}, we
 show that for the Newton method, the optimization error is comparable to the interpolation error.
 Moreover, we can also
 obtain an approximation-error bound for the Newton solution path
 in terms of function-value suboptimality. This is summarized below.
\begin{thm}
\label{thm:global_bd_newton}
   Under the assumptions in Theorem \ref{thm:uniform_bd_general} and \ref{thm:newton},
  we have that the approximate solution path $\tilde \theta(t)$ generated by the Newton method satisfies
 \begin{multline}
   \sup_{0 \leq t \leq t_{\max}}   \left\{ f_t(\tilde \theta(t)) - f_t(\theta(t)) \right\}
   \\
   \leq  8 \max \left\{(e^{\alpha_1} - 1)^2 \|\nabla L_n(\bm 0)\|_2^2, \,
\max_{1 \leq k \leq N} e^{-t_{k}}
  \left( \frac{e^{\alpha_{k+1}} - 1}{1 - e^{-t_{k}}} \right)^2
     \|\theta_k\|_2^2
  \right\} \,
    \label{eq:newton_uniform_general_bd2}
\end{multline}
when $t_{N-1} \leq t_{\max} < t_N$ for some $N \geq 1$; and
 \begin{multline}
   \sup_{0 \leq t \leq t_{\max}}   \left\{ f_t(\tilde \theta(t)) - f_t(\theta(t)) \right\}
  \leq \max\Bigg\{ 8 (e^{\alpha_{1}} - 1)^2 \|\nabla L_n(\bm 0)\|_2^2, \,
   \\
   8 \max_{1 \leq k \leq N-1} e^{-t_{k}}
  \left( \frac{e^{\alpha_{k+1}} - 1}{1 - e^{-t_{k}}} \right)^2
     \|\theta_k\|_2^2   \, ,
\frac{2\max(\|\theta(t_{\max})\|_2^2, \|\theta_N\|_2^2)}{(e^{t_N}-1)}  \Bigg\} \,
    \label{eq:newton_uniform_general_bd1}
\end{multline}
when $0 < t_N < t_{\max}$ for some $N \geq 2$ and
$\|\theta(t_{\max})\|_2 < \infty$.
\end{thm}

In the proof of the above theorem, it is shown that taking just one Newton step at each grid point
can ensure that the optimization error is comparable to the
interpolation error.
Specifically, it is shown in the proof of Theorem \ref{thm:global_bd_newton}
that for all $k \geq 1$,
\begin{multline}
\label{eq:newton_opt_int_error_comp}
 \underbrace{e^{t_{k+1}}
\max \left\{\left( \frac{1-e^{-t_{k+1}}}{1 - e^{-t_k}} \right)^2  \|g_k\|_2^2, \,
  \|g_{k+1}\|_2^2 \right\}}_{\text{optimization error}} \\
  \lesssim
   \underbrace{(e^{-t_k} - e^{-t_{k+1}})^2 \max \left\{
   \frac{e^{t_{k+1}} \|\theta_k\|_2^2 }{(1 - e^{-t_{k}})^2} , \,
   \frac{e^{t_k} \|\theta_{k+1}\|_2^2 }{(1 - e^{-t_{k+1}})^2} \right\}
   }_{\text{interpolation error}} \, ,
\end{multline}
where the LHS is the optimization error and
the RHS is the interpolation error in the bounds in Theorem \ref{thm:uniform_bd_general}.
In this sense, it is
wasteful to take more than one Newton step at each grid point.

Another important consequence of Theorem \ref{thm:newton} and
\ref{thm:global_bd_newton} is that
a principled scheme of choosing the step
sizes (or equivalently the grid points) can be designed
to ensure any prespecified level of suboptimality while minimizing the overall computations.
More specifically, for any $\epsilon > 0$ and $t_{\max} > 0$,
suppose that our goal is to design a step size scheme that satisfies
all the conditions in \eqref{eq:step_size_conditions} and at the same
time ensures that $\sup_{0 \leq t \leq t_{\max}}
\left\{ f_t(\tilde \theta(t)) - f_t(\theta(t)) \right\}
\lesssim \epsilon$.
In view of \eqref{eq:newton_uniform_general_bd1}
in Theorem \ref{thm:global_bd_newton},
this amounts to running the Newton method by choosing
a sequence of step sizes $\{\alpha_k\}$ satisfying both \eqref{eq:step_size_conditions} and
 \begin{align}
  & (e^{\alpha_{1}} - 1)^2 \|\nabla L_n(\bm 0)\|_2^2 \lesssim \epsilon \, ,
  e^{-t_k}\left( \frac{e^{\alpha_{k+1}} - 1}{1 - e^{-t_{k}}} \right)^2
     \|\theta_k\|_2^2 \lesssim (e^{\alpha_{1}} - 1)^2 \|\nabla L_n(\bm 0)\|_2^2 \, ,
     \label{eq:step_size_additional_conditions}
\end{align}
and terminating the Newton method at $k + 1 = N$ when
\begin{equation}
\label{eq:termination_criterion}
t_{N} > t_{\max} \text{ or }
 \frac{2\|\theta_N\|_2^2}{(e^{t_N}-1)}
   \leq (e^{\alpha_{1}} - 1)^2 \|\nabla L_n(\bm 0)\|_2^2 \, .
\end{equation}
If such a sequence of step sizes $\{\alpha_k\}$ exists, then by its construction and
\eqref{eq:newton_uniform_general_bd1}
of Theorem \ref{thm:global_bd_newton}, we have that
\begin{equation}
  \sup_{0 \leq t \leq t_{\max}}
\left\{ f_t(\tilde \theta(t)) - f_t(\theta(t)) \right\} \lesssim
(e^{\alpha_{1}} - 1)^2 \|\nabla L_n(\bm 0)\|_2^2 \lesssim \epsilon \, .
\end{equation}
Therefore, it remains to prove the existence of such a sequence satisfying
all the conditions in \eqref{eq:step_size_conditions} and
\eqref{eq:step_size_additional_conditions} for any $\epsilon > 0$, and that
the Newton method must terminate within finite steps.
This is shown in the theorem below.
\begin{thm}
\label{thm:newton_global_bd}
Suppose that $\|\theta(t_{\max})\|_2 < \infty$ with
$t_{\max} \in (0, \infty]$.
For any $\epsilon > 0$ and $0 < \alpha_{\max} \leq 10^{-1}$,
using the step sizes defined below
\begin{align}
\alpha_1 & \leq \min  \Bigg\{\alpha_{\max}, \,
 \ln\left(1+\frac{\sqrt{\epsilon}}{\|\nabla L_n(\bm 0)\|_2}\right), \, \nonumber \\
 & \qquad \qquad \qquad
\ln\left(1 + (\max(C_1, \sqrt{2}C_2)
\beta\|\nabla L_n(\bm 0)\|_2)^{-\min(2-\gamma_1, 1 -\gamma_2 / 2)} \right)
\Bigg \} \text{ and } \label{eq:newton_stepsize_cond_ideal_alpha1} \\
  \alpha_{k+1} & = \min\Bigg\{ \alpha_{\max}, 2\alpha_{k}, \,
  \ln \left(1 + \frac{e^{t_k/2}(e^{\alpha_1} - 1)\|\nabla L_n(\bm 0)\|_2(1-e^{-t_k})}
  {\|\theta_k\|_2}
  \right)\, ,
 \nonumber \\
& \quad  \ln\left( 1 + \left(C_2\beta e^{t_{k}}
  \max\left((e^{t_{k}} - 1)^{-\gamma_1},
  (e^{t_{k}} - 1)^{-1-\gamma_2/2}\right)
  \|\theta_k\|_2\right)^{-1} \right)
 \Bigg\}; k \geq 1  \label{eq:newton_stepsize_cond_ideal_alphak}
\end{align}
and the termination criterion in \eqref{eq:termination_criterion},
the Newton method terminates after a
finite number of iterations, and when terminated, the
generated solution path $\tilde \theta(t)$ satisfies
\begin{equation}
\label{eq:cor2_bd_global}
  \sup_{0 \leq t \leq t_{\max}}
\left\{ f_t(\tilde \theta(t)) - f_t(\theta(t)) \right\}
\lesssim \epsilon \, .
\end{equation}
\end{thm}
This result confirms the existence of a step size sequence that ensures
any prespecified suboptimality for the solution path generated by the Newton method.
The step size choices in \eqref{eq:newton_stepsize_cond_ideal_alpha1} and
\eqref{eq:newton_stepsize_cond_ideal_alphak} are motivated by
\eqref{eq:step_size_additional_conditions}.
Moreover, as we can see from \eqref{eq:newton_stepsize_cond_ideal_alpha1} and
 \eqref{eq:cor2_bd_global}, the suboptimality $\epsilon$
 is controlled by the initial step size $\alpha_1$.
Indeed, for small enough $\epsilon$, we can see that
$\epsilon = (e^{\alpha_1} - 1)^2\|\nabla L_n(\bm 0)\|_2^2$, which implies that
\begin{equation}
\label{eq:cor2_bd_global_alt}
   \sup_{0 \leq t \leq t_{\max}}
\left\{ f_t(\tilde \theta(t)) - f_t(\theta(t)) \right\}
\lesssim
(e^{\alpha_1} - 1)^2\|\nabla L_n(\bm 0)\|_2^2 \, .
\end{equation}
Importantly, the above result suggests that
even when $t_{\max} = \infty$, we can achieve arbitrarily small suboptimality for the entire
path using a finite number of grid points.
To the best of our knowledge, this type of theoretical analysis is
new in the literature for path following algorithms.

We next investigate how fast the Newton method explores the solution path by
deriving its computational complexity.
As we can see from the step size scheme in Theorem \ref{thm:newton_global_bd},
both the value of $\gamma_1$ and the speed that $\|\theta(t)\|_2$
grows as a function of $t$ will likely have a big
impact on how aggressively we can choose the step sizes $\alpha_k$.
In particular, if $\gamma_1 \geq 1$ and $\|\theta_k\|_2$ is bounded
(e.g., when $\theta^\star$ is finite), then
the last term in the $\min$ function of \eqref{eq:newton_stepsize_cond_ideal_alphak}
is at least of order
$\mathcal O(1)$, while the third term in the $\min$ function is increasing. Therefore,
aggressive step sizes can be taken in this case
until it reaches $\mathcal O(1)$, which will likely lead to
a fast exploration of the solution path.
On the other hand, if $\gamma_1 < 1$ or $\|\theta_k\|_2$
grows quite quickly to infinity as $k$ increases,
then the last term in the $\min$ function goes
 to zero as $k \rightarrow \infty$.
This means that the step sizes need to
 decrease to zero eventually,
 leading to a slower exploration of the solution path.
The following result gives an upper bound
on the number of Newton steps needed for the
Newton method when $\gamma_1 \geq 1$ and
$\|\theta(t_{\max})\|_2$ is treated as a finite constant.
\begin{thm}
\label{thm:complexity_newton}
 Suppose that $\|\theta(t_{\max})\|_2 < \infty$
and $\gamma_1 \geq 1$. For any $\epsilon > 0$, using the
  step sizes defined in \eqref{eq:newton_stepsize_cond_ideal_alpha1} and
  \eqref{eq:newton_stepsize_cond_ideal_alphak},
  and the termination criterion in \eqref{eq:termination_criterion},
  the number of Newton steps required to achieve $\epsilon$-suboptimality
  \eqref{eq:cor2_bd_global} is at most $\mathcal O(\epsilon^{-1/2})$.
\end{thm}
We remark the above result holds
even when $t_{\max} = \infty$.
Moreover, it is rather difficult to theoretically bound
the number of Newton steps when $\gamma_1 < 1$.
Fortunately, for many commonly used loss functions,
Assumption (A1) holds with some $\gamma_1 \geq 1$, as demonstrated by
the following proposition.
\begin{pro} \leavevmode
\label{pro:function_examples}
The Assumption (A1) holds for
\begin{itemize}
\item log-barrier function $L_n(\theta)=-\ln(\theta)$
with $\gamma_1=\frac{3}{2}$ and $\gamma_2=1$;
\item entropy-barrier function $L_n(\theta)=\theta\ln(\theta)-\ln(\theta)$
with $\gamma_1=\frac{3}{2}$ and $\gamma_2=1$;
\item logistic function $L_n(\theta)=\ln(1+e^{-\theta})$
with $\gamma_1=1$ and $\gamma_2=0$;
\item exponential function $L_n(\theta)=e^{-\theta}$
with $\gamma_1=1$ and $\gamma_2=0$.
\item square function $L_n(\theta)=\theta^2$
with any $\gamma_1\in [0,2)$ and $\gamma_2\in [0,2)$.
\end{itemize}
\end{pro}
We summarize these results in Table \ref{tab:function_examples}.
A detailed proof is provided in the Appendix.

\begin{table}[h]
\centering
\begin{tabular}{l l l l l l l}
\hline \hline
Function  & form of $L_n(\theta)$ & \textbf{dom}($L_n$) & $\gamma_1$ & $\gamma_2$  & Application \\
\hline
Log-barrier  & $-\ln(\theta)$ & $\mathbb R^{++}$ &$\frac{3}{2}$ &$1$  & Poisson regression \\
Entropy-barrier  & $\theta\ln(\theta)-\ln(\theta)$ & $\mathbb R^{++}$ &$\frac{3}{2}$ &1 & Interior-point \\
Logistic  & $\ln(1+e^{-\theta})$ & $\mathbb R$ &1 &0   & Logistic regression \\
Exponential  & $e^{-\theta}$ & $\mathbb R$ &1 &0  & Boosting \\
Square & ${\theta}^{2}$ & $\mathbb R$ &[0,2) &[0,2)  & Least square regression\\
\hline \hline
\end{tabular}
\caption{Some commonly used loss functions that satisfy Assumption (A1). }
\label{tab:function_examples}
\end{table}
As such, Theorem \ref{thm:complexity_newton} applies to all losses listed
in Table \ref{tab:function_examples} since Assumption (A1) is satisfied with
$\gamma_1 \geq 1$ for all losses. Thus the total number of Newton steps
 required to ensure \eqref{eq:cor2_bd_global}
is at most $\mathcal O(\epsilon^{-1/2})$ for these loss functions.

Finally, we note that our theoretical results for the Newton method
are widely applicable to a large class of functions.
It can even include loss functions that are not self-concordant,
which is a typical condition imposed to
establish complexity bound for the classical Newton method
\citep[see, e.g., ][]{nesterov1993} without making strong convexity
assumptions. For example, among the losses in Table \ref{tab:function_examples},
the logistic regression loss
function $\log(1+e^{-\theta})$ and the exponential loss
$e^{-\theta}$ are not self-concordant.
Indeed, a separate rate of convergence analysis is needed for the Newton method
when applied to logistic regression problems \citep[see, e.g., ][]{bach2010self}.
On the other hand, the generality of our analysis likely will lead to conservative
rates and step size choices for problems with better conditioning.
For instance, we expect that some of the above results can be improved and a better
step size scheme can be constructed if we assume that
the loss function $L_n(\theta)$ is strongly convex or
``locally'' strongly convex along the solution path $\theta(t)$.
Due to space limit, we leave this for future investigation.

%
\subsection{Gradient descent method}
We next bound $\|g_k\|_2$ for the gradient descent method
proposed in Section \ref{sec:newton_approx}.
We then use the bound to derive conditions on the number
of gradient steps needed to ensure that
the optimization error is comparable to the interpolation error.
For gradient descent method, we
impose the following Lipschitz gradient
assumption on $L_n(\theta)$.

\vskip .1in
\noindent
\textbf{Assumption (A2). }
Assume that $L_n(\theta)$ has
$L$-Lipschitz continuous gradient:
\begin{equation}
  \|\nabla L_n(\theta_1) - \nabla L_n(\theta_2)\|_2
  \leq L \|\theta_1 - \theta_2 \|_2 \, .
\end{equation}
\vskip -0.1in
\begin{thm}
\label{thm:gradient_convergence}
Let $m_k = m (1-e^{-t_{k}}) + e^{-t_{k}}$, $L_k = L (1-e^{-t_{k}}) + e^{-t_{k}}$,
where $m \geq 0$ is the strong convexity parameter for $L_n(\theta)$.
Under Assumption (A2) and the condition that
\begin{equation}
\label{eq:grad_cond_unified}
  n_{1} \geq \frac{\log(10m_1)}
  {-\log\left(1 - \frac{2m_{1}L_{1}}{m_{1}+L_{1}} \eta_{1} \right)}, \,
  n_{k+1} \geq \frac{\log(24)}
  {-\log\left(1 - \frac{2m_{k+1}L_{k+1}}{m_{k+1}+L_{k+1}} \eta_{k+1} \right)}
  \text{ and } \eta_{k} \leq \frac{2}{m_{k}+L_{k}}
\end{equation}
for any $k \geq 1$,
the iterates generated by the gradient descent method
(defined by \eqref{eq:grad_iterates}) satisfies
\begin{equation}
\label{eq:grad_approx_bound}
  \|g_k\|_2  \leq
  \frac{2 L_k}{m_{k-1}}  \left(1 - \frac{2m_{k}L_{k}}{m_{k}+L_{k}} \eta_{k}\right)^{n_{k}}
  \frac{(e^{\alpha_{k}} - 1)\|\theta(t_{k})\|_2}{(e^{t_{k}} - 1)}
\end{equation}
for any $k \geq 1$ and step sizes
$\alpha_k \leq \ln(2)$ satisfying
$2^{-1} \alpha_k \leq \alpha_{k+1} \leq 2\alpha_{k}; k \geq 1$.
\end{thm}
As we can see from the condition on $n_k$ in \eqref{eq:grad_cond_unified},
the number of gradient steps needed
at each grid point is likely to be more than one to ensure \eqref{eq:grad_approx_bound}.
This is in contrast to the Newton method, for which
only one Newton step is taken at each iteration. It will be shown later that
taking multiple gradient steps is necessary to ensure that the optimization error
is comparable to the interpolation error.
Moreover, we can see that
when $m > 0$, that is, when $L_n(\theta)$ is $m$-strongly convex with $m > 0$,
then the lower bound on $n_k$ behaves like a constant.
When $m = 0$, however, then $m_k = e^{-t_k}$ and
the number of gradient steps $n_k$ scales as $\mathcal O\left(e^{t_k}\right)$
in the worst case, suggesting that the number of gradient steps needed should increase
as $k$ increases.

Interestingly, unlike the Newton method,
the optimization error bound for gradient descent method
may not be dominated by the interpolation error. In order for
the optimization error to be comparable to the interpolation error,
more gradient steps need to be taken beyond what is required
in \eqref{eq:grad_cond_unified}.
The following theorem derives conditions on $n_k$ under which the
the optimization error is dominated by the interpolation error, and establishes
an approximation-error bound for the solution path generated by the gradient descent method
building on Theorem \ref{thm:uniform_bd_general} and \ref{thm:gradient_convergence}.
\begin{thm}
\label{thm:uniform_grad_bd}
  Under the assumptions in Theorem \ref{thm:gradient_convergence} with
  \eqref{eq:grad_cond_unified} replaced by
  \begin{align}
  \label{eq:grad_stepsize_relaxed}
  & n_{1} \geq \frac{\log(10m_1 L_1)}
  {-\log\left(1 - \frac{2m_{1}L_{1}}{m_{1}+L_{1}} \eta_{1} \right)}, \,
     n_{k+1} \geq \frac{\log(24) + \max(0, \log(L_{k+1}/m_{k}))}
  {-\log\left(1 - \frac{2m_{k+1}L_{k+1}}{m_{k+1}+L_{k+1}} \eta_{k+1} \right)}, \,
  \eta_{k} \leq \frac{2}{m_{k}+L_{k}}
  \end{align}
for $k \geq 1$, the approximate solution path $\tilde \theta(t)$ generated by the gradient descent
  method satisfies
  \begin{multline}
    \sup_{0 \leq t \leq t_{\max}}   \left\{ f_t(\tilde \theta(t)) - f_t(\theta(t)) \right\}
    \\
    \leq   2  \max \Bigg( (e^{\alpha_1} - 1)^2 \|\nabla L_n(\bm 0)\|_2^2 \, ,
     \max_{ 1 \leq k \leq N-1}
e^{-t_{k}} \left( \frac{e^{\alpha_{k+1}} - 1}{1 - e^{-t_{k}}} \right)^2 \|\theta_k\|_2^2
 \Bigg) \, .
    \label{eq:grad_uniform_general_bd_strong2}
\end{multline}
when $t_N \geq t_{\max}$ for some $N$; and
  \begin{multline}
    \sup_{0 \leq t \leq t_{\max}}   \left\{ f_t(\tilde \theta(t)) - f_t(\theta(t)) \right\}
    \leq 2 \max \Bigg( (e^{\alpha_1} - 1)^2 \|\nabla L_n(\bm 0)\|_2^2 \, ,  \\
     \max_{ 1 \leq k \leq N-1}
   e^{-t_{k}} \left( \frac{e^{\alpha_{k+1}} - 1}{1 - e^{-t_{k}}} \right)^2 \|\theta_k\|_2^2 \, ,
     \frac{\|\theta(t_{\max})\|_2^2}
   {(e^{t_N} - 1)}  \Bigg)
    \label{eq:grad_uniform_general_bd_strong1}
\end{multline}
when $t_N \leq t_{\max}$ and $\|\theta(t_{\max})\|_2 < \infty$.
\end{thm}
Compared to Theorem \ref{thm:gradient_convergence},
the conditions \eqref{eq:grad_stepsize_relaxed} on $n_k$ in the above theorem
are stronger than the conditions \eqref{eq:grad_cond_unified},
which is to ensure that the optimization error is dominated by the interpolation error.
Moreover, it is unnecessary to run more than those required by
the conditions in \eqref{eq:grad_stepsize_relaxed},
as taking beyond this many gradient steps would not improve the overall
approximation error for the entire path (at least in terms of order).

Similar to the Newton method,
a novel step size scheme can be designed
to ensure that the approximation error is small for all $k \geq 1$.
In particular, we choose
\begin{align}
\alpha_1 & \leq \min  \Bigg\{\alpha_{\max}, \,
 \ln\left(1+\frac{\epsilon^{1/2}}{\|\nabla L_n(\bm 0)\|_2}\right)
\Bigg \} \text{ and } \label{eq:GD_stepsize_cond_alpha1} \\
  \alpha_{k+1} & = \min\Bigg\{ \alpha_{\max}, 2\alpha_{k}, \,
  \ln \left(1 + \frac{\epsilon^{1/2} e^{t_k/2}(1-e^{-t_k})}
  {\|\theta_k\|_2}
  \right) \Bigg\}; k \geq 1 \, ,  \label{eq:GD_stepsize_cond_alphak}
\end{align}
where $\alpha_{\max} = \ln(2)$,
and terminate the algorithm at $k + 1 = N$ when
\begin{equation}
\label{eq:termination_criterion_GD}
t_{N} > t_{\max} \text{ or }
 \frac{2\|\theta_N\|_2^2}{(e^{t_N}-1)}
 \leq \epsilon \, .
\end{equation}
Similar to the Newton method, we show that the solution path generated by the
gradient descent method
using the above step size scheme and termination criterion
achieves $\epsilon$-suboptimality (up to a multiplicative constant).
This is summarized in the following theorem.

\begin{thm}
  \label{thm:gd_bd}
  Suppose that $\|\theta(t_{\max})\|_2 < \infty$ with
   $t_{\max} \in (0, \infty]$, and
\eqref{eq:grad_stepsize_relaxed} in Theorem \ref{thm:uniform_grad_bd} is satisfied.
For any $\epsilon > 0$,  using the step sizes and
the termination criterion specified above in
\eqref{eq:GD_stepsize_cond_alpha1}, \eqref{eq:GD_stepsize_cond_alphak},
and \eqref{eq:termination_criterion_GD},
the gradient descent method
terminates after a
finite number of iterations, and when terminated, the generated
solution path $\tilde \theta(t)$ satisfies
\begin{equation}
\label{eq:GD_thm_bd_global}
  \sup_{0 \leq t \leq t_{\max}}
\left\{ f_t(\tilde \theta(t)) - f_t(\theta(t)) \right\}
\lesssim \epsilon \, .
\end{equation}
\end{thm}
Next, we derive the computational complexity of the gradient descent method.
To make it directly comparable to the Newton method, we
consider the case $m = 0$.
In this case, in order for the optimization error to be comparable to the interpolation error,
$n_{k+1}$ must satisfy \eqref{eq:grad_stepsize_relaxed}, which
can be shown to be equivalent to
$n_{k+1} \geq \mathcal O(e^{t_k}(t_k + 1))$.
Building on this,
an upper bound
on the number of gradient steps needed can be derived when
$\|\theta(t_{\max})\|_2$ is treated as a finite constant.
This is summarized in the following theorem.
\begin{thm}
\label{thm:complexity_gd}
 Suppose that $\|\theta(t_{\max})\|_2 < \infty$, and
the assumptions in Theorem \ref{thm:uniform_grad_bd}
 are met with $m = 0$. Using $\eta_k =  \mathcal O(\min(1,L^{-1}))$, the
  step sizes defined in \eqref{eq:GD_stepsize_cond_alpha1} and
  \eqref{eq:GD_stepsize_cond_alphak},
  and the termination criterion in \eqref{eq:termination_criterion_GD},
  the number of gradient steps required to achieve $\epsilon$-suboptimality
  \eqref{eq:GD_thm_bd_global} is
  at most $\mathcal O\left(\epsilon^{-1} \ln(\epsilon^{-1})\right)$.
\end{thm}
Compared with the Newton method that requires $\mathcal O(\epsilon^{-1/2})$ number
of Newton steps, gradient descent method requires substantially more updates.
Of course, since the per-iteration
cost of the gradient descent method is much lower than that of Newton method,
an overall computational-complexity comparison depends on how problem
dimension scales with suboptimality
$\epsilon$.
In general, we expect that the Newton method
may be more suitable for small to
medium scale problems or when a small suboptimality is desired,
whereas gradient descent method may be more suitable for
large scale problems with medium accuracy.
This will also be confirmed through some numerical experiments in Section \ref{sec:sims}.
As a side remark, a hybrid approach combining the gradient descent method and
the Newton method is likely to work better than either one. Due to space limit,
we choose to investigate this strategy in the future.

Moreover, for the unregularized problem,
it is well-known that the number of gradient steps required for the
regular gradient descent method to achieve
an $\epsilon$-suboptimality (i.e., $L_n(\theta_k) - L_n(\theta^\star) <
\epsilon$) is $\mathcal O(\epsilon^{-1})$ when $m = 0$.
In view of this and the above result, one can essentially claim that
for the gradient
descent method starting from $\theta_0 = \bm 0$,
computing the entire solution path for the $\ell_2$-regularized problem
requires roughly the same amount of computation as compared
to computing a single unregularized solution (up to a logarithm term $\ln(\epsilon^{-1})$).

The implementation of the gradient descent method requires the specification
of $n_k$ and $\eta_k$, both of which depend on unknown problem-specific parameters
$m$ and $L$ (see \eqref{eq:grad_stepsize_relaxed}). In practice, we implement
the gradient method using a backtracking line search \citep{boyd2004convex} and
terminates the gradient descent method at $t_k$ when
\begin{equation}
\label{eq:gd_termination_criterion_at_tk}
  \|g_k\|_2 \leq \frac{e^{\alpha_{k}} - 1}{C_0(e^{t_{k}} - 1)}\|\theta_k\|_2 \, ,
\end{equation}
for some absolute constant $C_0$.
In the proof of Theorem \ref{thm:uniform_grad_bd}, it is shown that
if $n_k$ and $\eta_k$ satisfy the conditions in \eqref{eq:grad_stepsize_relaxed},
then \eqref{eq:gd_termination_criterion_at_tk} holds
for $C_0 = 12$.
Here if we use \eqref{eq:gd_termination_criterion_at_tk}
directly as a termination criterion for the gradient descent method at $t_k$,
we can still establish the approximation-error bound in
Theorem \ref{thm:uniform_grad_bd} and \ref{thm:gd_bd}.
\begin{cor}
\label{cor:GD_bound}
  Suppose that $\|\theta(t_{\max})\|_2 < \infty$ with
   $t_{\max} \in (0, \infty]$. Moreover,
   we assume that at each $t_k$, we run the gradient descent method with backtracking line
   search until  \eqref{eq:gd_termination_criterion_at_tk} is satisfied for some
   absolute constant $C_0$. Then for any $\epsilon > 0$,  using the step sizes and
the termination criterion specified in
\eqref{eq:GD_stepsize_cond_alpha1}, \eqref{eq:GD_stepsize_cond_alphak},
and \eqref{eq:termination_criterion_GD},
the gradient descent method
terminates after a
finite number of iterations, and when terminated, the generated
solution path $\tilde \theta(t)$ satisfies
\begin{equation}
\label{eq:GD_cor_bd_global}
  \sup_{0 \leq t \leq t_{\max}}
\left\{ f_t(\tilde \theta(t)) - f_t(\theta(t)) \right\}
\lesssim \epsilon \, .
\end{equation}
\end{cor}
Again, the advantage of using the backtracking line search and the termination criterion
 \eqref{eq:gd_termination_criterion_at_tk} for the gradient descent is that
it avoids having to specify $n_k$ and $\eta_k$, both of which may depend on
unknown problem-specific parameters $m$ and $L$.

\section{Numerical studies}
\label{sec:sims}
In this section, we use $\ell_2$-regularized logistic regression
as an illustrating
example to study the operating characteristics of the various
proposed methods.
Let $X=(X_1,\ldots,X_n)^\top$ and $Y = (Y_1,\ldots, Y_n)^\top$
denote the design matrix and the binary response vector, where
$X_i \in \mathbb R^p$ and $Y_i \in \{+1, -1\}$; $i =  1,\ldots, n$.
The empirical loss function for logistic regression  is
\begin{equation}
  \label{eq:logistic_loss}
L_n(\theta) = \frac{1}{n}\sum_{i=1}^n \log(1+e^{-Y_i X_i^\top\theta}) \, .
\end{equation}
We first verify that the above loss function satisfies Assumption (A1).
\begin{pro}
\label{pro:logistic_loss}
The logistic regression loss function
$L_n(\theta)$ defined in \eqref{eq:logistic_loss}
 satisfies Assumption (A1) with
 $\gamma_1 = 1, \gamma_2 = 0$, and $\beta = 2
 \max_{1 \leq i \leq n} \|X_i\|_2$.
\end{pro}
In view of the above results,
Theorem \ref{thm:complexity_newton} can be applied
to logistic regression if $\|\theta(t_{\max})\|_2 < \infty$.
We note that for logistic regression, the MLE could be at the ``infinity''
i.e., $\|\theta^\star\|_2 = \infty$, when
the two classes are separable \citep[see, e.g., ][]{geyer2009}.

In our numerical experiments, we consider
six methods:
Euler method, second-order Runge-Kutta method, Newton method,
the method proposed by \cite{rosset2004tracking}, gradient descent method, and
 glmnet \citep{friedman2010regularization}.
The first four methods are ``second-order''
algorithms in the sense that they all involve
solving linear systems.
Gradient descent method only requires gradient evaluations, and
glmnet uses warm start strategies and cyclical coordinate descent method
to compute an approximate solution path.
We implement all methods in R using Rcpp \citep{eddelbuettel2011rcpp,eddelbuettel2013seamless},
except for glmnet for which we use the R package glmnet.
We remark that the method of \cite{rosset2004tracking} is
also a path-following algorithm based on Newton updates.
Compared with our proposed Newton method,
it considers equally-spaced grid points using $C(t) = 1/t$ and
starts with an initial solution $\theta(t_{\max})$ at $t_{\max}$.
As will be demonstrated later, this makes it less efficient compared with the
proposed Newton method.
Finally, we point out that the proposed Newton method and gradient descent
method can be applied to the case $t_{\max}=\infty$ for the nonseparable case,
while all the other four methods can be only applied to
the case $t_{\max}<\infty$.
Throughout, we use $t_{\max}=10$ in all of the numerical experiments.
Increasing $t_{\max}$ further will make the proposed methods even more competitive
in the comparisons.

We first compare all methods in terms of runtime and suboptimality.
Two scenarios will be considered depending on whether the two classes are separable or not.
For the nonseparable case, we sample the components of the response vector
$Y \in \mathbb R^n$ from a Bernoulli distribution,
where $\mathbb P(Y_{i}=+1)=1/2$ and $\mathbb P(Y_{i}=-1)=1/2$ for $i=1,2, \ldots ,n$.
Conditioned on $Y_i$, we generate $X_i$'s
independently from $N_p(Y_i \mu, \sigma^2 I_{p \times p})$,
where $\mu \in \mathbb R^p$ and $\sigma^2>0$.
Note that $\mu$ and $\sigma^2$ controls the Bayes risk, which
is $\Phi(-\|\mu\|_2 / \sigma)$ under the 0/1 loss,
where $\Phi(\cdot)$ is the cumulative distribution function of a
standard normal random variable.
Here we choose $\mu=(1/\sqrt{p},\ldots,1/\sqrt{p})$ and $\sigma^2=1$
so that the Bayes risk is $\Phi(-1)\approx 0.15$.
For the separable case, we generate
$X_i$'s independently from $N_p(Y_i\mu, I_{p \times p})$ where
$\mu=(1/\sqrt{p},\ldots,1/\sqrt{p})$
until $Y_i \mu^\top X_i>1$,
which makes the two classes linearly separable.
In  fact, the two classes can be separated by
the hyperplane $\mu^\top X_i= 0$.
For both scenarios, three choices of problem dimensions are considered:
 $(n,p)=(1000, 500)$, $(n,p)=(1000,1000)$, and $(n,p)=(1000, 2000)$.


To assess the accuracy for the approximate solution path $\tilde \theta(t)$
generated by each method,
we use the global approximation error
$\sup_{0 \leq t \leq t_{\max}} \{ f_t(\tilde \theta(t)) - f_t(\theta(t)) \}$,
where $\tilde \theta(t)$ is the linear interpolation
of the iterates $\theta_k$ generated by each method.
To approximate the global approximation error,
we sample $N$ points $s_1, \ldots, s_N$
uniformly from $(0, t_{\max})$ and use
$\max_{1\leq i\leq N}\{ f_{s_i}(\tilde \theta({s_i})) - f_{s_i}(\theta({s_i})) \}$
as an approximation of
$\sup_{0 \leq t \leq t_{\max}} \{ f_t(\tilde \theta(t)) - f_t(\theta(t)) \}$.
Here the exact solutions $\theta({s_i})$ at $s_i$'s are
calculated using the CVX solver \citep{grant2014cvx, grant2008graph}.
In all simulations, we use $N=100$.

We first compare the four ``second-order'' methods:
Newton, Euler, Runge-Kutta, and the method of \cite{rosset2004tracking}
as they all involve solving linear systems.
In order to make a fair comparison among these four methods,
we design our experiments so that their runtime are about the same.
This can be achieved by controlling the step sizes in these methods
to ensure that they all take the same number of Newton steps.
Specifically, for any particular choice of initial step size,
we first run the proposed Newton method,
record the number of Newton steps taken (denoted as $N_{\text{Newton}}$),
and define 
$\alpha=t_{\max} / N_{\text{Newton}}$.
Then, for the Euler method and the second-order Runge-Kutta method,
we use a constant step scheme with $\alpha_k = \alpha$ and $\alpha_k = 2 \alpha$.
For the method of \cite{rosset2004tracking}, we choose the $N_{\text{Newton}}$
grid points equally spaced with $C(t) = 1/t$. This is to
ensure that all four methods have identical computational complexity.
We also consider two initial step sizes: $\alpha_1=0.01,0.1$
for the Newton method to see the impact of $\alpha_1$ on the suboptimality.

\begin{figure}[ht!]
\begin{subfigure}{0.49\textwidth}
\includegraphics[width=\linewidth]{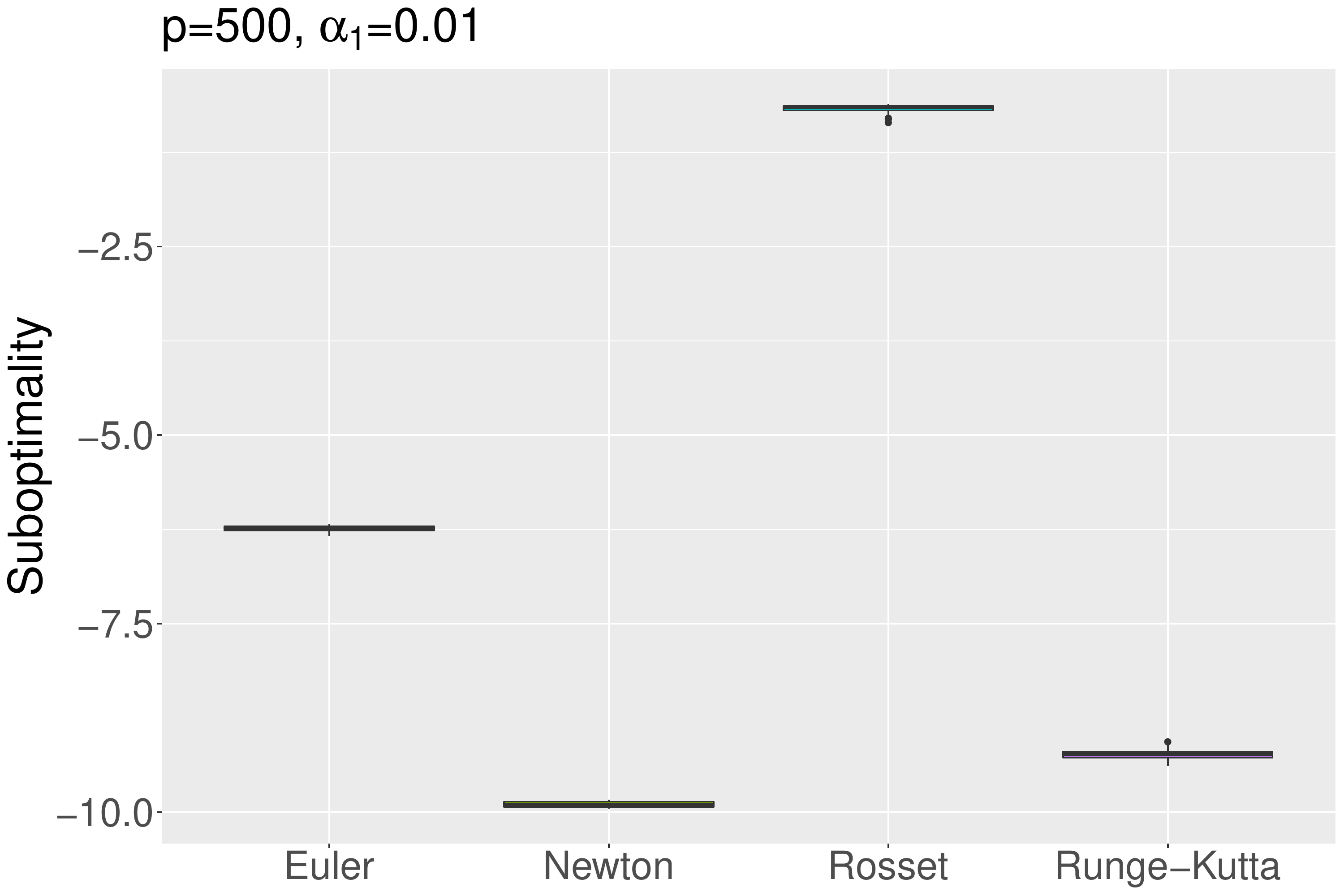}
\end{subfigure}
\hspace*{\fill}
\begin{subfigure}{0.49\textwidth}
\includegraphics[width=\linewidth]{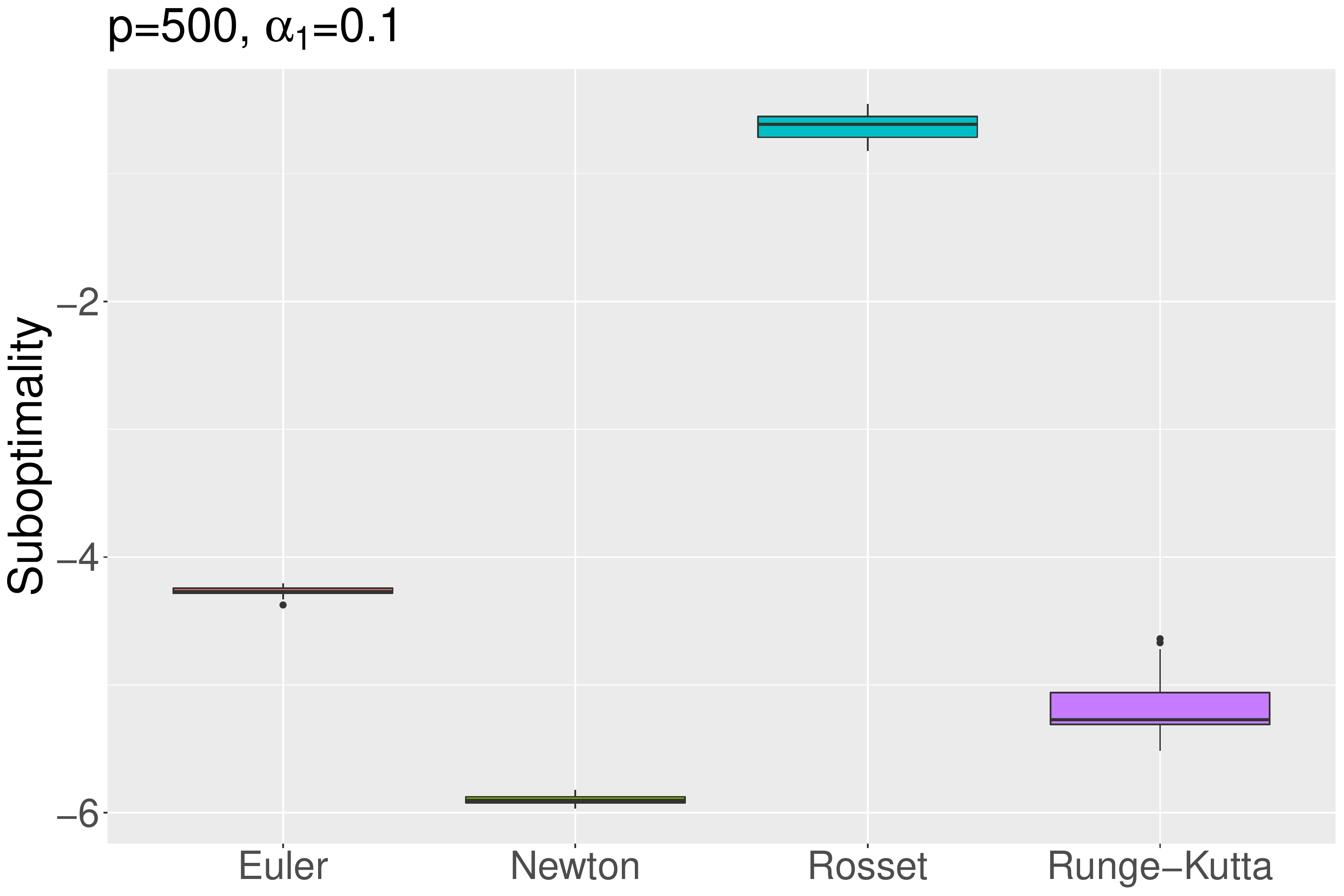}
\end{subfigure}

\medskip
\begin{subfigure}{0.49\textwidth}
\includegraphics[width=\linewidth]{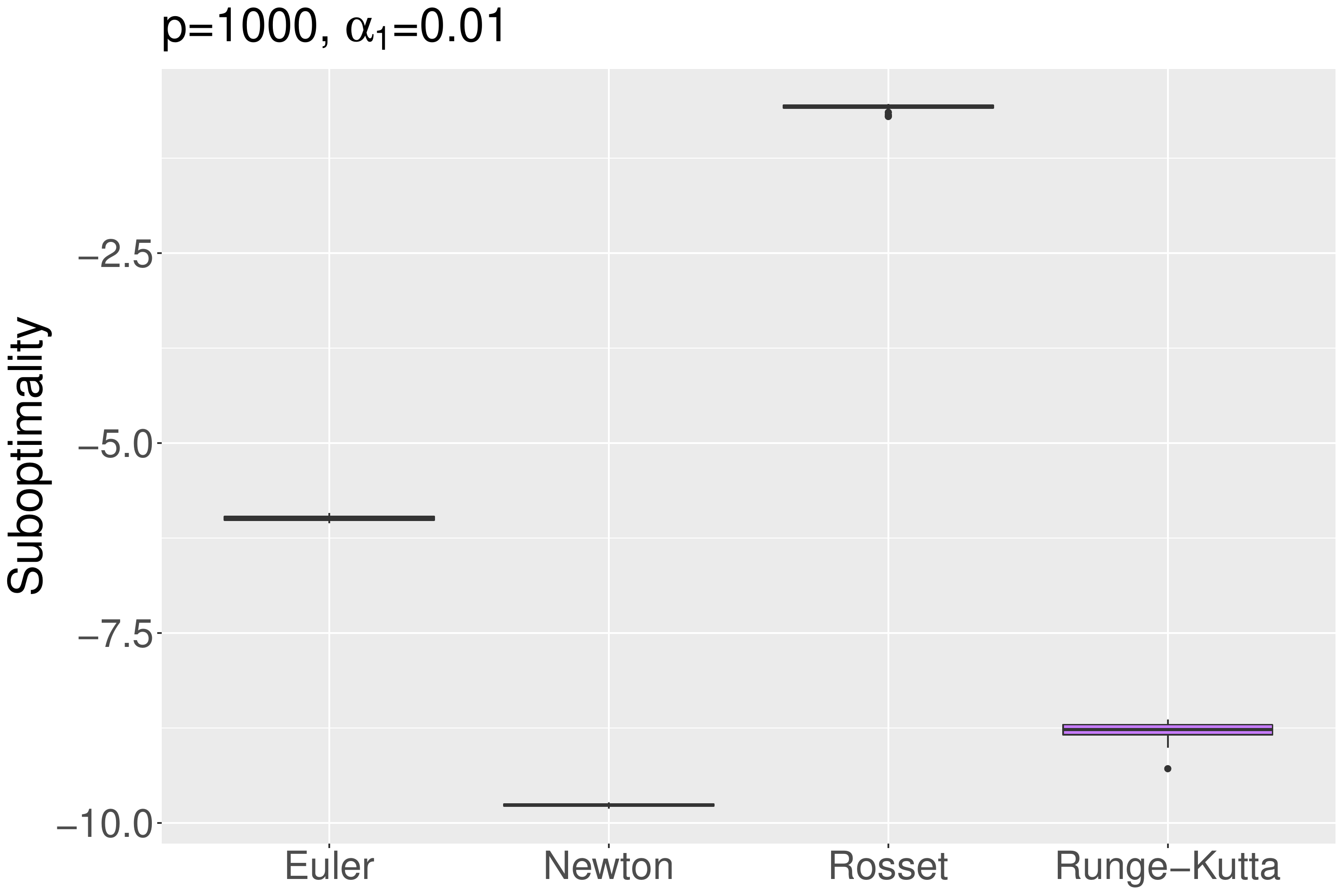}
\end{subfigure}
\hspace*{\fill}
\begin{subfigure}{0.49\textwidth}
\includegraphics[width=\linewidth]{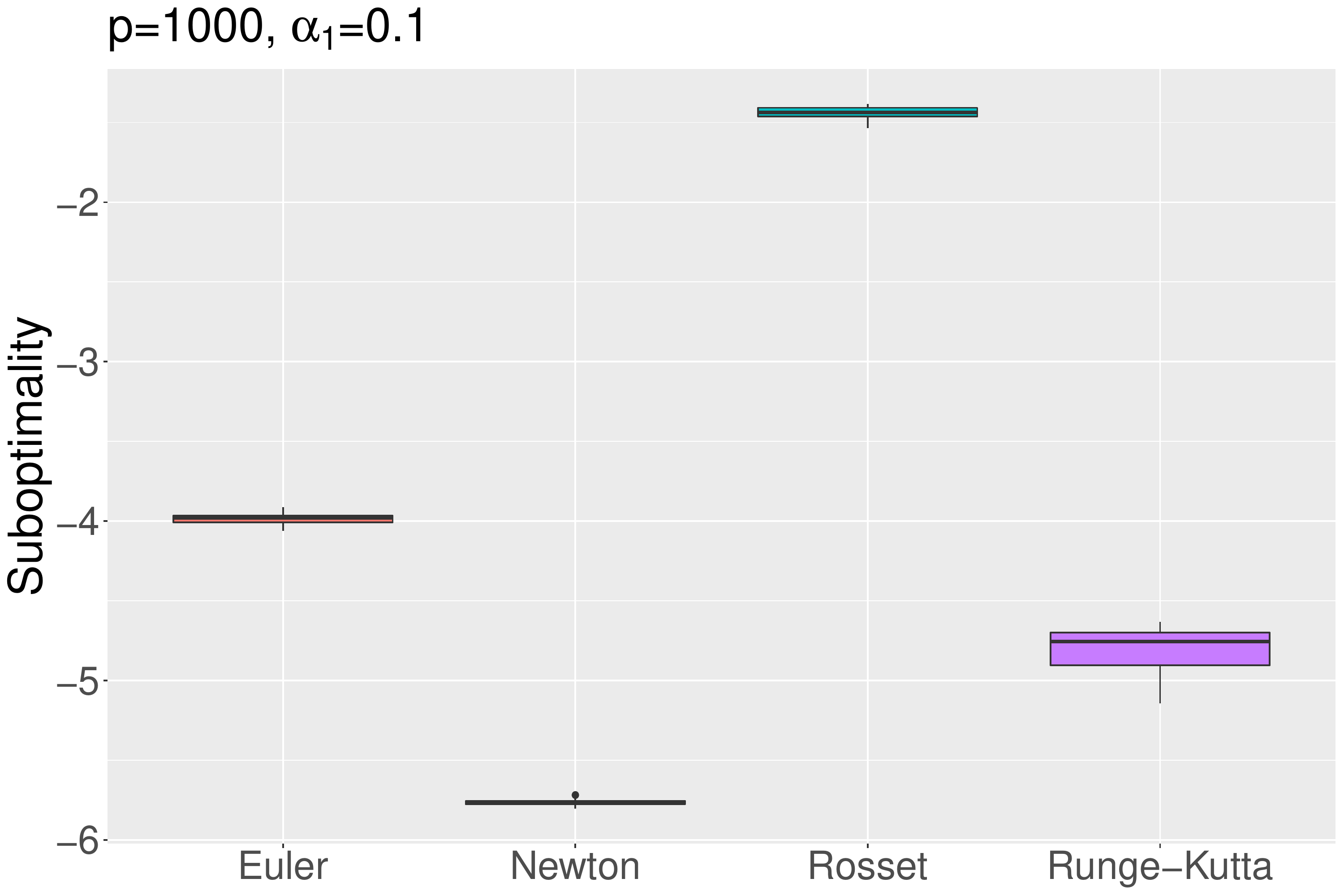}
\end{subfigure}

\medskip
\begin{subfigure}{0.49\textwidth}
\includegraphics[width=\linewidth]{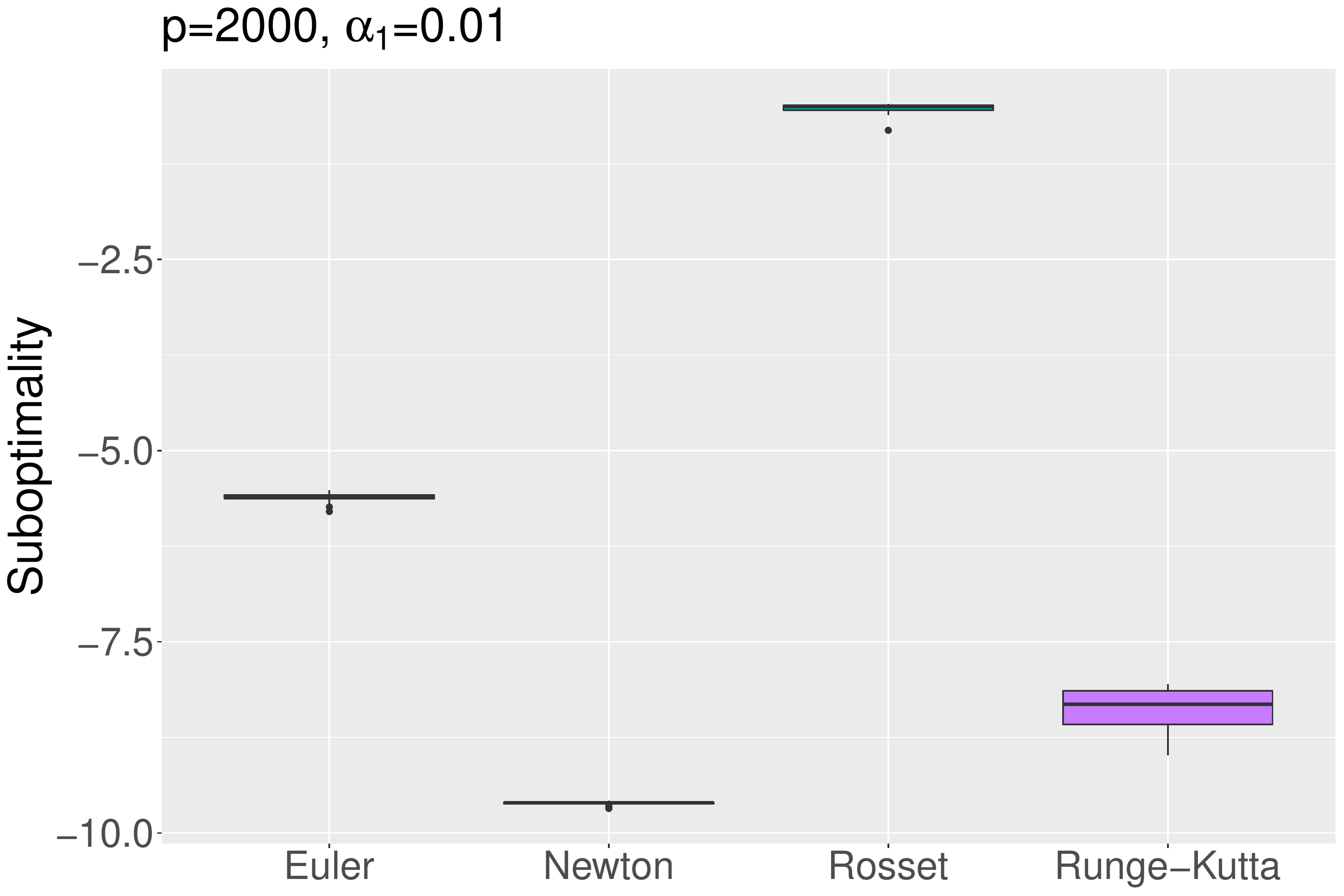}
\end{subfigure}
\hspace*{\fill}
\begin{subfigure}{0.49\textwidth}
\includegraphics[width=\linewidth]{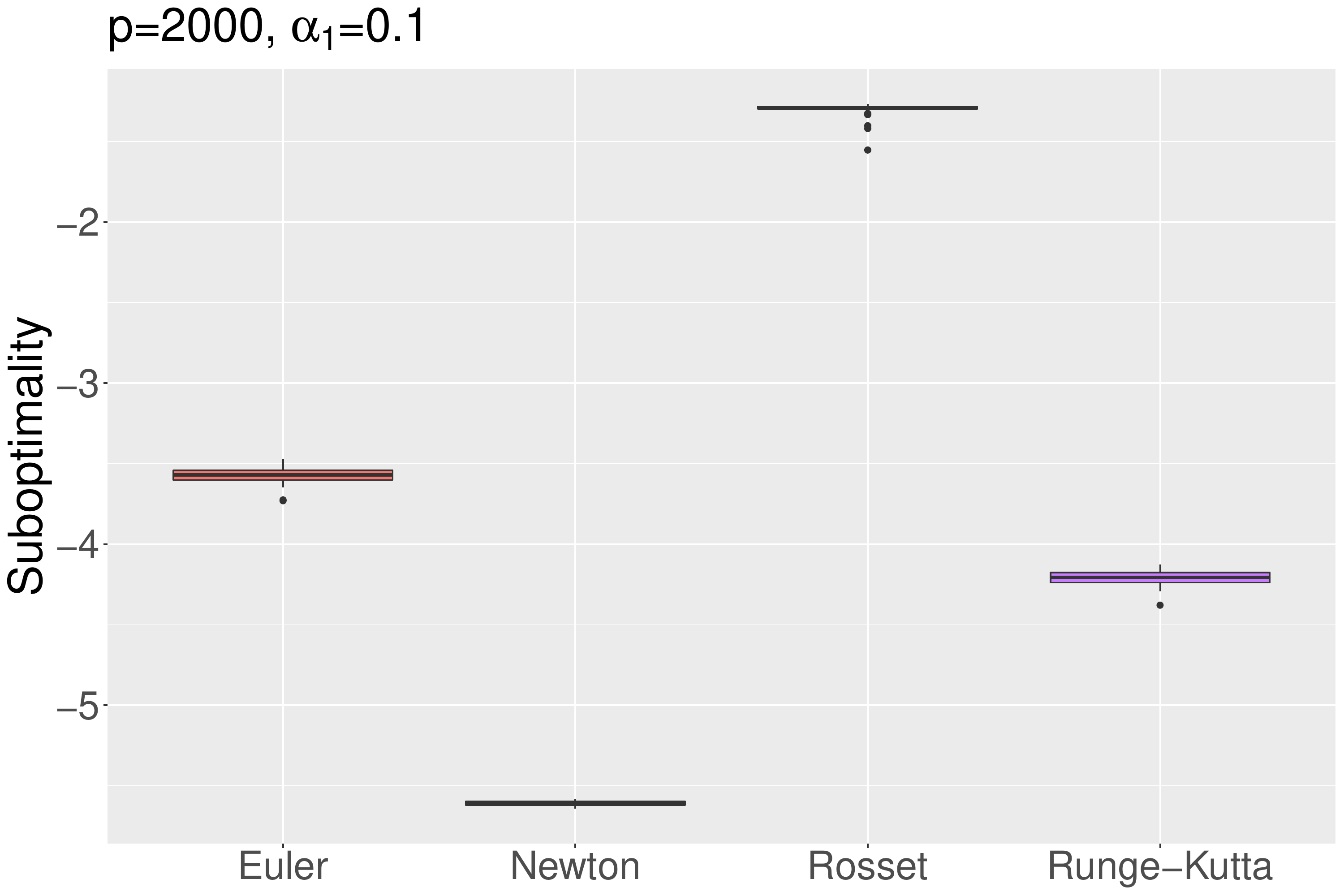}
\end{subfigure}

\caption{
Suboptimalities $\sup_{0 \leq t \leq 10} \{ f_t(\tilde \theta(t)) - f_t(\theta(t)) \}$ (in log scale)
of the approximate solution paths generated by
the proposed Newton method (Newton), the second-order Runge-Kutta method ({Runge-Kutta}),
the Euler method (Euler), and the method of \cite{rosset2004tracking} (Rosset) for
$\ell_2$-regularized logistic regression when the data is nonseparable.
}
\label{figure:nonsep_gk}
\end{figure}

\begin{figure}[ht!]
\begin{subfigure}{0.48\textwidth}
\includegraphics[width=\linewidth]{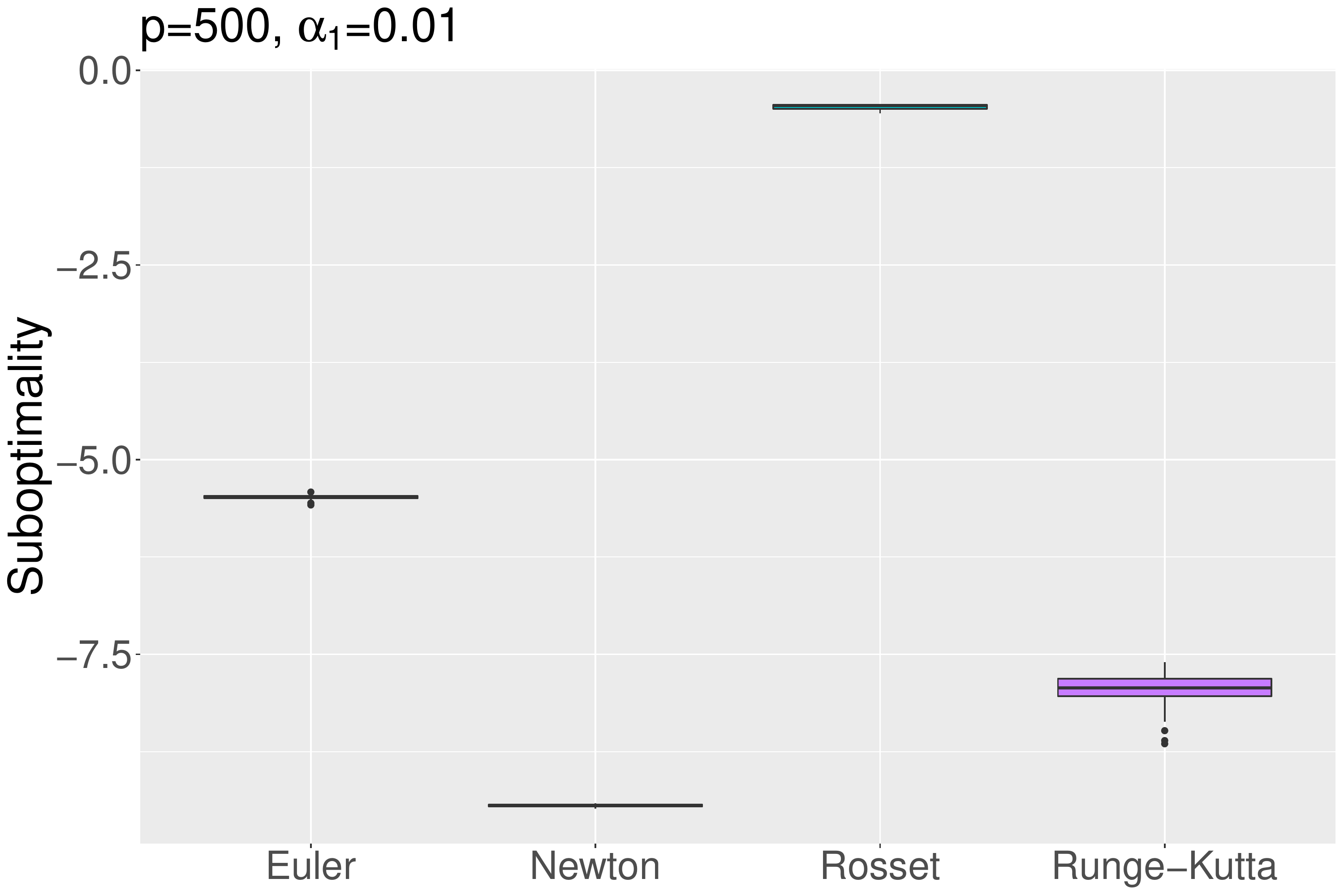}
\end{subfigure}
\hspace*{\fill}
\begin{subfigure}{0.48\textwidth}
\includegraphics[width=\linewidth]{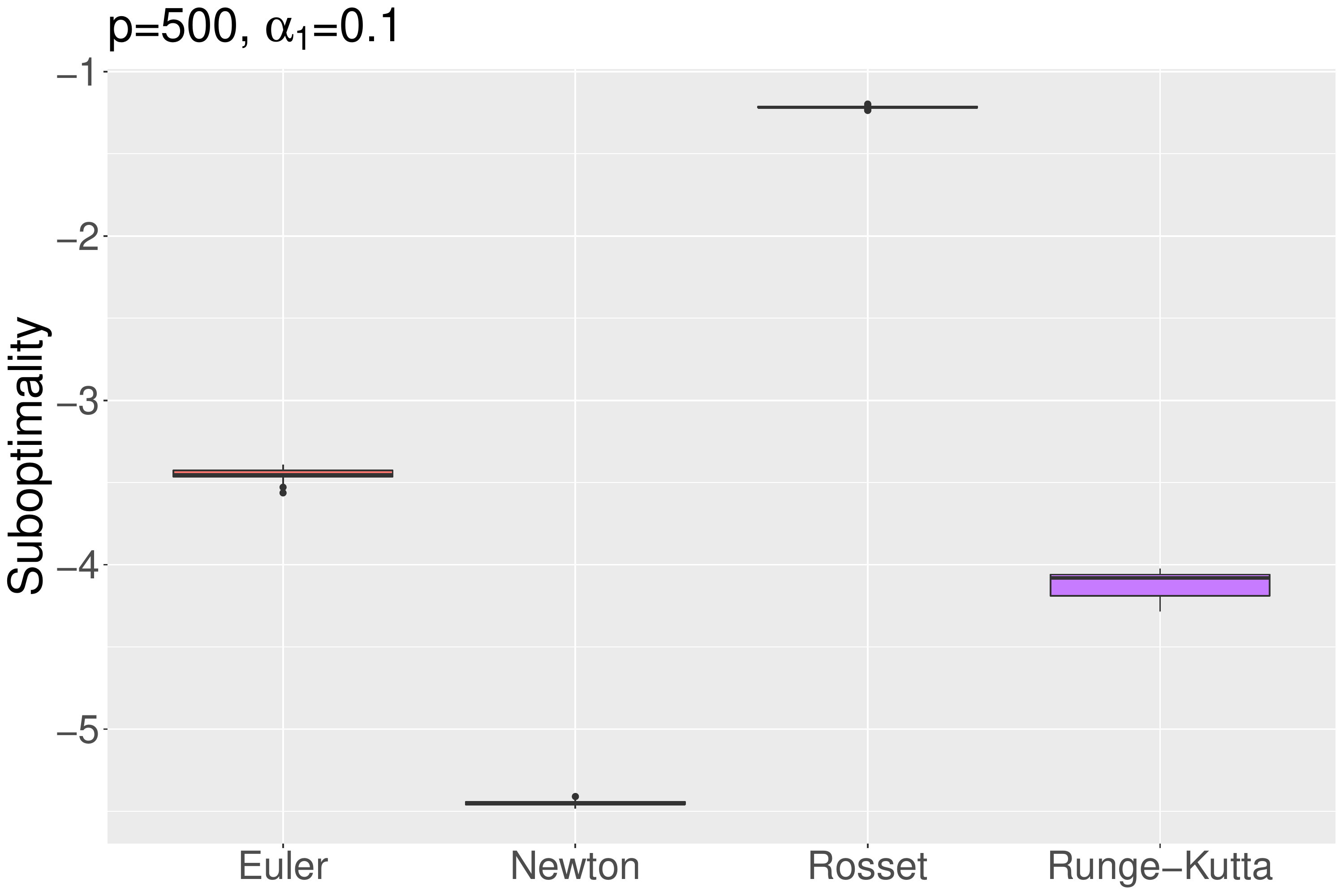}
\end{subfigure}

\medskip
\begin{subfigure}{0.48\textwidth}
\includegraphics[width=\linewidth]{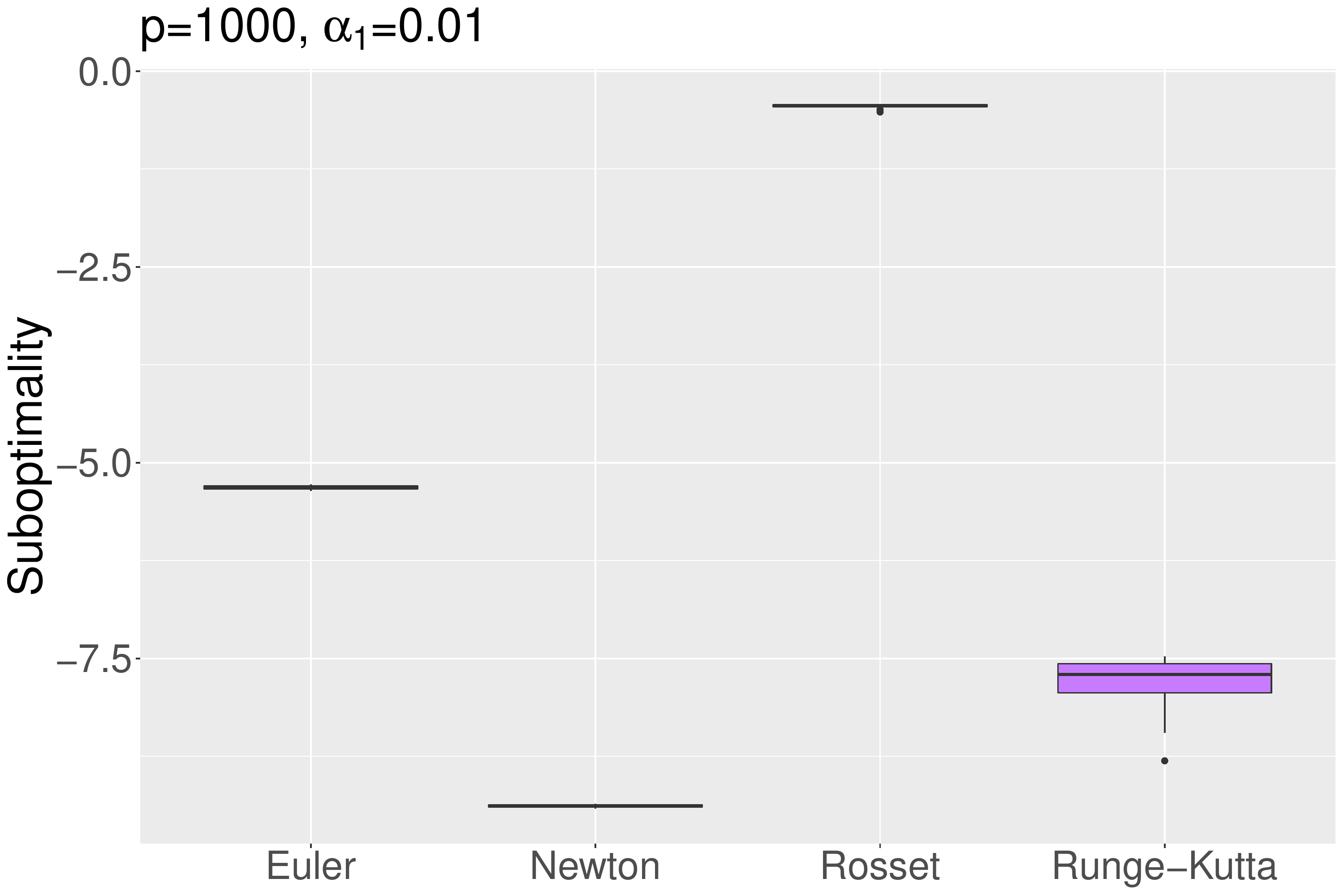}
\end{subfigure}
\hspace*{\fill}
\begin{subfigure}{0.48\textwidth}
\includegraphics[width=\linewidth]{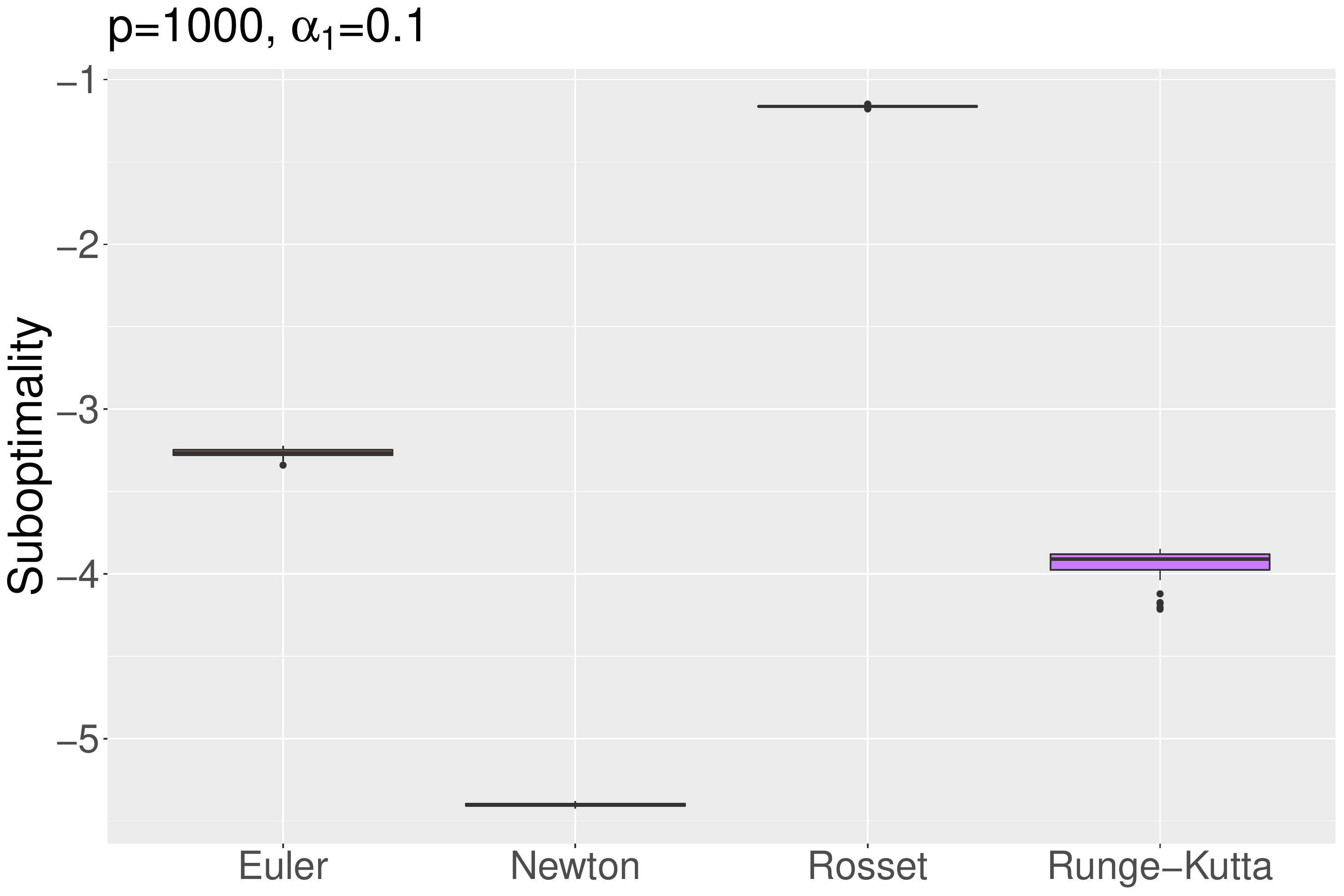}
\end{subfigure}

\medskip
\begin{subfigure}{0.48\textwidth}
\includegraphics[width=\linewidth]{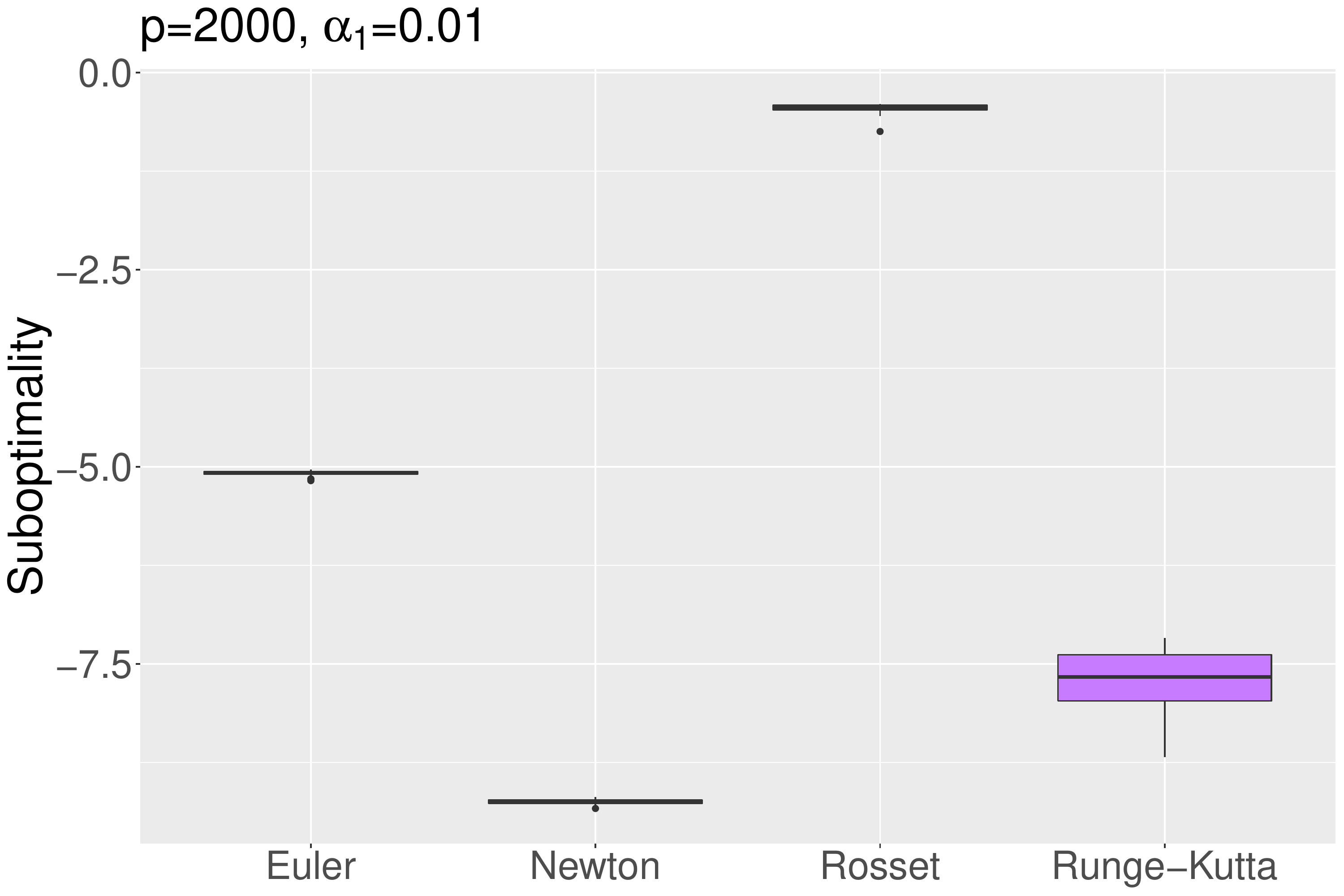}
\end{subfigure}
\hspace*{\fill}
\begin{subfigure}{0.48\textwidth}
\includegraphics[width=\linewidth]{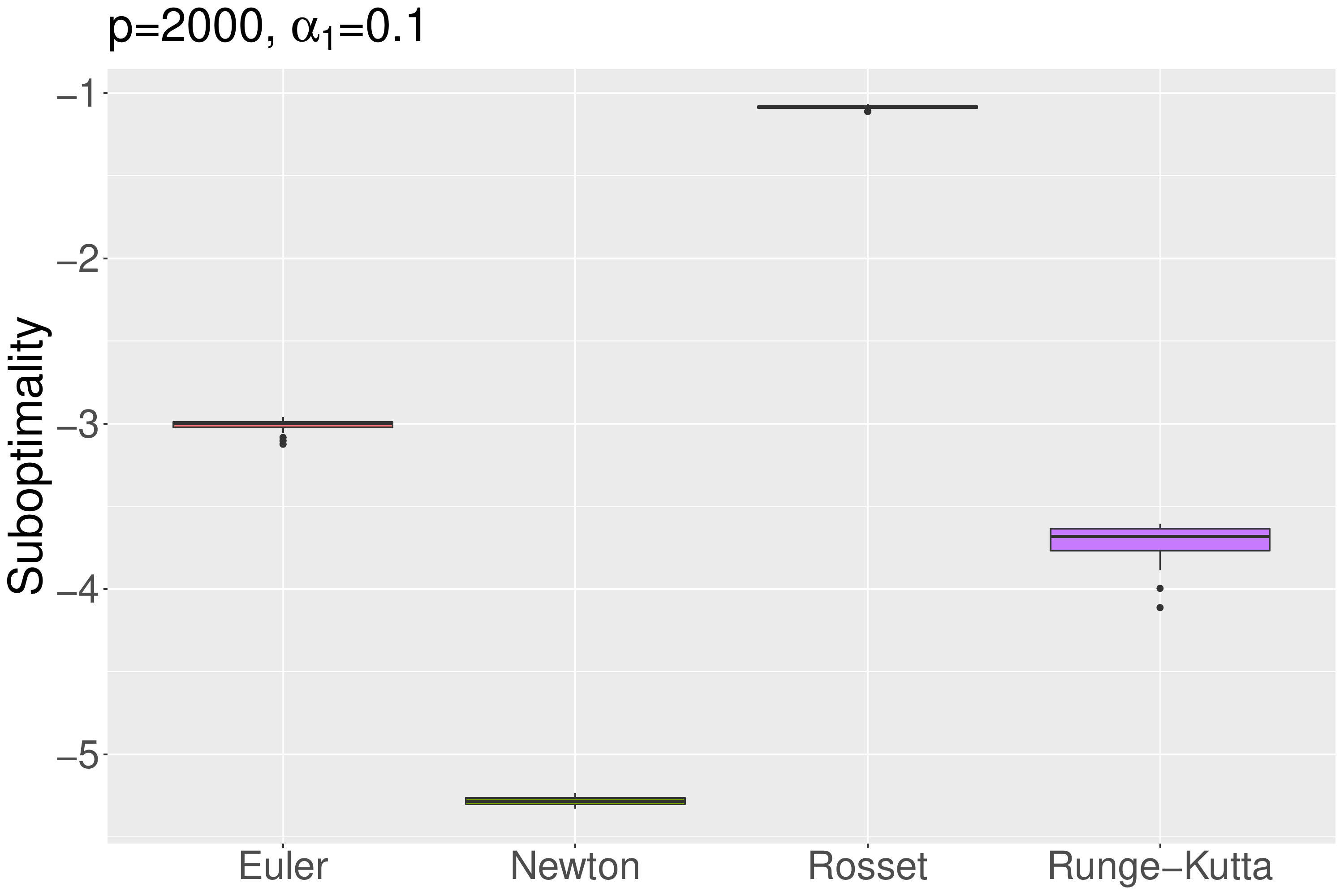}
\end{subfigure}
\caption{
Suboptimalities $\sup_{0 \leq t \leq 10} \{ f_t(\tilde \theta(t)) - f_t(\theta(t)) \}$ (in log scale)
of the approximate solution paths generated by
the proposed Newton method (Newton), the second-order Runge-Kutta method (Runge-Kutta),
the Euler method (Euler), and the method of \cite{rosset2004tracking} (Rosset) for
$\ell_2$-regularized logistic regression when the data is separable.
}
\label{figure:sep_gk}
\end{figure}

\begin{figure}[ht!]
\begin{subfigure}{0.48\textwidth}
\includegraphics[width=\linewidth]{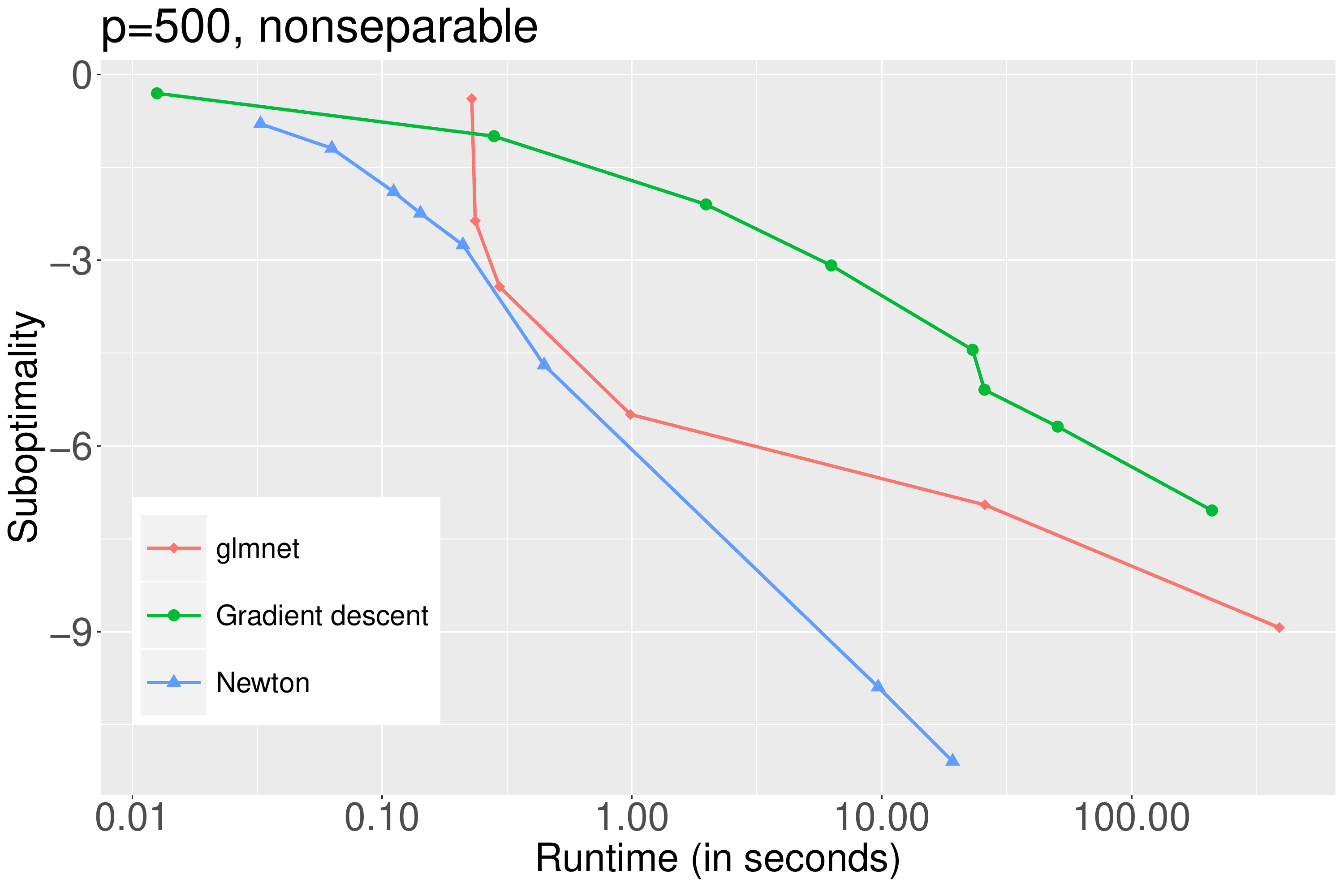}
\end{subfigure}
\hspace*{\fill}
\begin{subfigure}{0.48\textwidth}
\includegraphics[width=\linewidth]{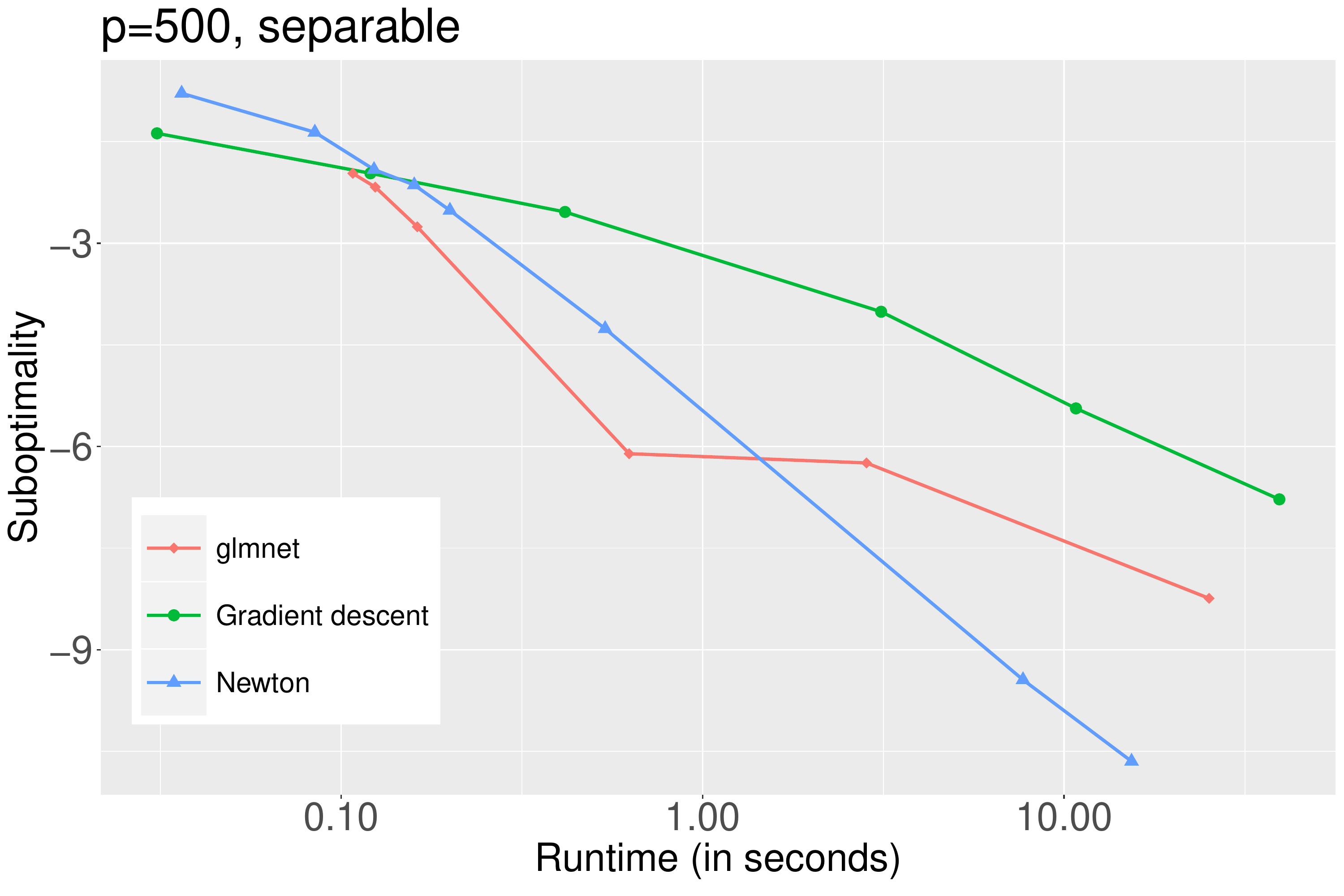}
\end{subfigure}

\medskip
\begin{subfigure}{0.48\textwidth}
\includegraphics[width=\linewidth]{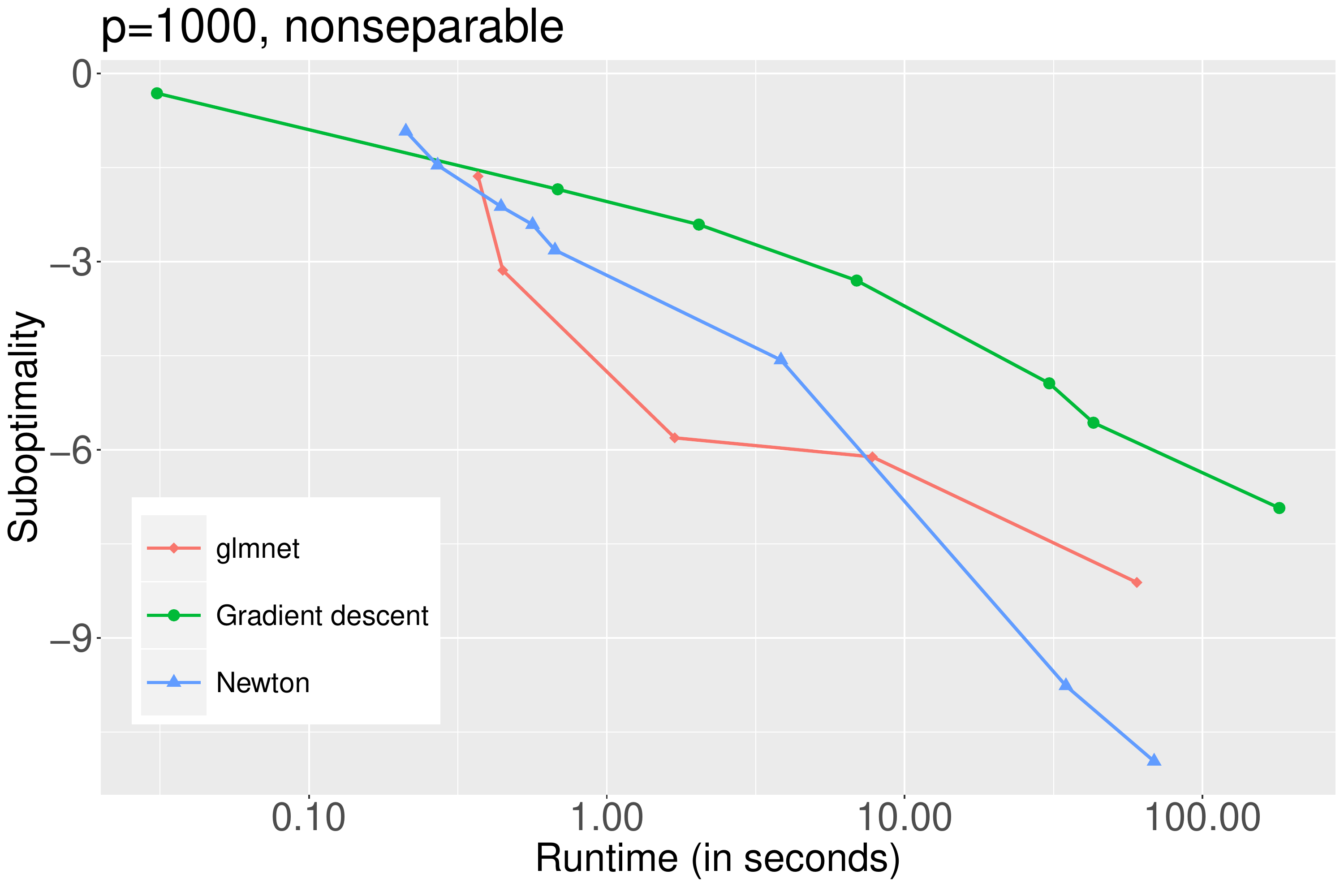}
\end{subfigure}
\hspace*{\fill}
\begin{subfigure}{0.48\textwidth}
\includegraphics[width=\linewidth]{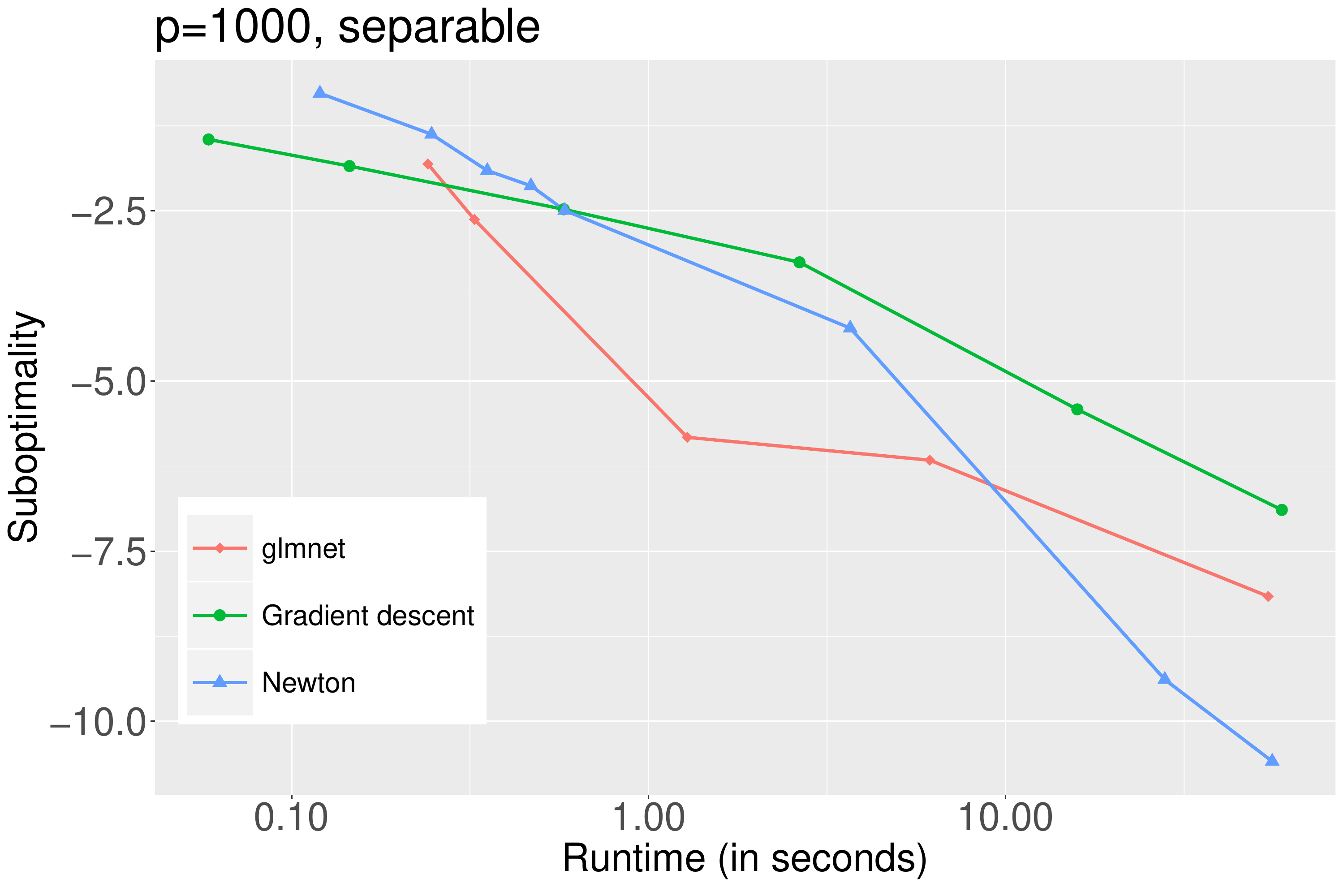}
\end{subfigure}

\medskip
\begin{subfigure}{0.48\textwidth}
\includegraphics[width=\linewidth]{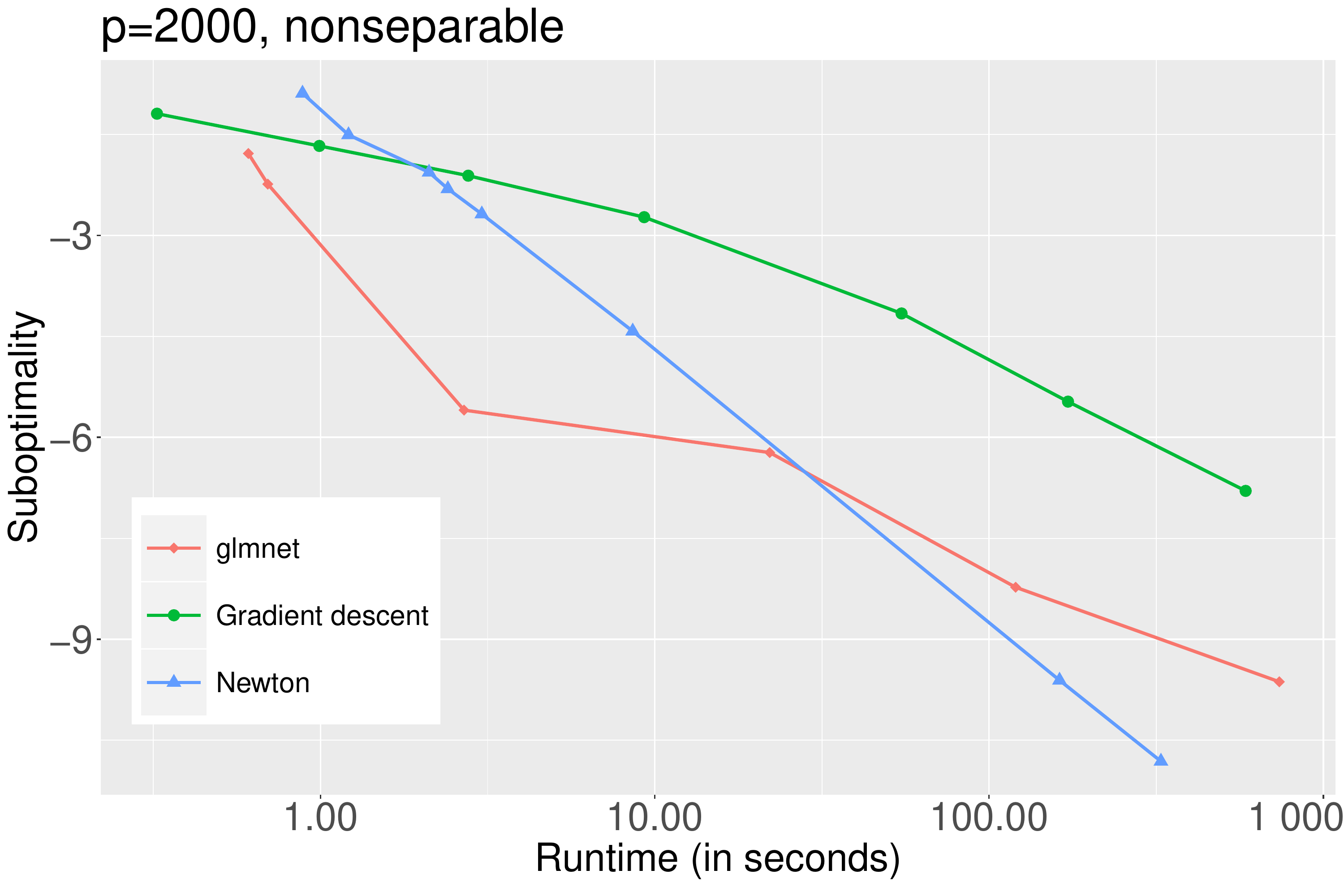}
\end{subfigure}
\hspace*{\fill}
\begin{subfigure}{0.48\textwidth}
\includegraphics[width=\linewidth]{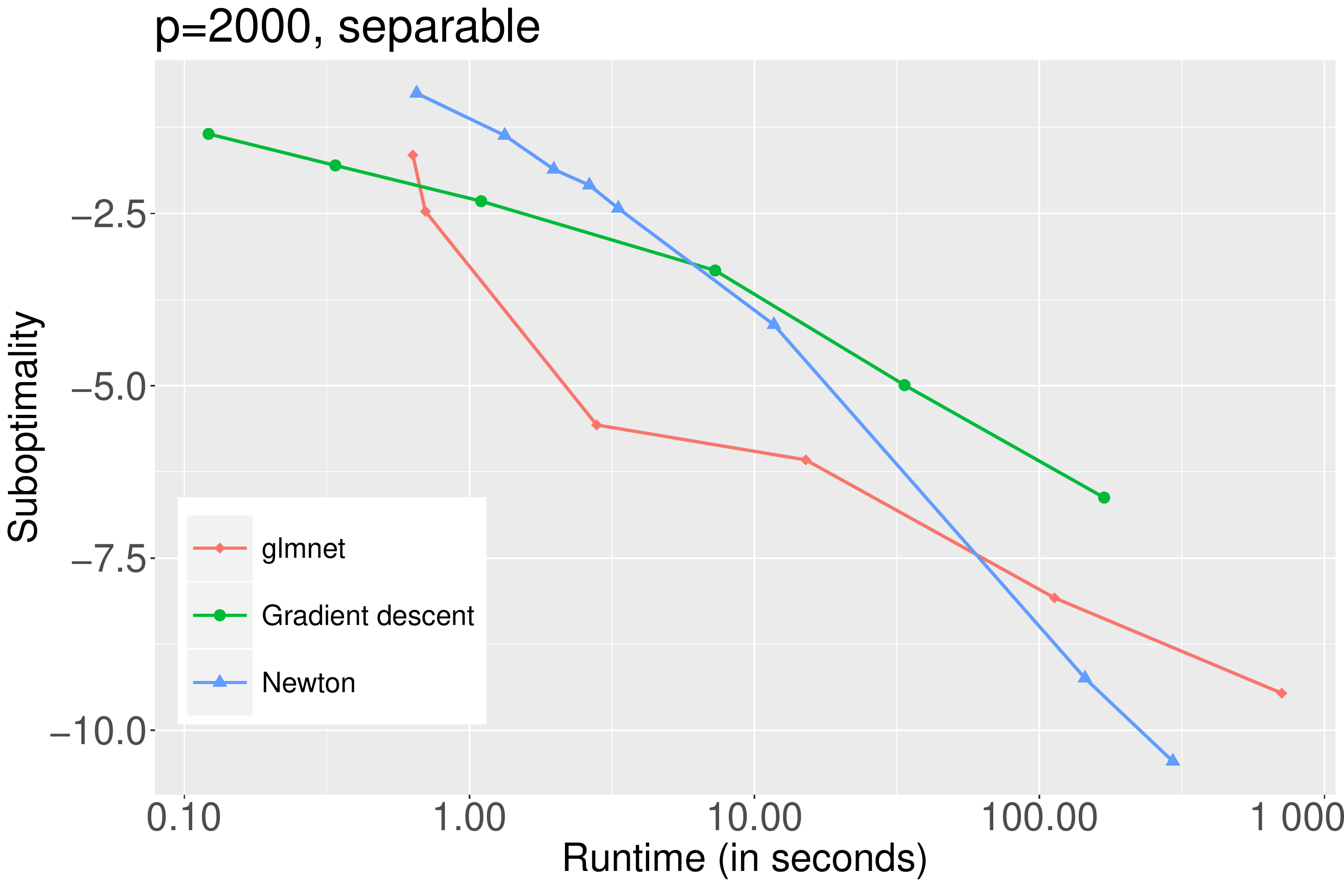}
\end{subfigure}
\caption{Runtime v.s. suboptimality
for the proposed Newton method, gradient descent method, and glmnet under six different scenarios,
when applied to $\ell_2$-regularized logistic regression.
}
\label{figure:time_nonsep}
\end{figure}

Figure \ref{figure:nonsep_gk} and \ref{figure:sep_gk} present the
global approximation errors (on a log scale) of the aforementioned four
second-order methods
for nonseparable and separable cases, respectively.
Note that two initial step sizes $\alpha_1 = .01, .01$ are
used for the proposed Newton method, and the other methods use
the corresponding initial step sizes so that the overall computations are
comparable to that of the Newton method.
Among the four methods, the proposed Newton method performs the best, followed by the
second-order Runge-Kutta method, the Euler method,
 and the method of \cite{rosset2004tracking}.
The method of \cite{rosset2004tracking} is much worse compared to other methods
due to the way it chooses the grid points.


Next we compare the Newton method and gradient descent method against
glmnet in terms of both runtime and approximation error.
In this case, it is difficult to control the initial step sizes so that they
have similar runtime. As such, we choose to look at the trade-off
curve of runtime and
approximation error for these three methods.
Figure \ref{figure:time_nonsep} presents plots of runtime versus
approximation error based on $100$ simulations,
as we vary the initial step size for each method.
We can see from Figure \ref{figure:time_nonsep} that
the proposed Newton method runs the fastest when the desired
suboptimality is small (high precision), especially when the problem dimension is small.
Also, as expected, the gradient method runs the
slowest when the desired suboptimality is small.
Interestingly, the glmnet performs better than the gradient descent method
in most cases, but worse than the Newton method when the desired suboptimality is small.
This could be partially explained by the fact that the coordinate descent algorithms
can usually be viewed as a type of methods that is between ``first-order'' and
``second-order'' method.

In summary, in terms of approximation error and computational efficiency,
the Newton method and
the second-order Runge-Kutta method both work quite well when the
problem dimension is not too large or the desired suboptimality is small.
For large-scale problems, however, gradient descent method and glmnet
seem to be more scalable, although
glmnet produces solution paths with better suboptimality.

Lastly, we investigate how the initial step size of various solution path
algorithms would affect their statistical performances. As we have argued before,
the initial step size determines the approximation error.
To assess the accuracy of the approximation to the true statistical risk,
we consider a generative model
for logistic regression.
Specifically, we first generate
the predictors $X_1,\ldots, X_n\in \mathbb R^p$ from normal distribution
$N_p( 0,  I_{p \times p})$.
Given predictor $X_i$,
we draw the binary response $Y_i \in \{-1, +1\}$ from
Bernoulli distributions
with {$\mathbb P_{\theta}(Y_i=+1)=\exp(X_i^\top \theta) / (1+\exp(X_i^\top \theta))$}
for $i=1,2\ldots, n$,
where the true regression coefficient $\theta \in \mathbb R^p$ is drawn from
$N_p( 0, (16/p) \cdot  I_{p \times p})$.
Three choices of problem dimensions
 $(n,p)=(500, 100)$, $(n,p)=(500,500)$, and $(n,p)=(500, 1000)$ will be considered.
The statistical risk of an approximate solution path
$\tilde \theta(t)$ is quantified by the Kullback–Leibler divergence:
\begin{figure}[ht!]
\begin{subfigure}{0.48\textwidth}
\includegraphics[width=\linewidth]{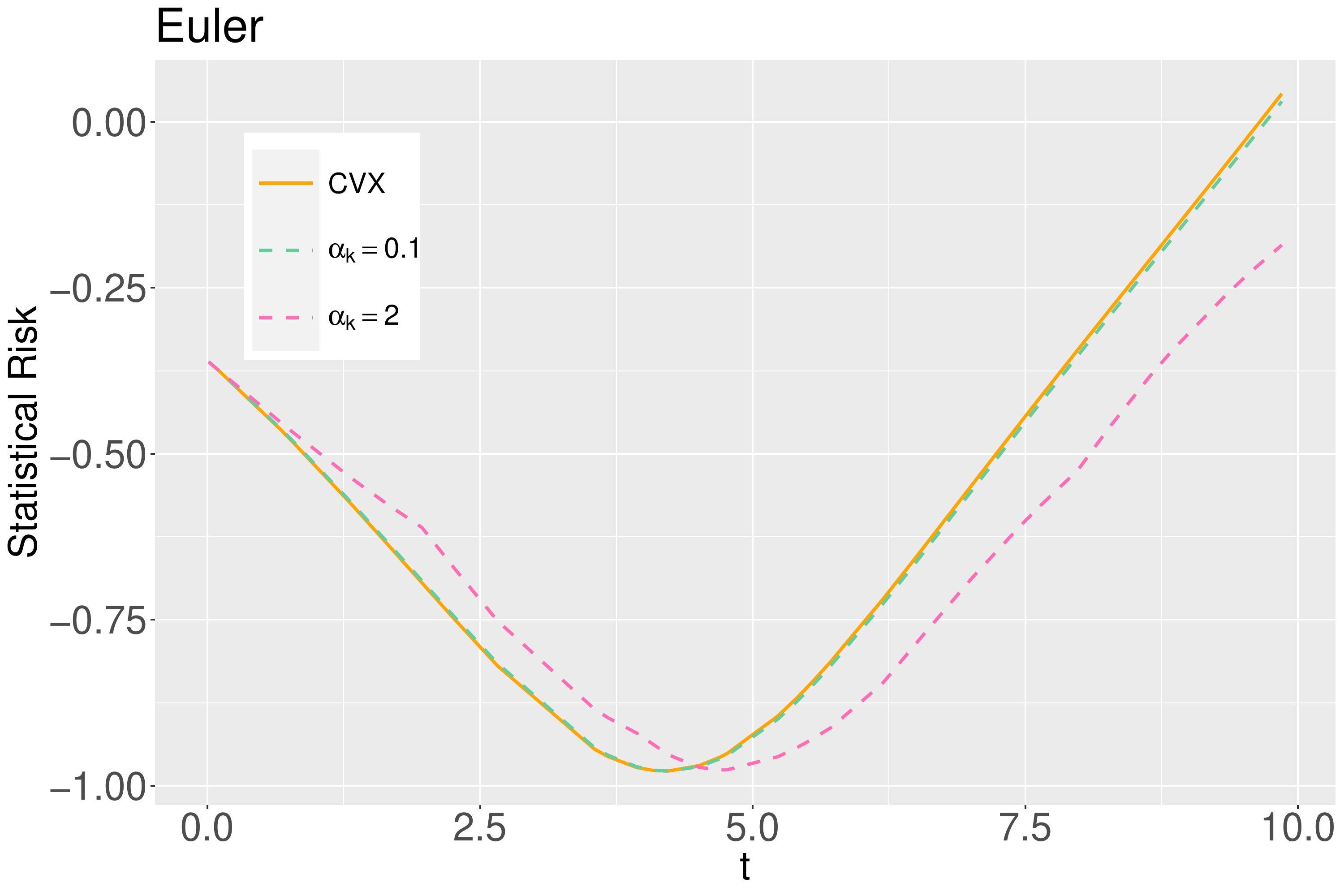}
\end{subfigure}
\hspace*{\fill}
\begin{subfigure}{0.48\textwidth}
\includegraphics[width=\linewidth]{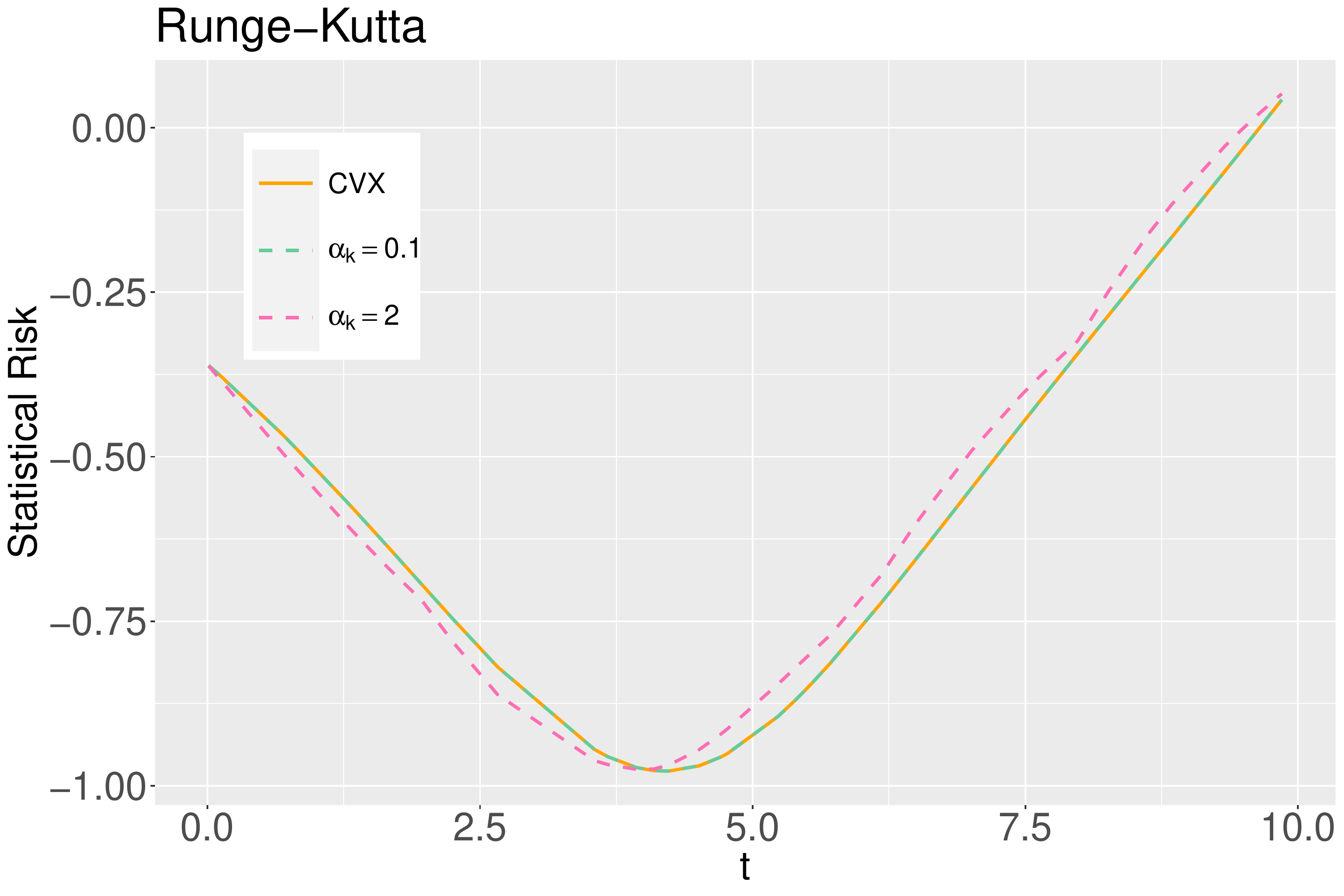}
\end{subfigure}


\medskip
\begin{subfigure}{0.48\textwidth}
\includegraphics[width=\linewidth]{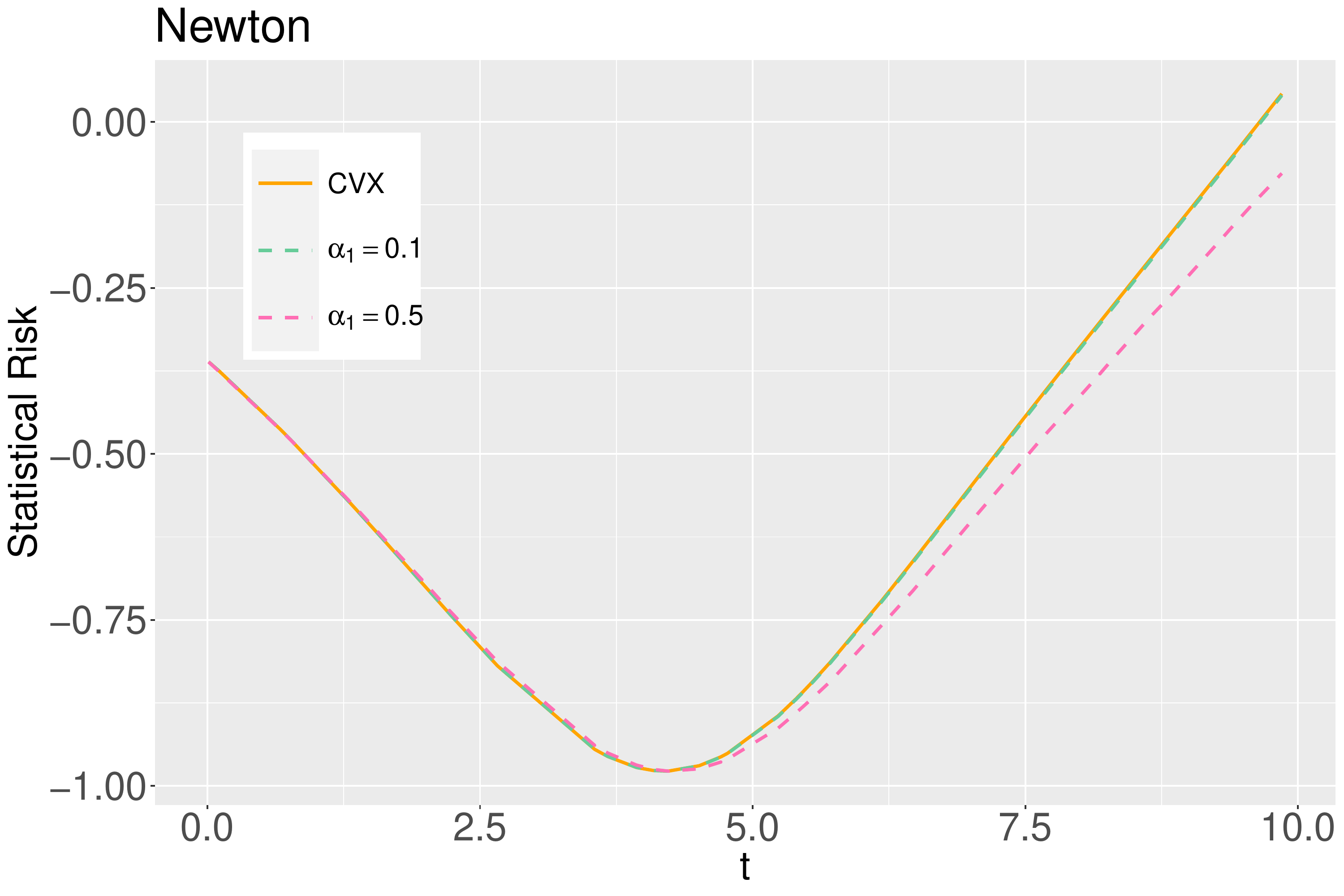}
\end{subfigure}
\hspace*{\fill}
\begin{subfigure}{0.48\textwidth}
\includegraphics[width=\linewidth]{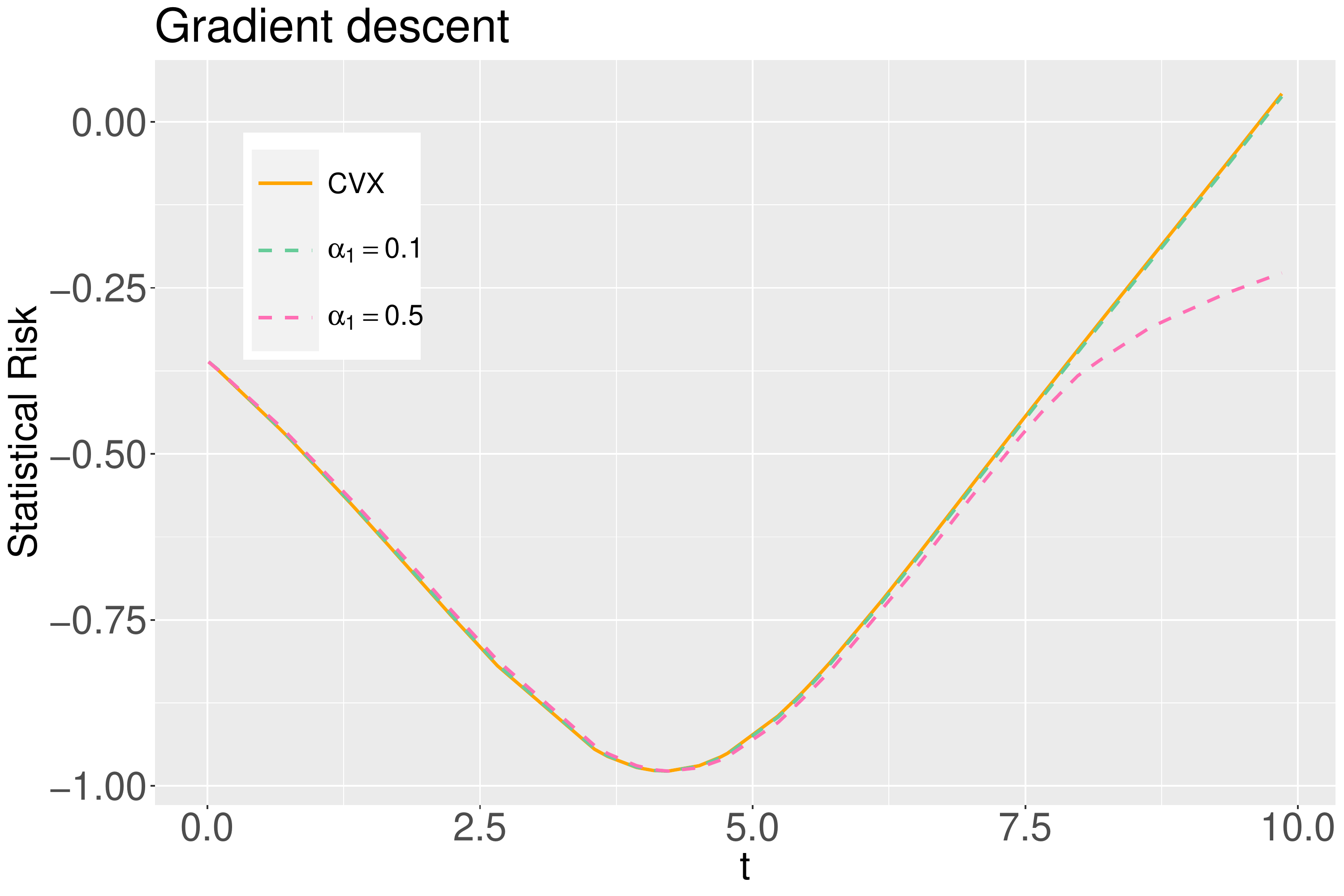}
\end{subfigure}
\caption{
Approximate risk curve $\log_{10}(R(\tilde \theta(t), \theta))$ of the proposed
algorithms applied to
$\ell_2$-regularized logistic regression when problem dimension is $(n,p)=(500,100)$.
The CVX (orange) curve denotes the true risk curve
$\log_{10}(R(\theta(t), \theta))$ with
$\theta(t)$ computed using the CVX solver.
For algorithms with constant step size (Euler and Runge-Kutta),
 $\alpha_k$ denotes the step size;
while $\alpha_1$ denotes the initial step size for Newton and gradient descent method.
}
\label{figure:risk_no_GD_low_dimension}
\end{figure}
\begin{equation*}
R(\tilde \theta(t), \theta)=
\mathbb E_{\theta} \log(1+\exp(-Y X^\top \tilde \theta(t)))-
\mathbb E_{\theta} \log(1+\exp(-Y X^\top \theta))\, .
\end{equation*}
Note that the statistical risk for the exact solution path $\theta(t)$ is
$R(\theta(t),\theta)$, which we refer to as the true risk curve (as a function of $t$). Here,
we calculate the exact solution path $\theta(t)$
 using CVX \citep{grant2014cvx, grant2008graph}.
Again, the goal is to see the impact of the initial step size on how close the approximate
risk curve $R(\tilde \theta(t),\theta)$ is to the true risk curve $R(\theta(t),\theta)$.

\begin{figure}[ht!]
\begin{subfigure}{0.48\textwidth}
\includegraphics[width=\linewidth]{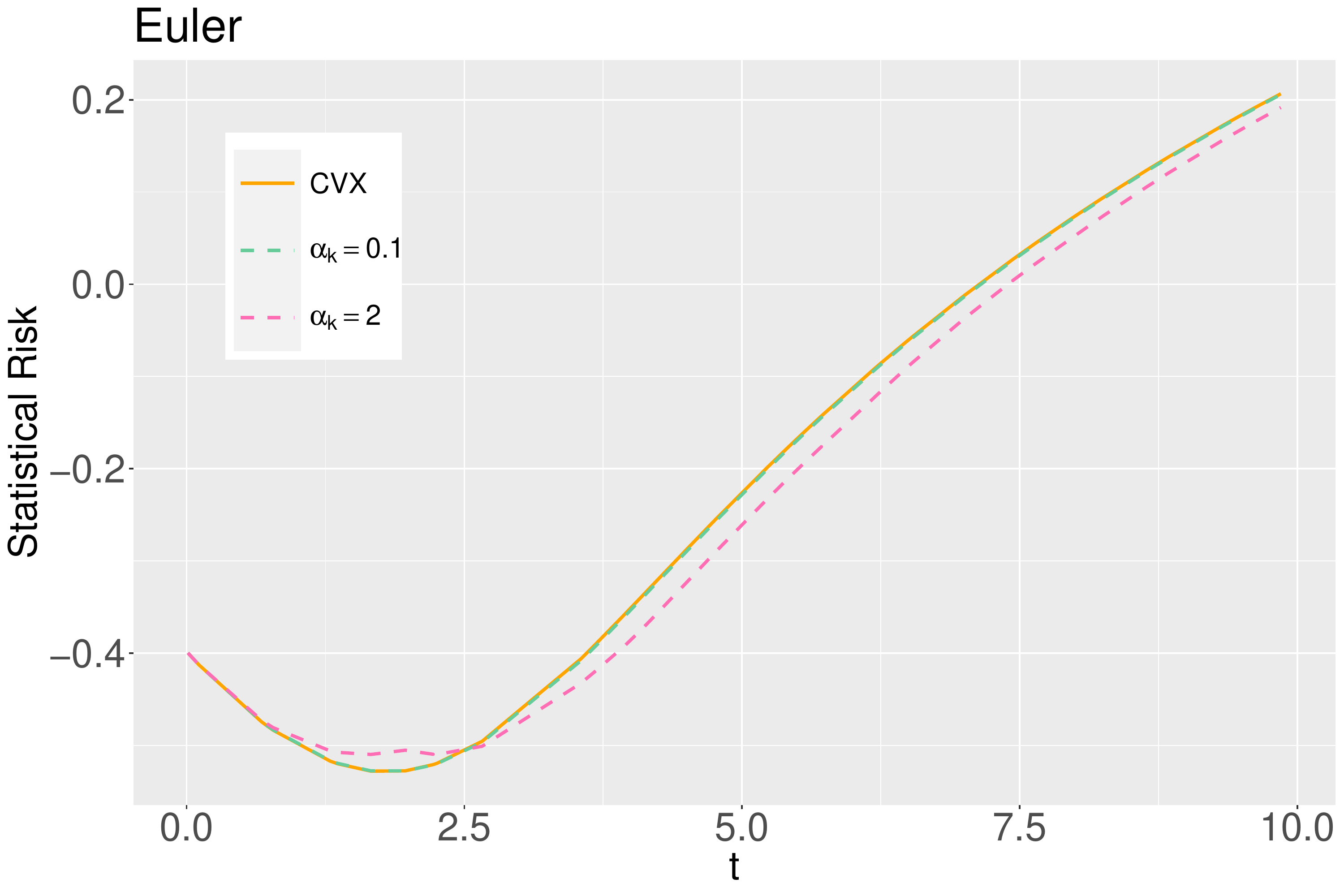}
\end{subfigure}
\hspace*{\fill}
\begin{subfigure}{0.48\textwidth}
\includegraphics[width=\linewidth]{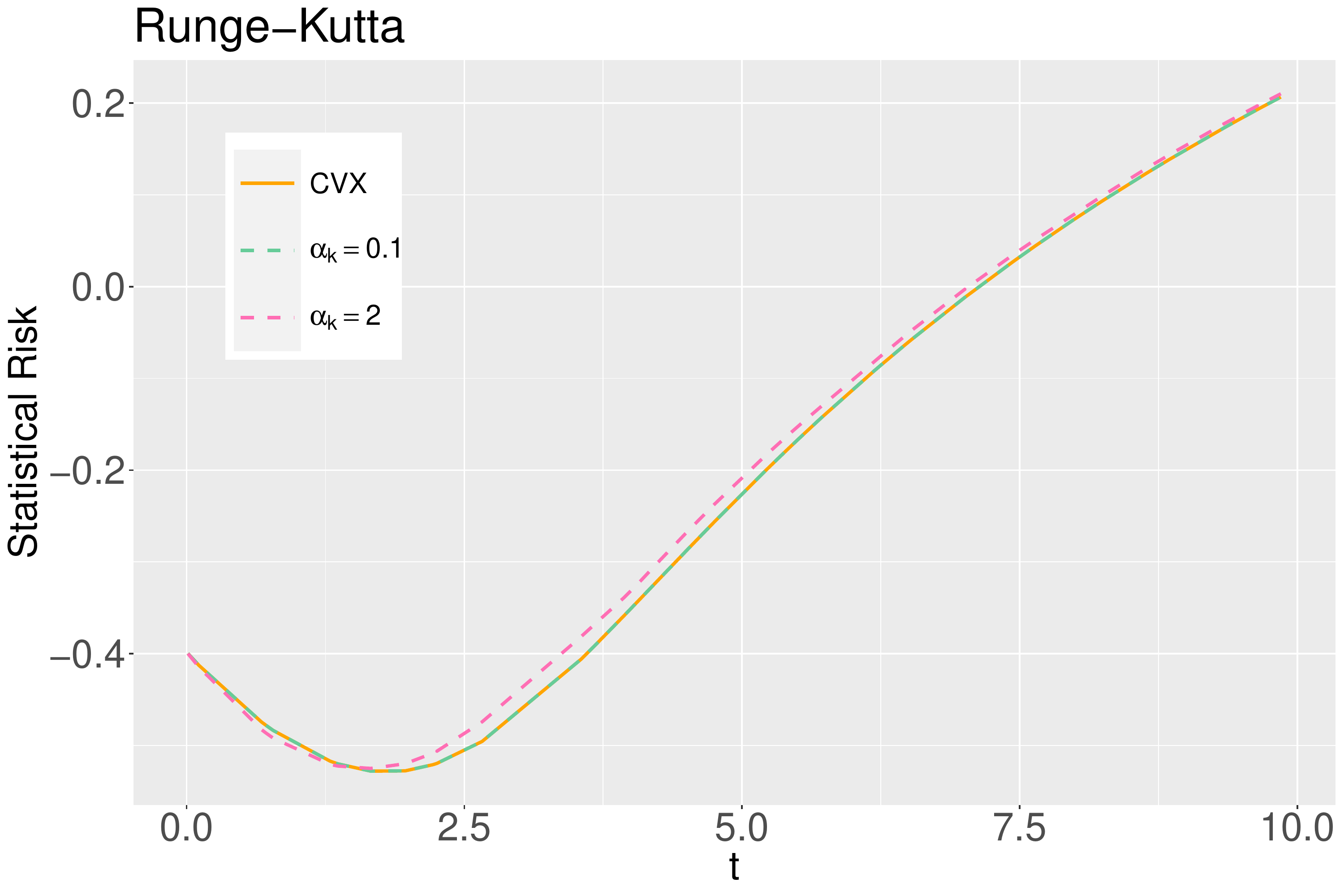}
\end{subfigure}

\medskip
\begin{subfigure}{0.48\textwidth}
\includegraphics[width=\linewidth]{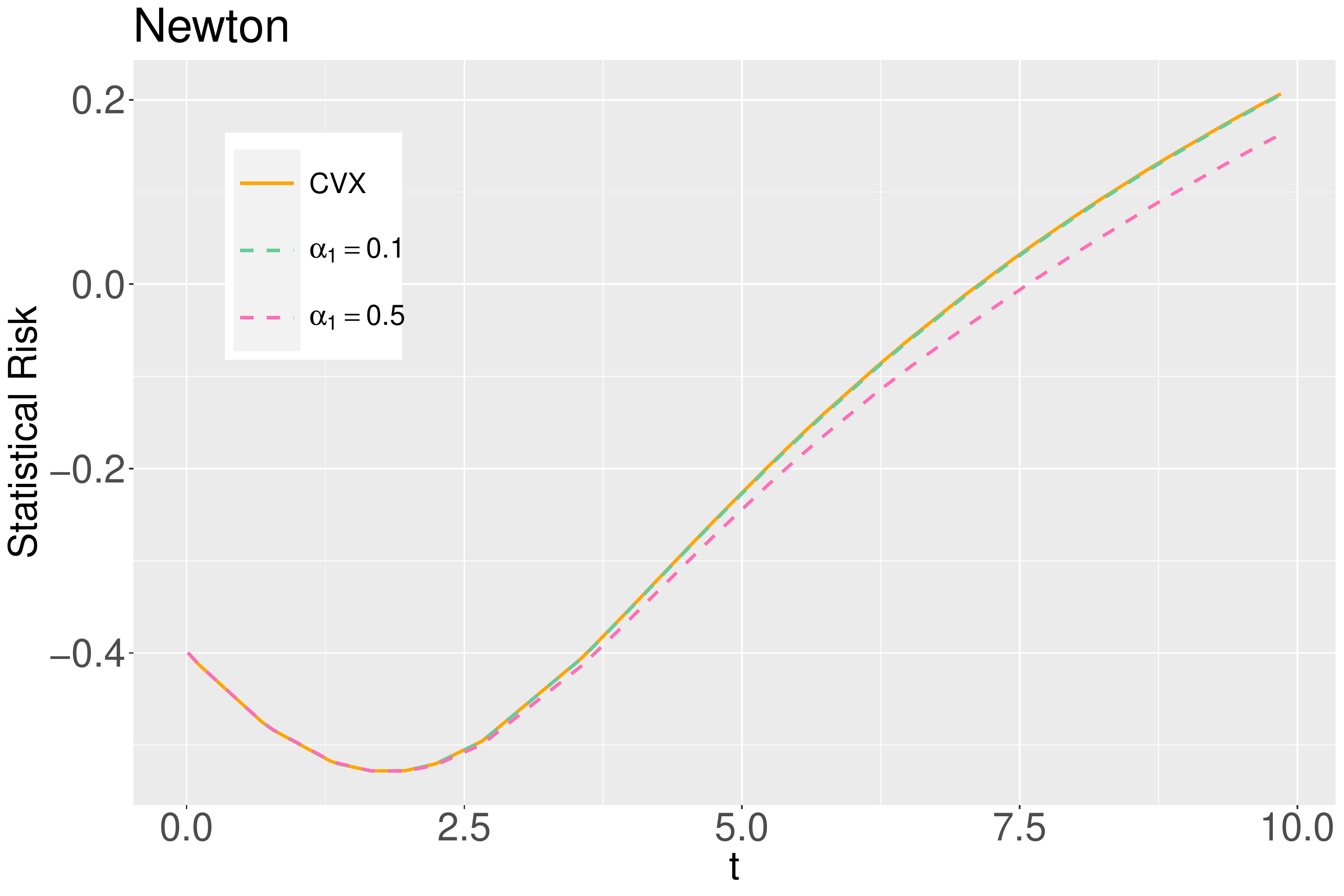}
\end{subfigure}
\hspace*{\fill}
\begin{subfigure}{0.48\textwidth}
\includegraphics[width=\linewidth]{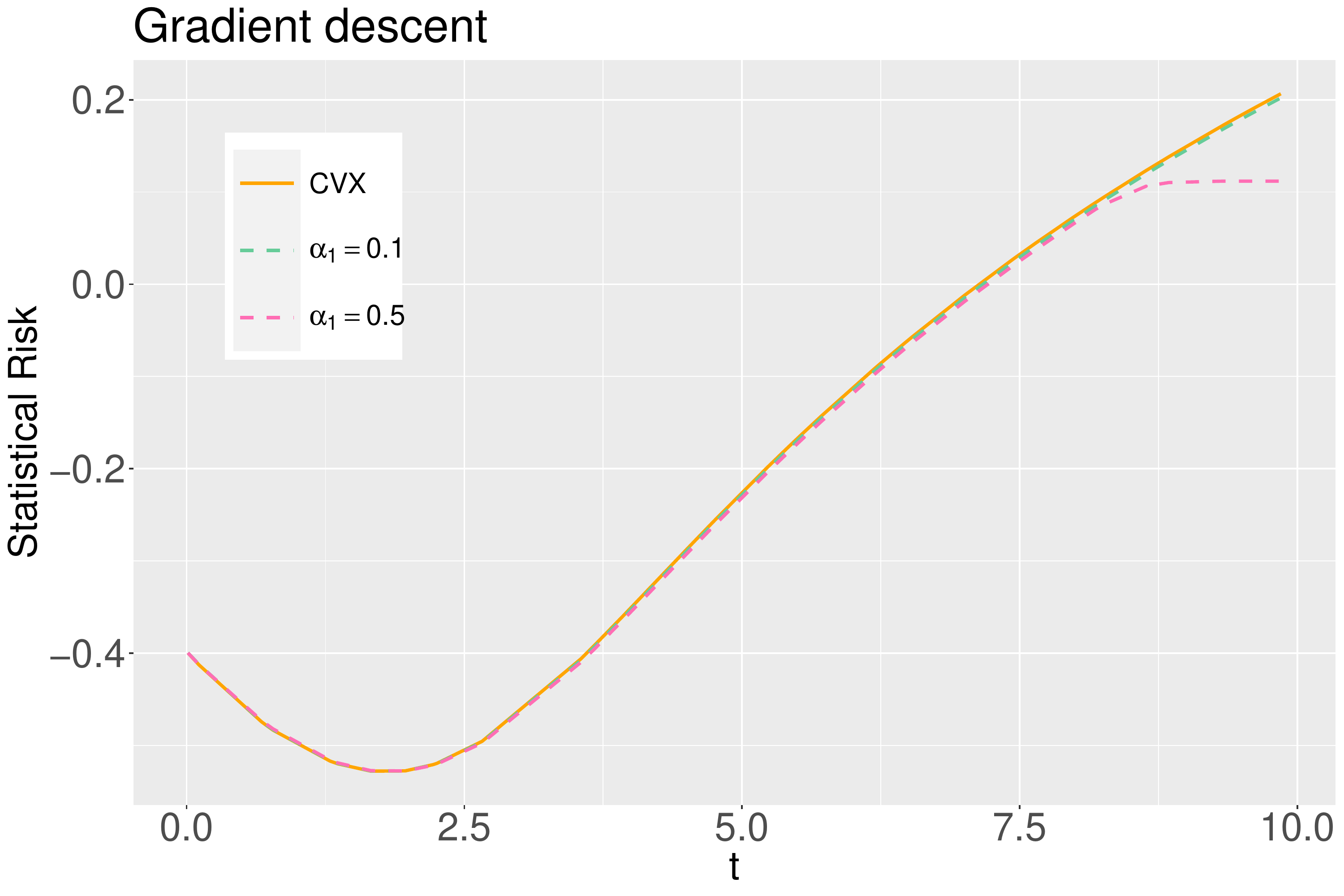}
\end{subfigure}
\caption{
Approximate risk curve
{$\log_{10}(R(\tilde \theta(t), \theta))$} of the proposed
algorithms applied to
$\ell_2$-regularized logistic regression when problem dimension is $(n,p)=(500,500)$.
The CVX (orange) curve denotes the true risk curve
{$\log_{10}(R(\theta(t), \theta))$} with
$\theta(t)$ computed using the CVX solver.
For algorithms with constant step size (Euler and Runge-Kutta),
 $\alpha_k$ denotes the step size;
while $\alpha_1$ denotes the initial step size for Newton and gradient descent method.
}
\label{figure:risk_no_GD_mid_dimension}
\end{figure}

\begin{figure}[ht!]
\begin{subfigure}{0.48\textwidth}
\includegraphics[width=\linewidth]{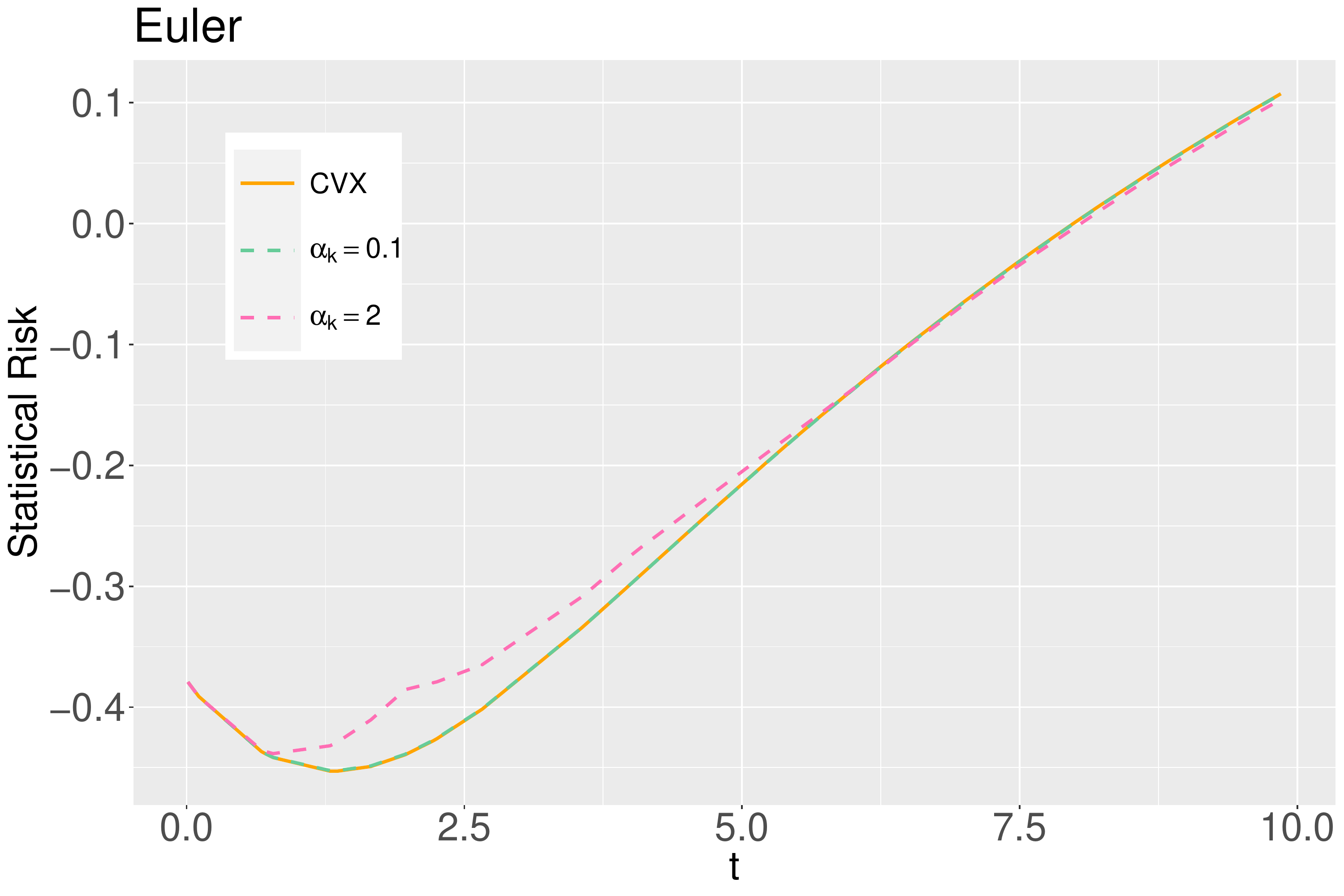}
\end{subfigure}
\hspace*{\fill}
\begin{subfigure}{0.48\textwidth}
\includegraphics[width=\linewidth]{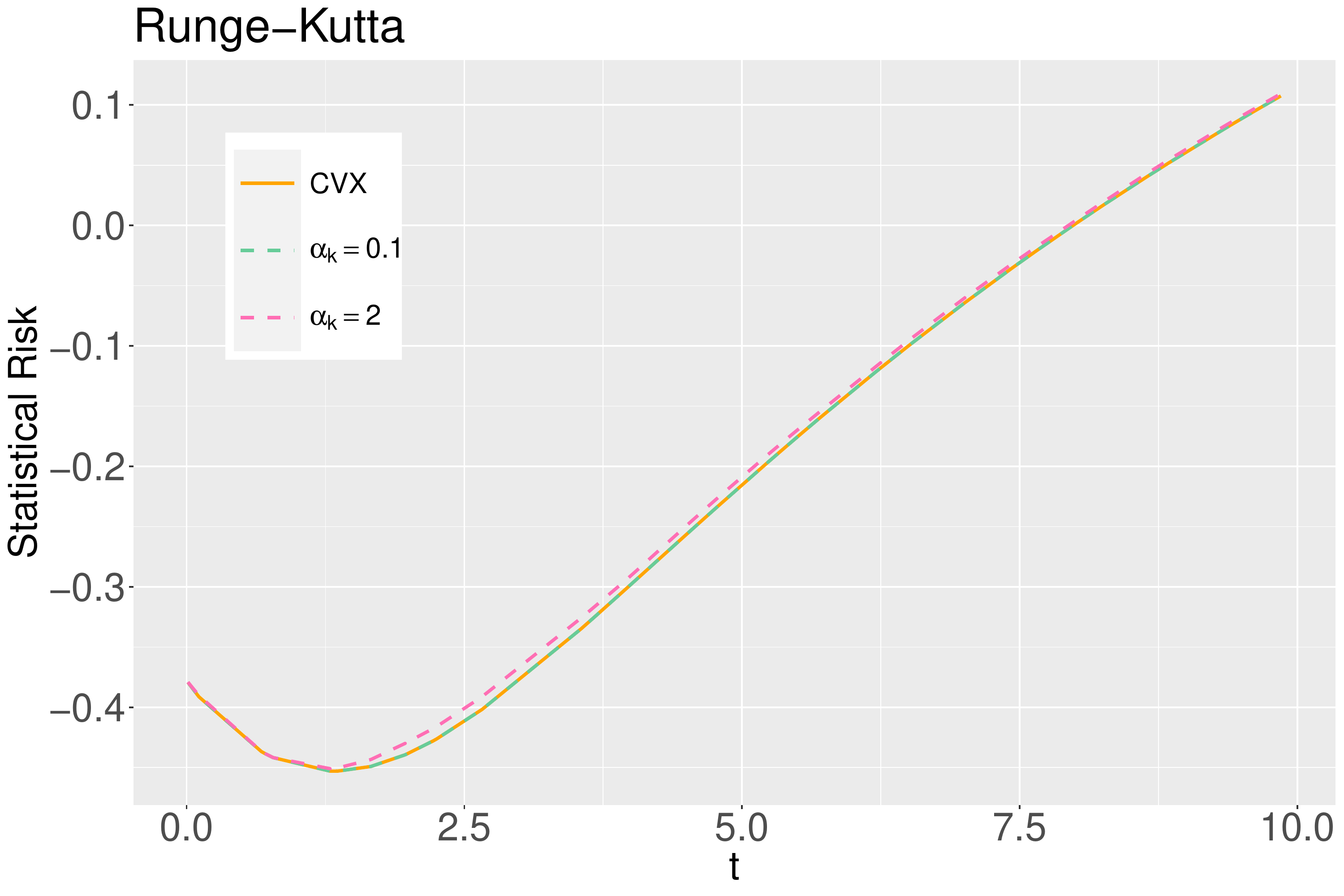}
\end{subfigure}

\medskip
\begin{subfigure}{0.48\textwidth}
\includegraphics[width=\linewidth]{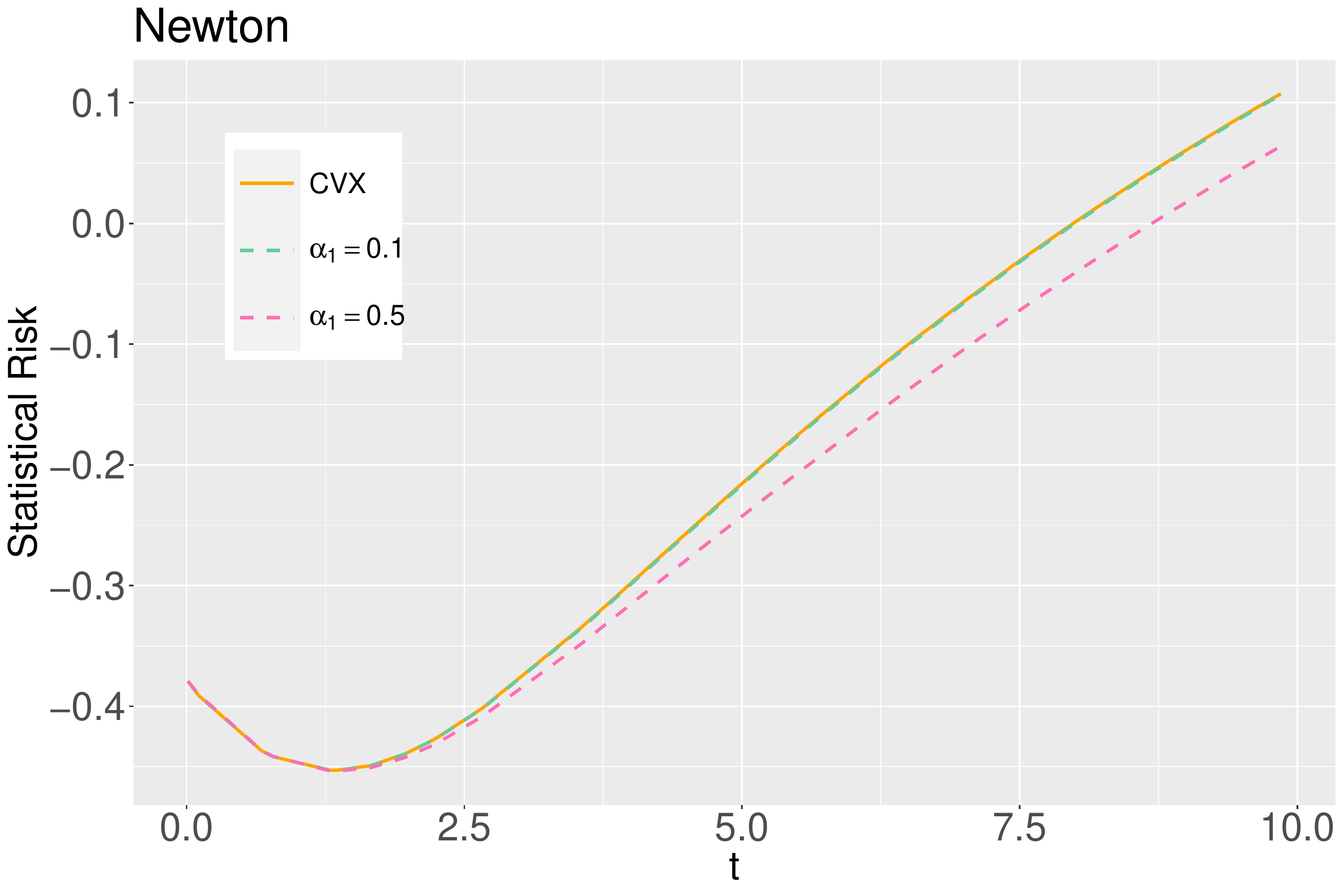}
\end{subfigure}
\hspace*{\fill}
\begin{subfigure}{0.48\textwidth}
\includegraphics[width=\linewidth]{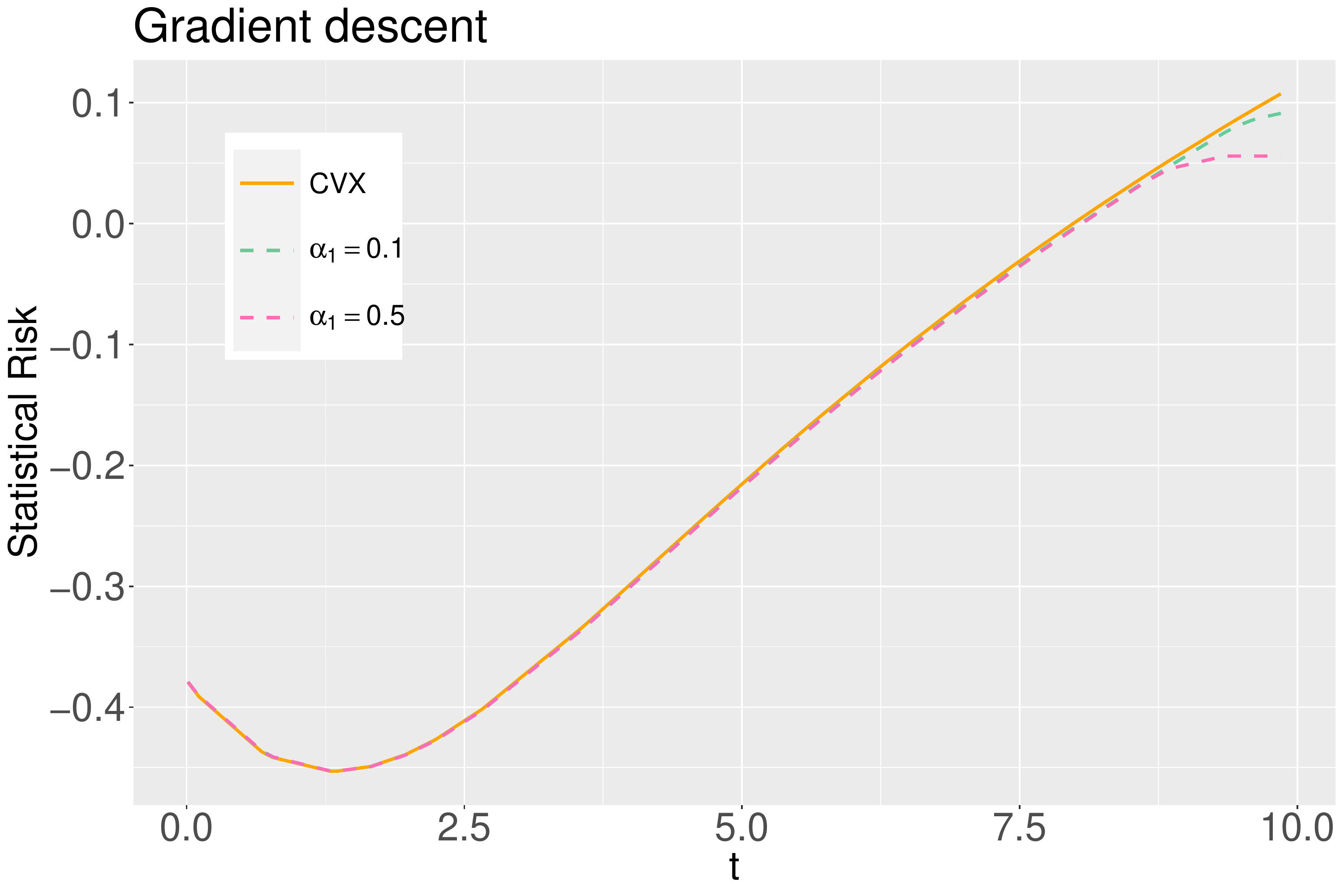}
\end{subfigure}
\caption{
Approximate risk curve $\log_{10}(R(\tilde \theta(t), \theta))$ of the proposed
algorithms applied to
$\ell_2$-regularized logistic regression when problem dimension is $(n,p)=(500,1000)$.
The CVX (orange) curve denotes the true risk curve
$\log_{10}(R(\theta(t), \theta))$ with
$\theta(t)$ computed using the CVX solver.
For algorithms with constant step size (Euler and Runge-Kutta),
 $\alpha_k$ denotes the step size;
while $\alpha_1$ denotes the initial step size for Newton and gradient descent method.
}
\label{figure:risk_no_GD_high_dimension}
\end{figure}

Figures \ref{figure:risk_no_GD_low_dimension}--\ref{figure:risk_no_GD_high_dimension}
plot the approximate risk curve $R(\tilde \theta(t),\theta)$
against the true risk curve (on a log scale)
by varying the initial step sizes for the proposed methods.
Note that under all scenarios, when the initial step size is $0.1$ (i.e., $\alpha_1 = 0.1$),
the approximate risk curves approximate the true risk curve quite well
for all four methods. This seems to suggest that good approximation error leads to
good approximation of the risk curve.
As the initial step size increases,
interestingly, we observe that
Runge-Kutta continues
to provide reasonable good results, suggesting
that they are more tolerant of a large initial step size (see the results
when $\alpha_k = 2$ for Runge-Kutta methods on
Figures \ref{figure:risk_no_GD_low_dimension}--\ref{figure:risk_no_GD_high_dimension}).
On the other hand, the Newton method and the gradient descent method
requires the initial step
sizes to be much smaller to obtain reasonable risk curve approximation.
That says, this does not necessarily
imply that the Newton method is less efficient
than the ODE-based methods, because the Newton method will adaptively
increase step sizes while
the ODE-based methods always fix their step sizes.

\section{Discussion}
\label{sec:discussion}
In this article, we established a formal connection between
$\ell_2$-regularized solution path and the solution of an ODE.
This connection provides an interesting algorithmic view of
$\ell_2$ regularization. In particular, the solution path
turns out to be similar to the iterates of a hybrid algorithm that
combines the gradient descent update and the Newton update. Moreover,
we proposed various new path-following algorithms to approximate the
$\ell_2$-regularized solution path. Global approximation-error bounds
for these methods are also derived, which in turn suggest some interesting
schemes for choosing the grid points. Computational complexities are also derived using
the proposed grid point schemes.

One important aspect
 we did not touch on is the statistical properties of
$\ell_2$-regularized solution path, which has been studied extensively
in the literature
\citep[see, e.g., ][and references therein]{dobriban2018}.
Interestingly,
\citet{ali2019continuous}, in the context of least squares regression,
connects the statistical properties of gradient descent iterates
to that of ridge regression solution path.
In particular, they show that the statistical risk of
the gradient descent path is no more than $1.69$ times that
of ridge regression, along the entire path. Motivated by
our proposed homotopy method based on damped gradient descent updates
\eqref{eq:grad_iterates},
it would be interesting
to investigate whether a
 damped version of gradient descent algorithm
would enjoy a more favorable statistical risk compared to regular
gradient descent. Further investigation is necessary.


\acks{
We would like to thank the Associate Editor and 
reviewers for their insightful comments and encouragement to revise our paper. 
The feedback substantially improved the paper. 
We would also like to acknowledge support for this project
from the National Science Foundation 
 (DMS-17-12580, DMS-17-21445 and DMS-20-15490). }


\newpage

\appendix

\section{Proofs of main results}


This section collects the proofs of
 Theorem \ref{thm:equiv_to_ode}--\ref{thm:gd_bd},
Corollary \ref{cor:solution_prop}--\ref{cor:GD_bound}, and
Proposition \ref{pro:function_examples} and \ref{pro:logistic_loss}.
Throughout this section, some standard results for $m$-strongly convex
functions will be repeatedly used in the proofs,
which are stated below. We omit their proofs as all of them can be found in
standard convex analysis textbooks \citep[see, e.g., ][]{boyd2004convex}.

Suppose that $f(\cdot)$ is a $m$-strongly convex function with minimizer $x^\star$. Then
for any $x$ and $y$,
\begin{align}
  & m \|x - y\|_2^2 \leq \langle \nabla f(x) - \nabla f(y) \, , x - y \rangle \leq \frac{1}{m}
  \|\nabla f(x) - \nabla f(y)\|_2^2 \, , \label{eq:str_con_prop1} \\
  & \frac{m}{2} \|x - x^\star\|_2^2  \leq f(x) - f(x^\star) \leq  \frac{1}{2m} \|\nabla f(x)\|_2^2
  \text{ and } \|x - x^\star\|_2 \leq \frac{1}{m} \|\nabla f(x)\|_2 \, .
  \label{eq:str_con_prop2}
\end{align}

\vskip .1in
\noindent
\textbf{Proof of Theorem \ref{thm:equiv_to_ode}. }
We first show differentiability of $\theta(t)$ at $t = 0$.
By the optimality of $\theta(t)$ and strong convexity of
the objective function, we have that for any $t \geq 0$,
\begin{equation*}
  \|\theta(t) - \bm 0\|_2^2 \leq
  (\theta(t) - \bm 0)^\top \left(\bm 0 - C(t) \nabla L_n(\bm 0)\right)
  \leq C(t) \|\theta(t)\|_2
  \left\|\nabla L_n(\bm 0)\right\|_2 \, ,
\end{equation*}
where we have used \eqref{eq:str_con_prop1}. This implies that $\|\theta(t)\|_2 \leq C(t)
\left\|\nabla L_n(\bm 0)\right\|_2$.
Thus $\theta(t)$ is continuous at $t = 0$ since $\lim_{t \rightarrow 0} C(t)
= C(0) = 0$ and $\bm 0 \in \textbf{dom}\, L_n$.
Moreover,
\begin{equation}
  \label{eq:theta_at_zero}
\frac{\theta(t)}{t} = -\frac{C(t)}{t} \nabla L_n(\theta(t))
\rightarrow - C'(0) \nabla L_n(\bm 0) \text{ as } t \rightarrow 0 \, ,
\end{equation}
where we have used the continuity of $\nabla L_n(\theta)$ and $\theta(t)$ at
$\theta = \bm 0$ and $t = 0$, respectively.
Therefore, $\theta(t)$ is differentiable at $t = 0$.

Next we show the differentiability of $\theta(t)$ for $t > 0$.
Denote by $f_t(\theta) = C(t) L_n(\theta) + \frac{1}{2} \|\theta\|_2^2$.
Since $f_t(\cdot)$ is $1$-strongly convex for all $t \geq 0$, by
using \eqref{eq:str_con_prop1} and the fact that $\nabla f_{t'}(\theta(t'))
= \nabla f_{t}(\theta(t)) = \bm 0$,
we have that for any $t > 0$
\begin{align}
  \|\theta(t') - \theta(t)\|_2^2 &\leq
  \langle \nabla f_{t'}(\theta(t')) - \nabla f_{t'}(\theta(t)) \, ,
  \theta(t') - \theta(t) \rangle =
  - \langle \nabla f_{t'}(\theta(t)) \, ,
  \theta(t') - \theta(t) \rangle \nonumber \\
  & =  - \langle C(t')\nabla L_n(\theta(t)) + \theta(t) \, ,
  \theta(t') - \theta(t) \rangle \nonumber \\
  & =  - \langle - C(t') \theta(t) / C(t)  + \theta(t)
  \, , \theta(t') - \theta(t) \rangle \label{eq:mono_theta}\\
  & \leq  \frac{|C(t') - C(t)|}{C(t)}
  \|\theta(t)\|_2 \|\theta(t') - \theta(t)\|_2 \, , \nonumber
\end{align}
which implies that
\begin{equation}
  \label{eq:lip_bound}
  \|\theta(t') - \theta(t)\|_2
  \leq \frac{|C(t') - C(t)|}{C(t)} \|\theta(t)\|_2 \, ,
\end{equation}
when $t > 0$. This gives a bound on how fast can $\theta(t)$ can vary
as $t$ increases. Next, we use this to establish differentiability of
$\theta(t)$.
Note that for any $t, t' \geq 0$
\begin{equation}
  \label{eq:gradient_eq}
  C(t) \nabla L_n(\theta(t)) + \theta(t) = 0 \text{ and }
  C(t') \nabla L_n(\theta(t')) + \theta(t') = 0 \, .
\end{equation}
Taking the difference, we obtain that
\begin{equation}
  \label{eq:minimum_growth_rate}
  \theta(t') - \theta(t) = - (C(t') - C(t)) \nabla L_n(\theta(t))
  - C(t') \nabla^2 L_n(\theta(t))(\theta(t') - \theta(t)) -
  C(t') \Delta \, ,
\end{equation}
where $\Delta = \nabla L_n(\theta(t')) - \nabla L_n(\theta(t)) -
\nabla^2 L_n(\theta(t)) (\theta(t') - \theta(t))$.
Rearranging and dividing both sides by $t'-t$, we obtain that
\begin{equation}
  \label{eq:theta_grad_der}
  \frac{\theta(t') - \theta(t)}{t' - t} = - \left(C(t')  \nabla^2 L_n(\theta(t))
  + I\right)^{-1} \left(\frac{C(t') - C(t)}{t' - t} \nabla L_n(\theta(t)) +
  C(t') \frac{\Delta}{t'-t} \right) \, ,
\end{equation}
where the matrix $C(t')  \nabla^2 L_n(\theta(t)) + I$
is invertible because $\nabla^2 L_n(\theta(t)) \succeq 0$.
Since $C(t)$ is differentiable, it remains to show that
$\|C(t') \frac{\Delta}{t'-t}\|_2 \rightarrow 0$ as $t' \rightarrow t$.
By Assumption (A0) and \eqref{eq:lip_bound},
we obtain that
  \begin{eqnarray*}
   && \left \|C(t') \frac{\Delta}{t'-t} \right \|_2  \\
   & \leq &
   \frac{C(t')}{|t'-t|}
   \int_0^1 \left\|
   \left[\nabla^2 L_n(\theta(t) +
   \tau(\theta(t') - \theta(t))) - \nabla^2 L_n(\theta(t))\right]
    (\theta(t') - \theta(t))\right\|_2 \,
   d \tau \\
    &\leq&
    \frac{C(t')}{|t'-t|}
    \sup_{0 \leq \tau \leq 1}
    \left| \rho \left(\nabla^2 L_n(\theta(t) +
    \tau(\theta(t') - \theta(t))) - \nabla^2 L_n(\theta(t))\right)
    \right| \|\theta(t') - \theta(t)\|_2
   \\
   &\leq& \frac{C(t')}{C(t)}
   \frac{|C(t') - C(t)|}{|t'-t|} \|\theta(t)\|_2
   \sup_{0 \leq \tau \leq 1}
   \left| \rho \left(\nabla^2 L_n(\theta(t) +
   \tau(\theta(t') - \theta(t))) - \nabla^2 L_n(\theta(t))\right)
   \right|
  \rightarrow 0
 \end{eqnarray*}
 as $t' \rightarrow t$,
 where $\rho(A)$ denotes the spectral norm of a matrix $A$, and
 we have used the fact that
 \begin{equation}
   \sup_{0 \leq \tau \leq 1} \left| \rho \left(\nabla^2 L_n(\theta(t) +
   \tau(\theta(t') - \theta(t))) - \nabla^2 L_n(\theta(t))\right)
   \right| \rightarrow 0 \text{ as } t' \rightarrow t
 \end{equation}
by Assumption (A0)
 and
$\frac{C(t') - C(t)}{t'-t} \rightarrow C^\prime(t)$
as $t' \rightarrow t$ since
$C(t)$ is differentiable.
Combining this with \eqref{eq:theta_grad_der}, it follows that
\begin{equation*}
  \theta^\prime(t) = \lim_{t' \rightarrow t}
  \frac{\theta(t') - \theta(t)}{t' - t}
  = -  C'(t) \left(C(t)  \nabla^2 L_n(\theta(t))
  + I\right)^{-1} \nabla L_n(\theta(t)) \, .
\end{equation*}
This completes the proof of Theorem \ref{thm:equiv_to_ode}.

\vskip .1in
\noindent
\textbf{Proof of Corollary \ref{cor:solution_prop}. }
To prove (i), rearranging terms in \eqref{eq:mono_theta}, we obtain that
\begin{equation*}
  (C(t') - C(t)) \left( \|\theta(t')\|_2^2 -
   \|\theta(t) \|_2^2 \right)
 \geq (C(t) + C(t')) \|\theta(t) - \theta(t')\|_2^2 \geq 0 \, ,
\end{equation*}
which implies that
$\|\theta(t)\|_2$ is nondecreasing in $t$.
For nonincreasingness of $L_n(\theta(t))$,
note that
\begin{eqnarray*}
  C(t') L_n(\theta(t')) + \frac{1}{2} \|\theta(t')\|_2^2
  &\leq& C(t') L_n(\theta(t)) + \frac{1}{2} \|\theta(t)\|_2^2 \\
  &\leq& (C(t') - C(t))L_n(\theta(t)) +
  C(t) L_n(\theta(t')) + \frac{1}{2} \|\theta(t')\|_2^2 \, ,
\end{eqnarray*}
which implies that $(C(t') - C(t))(L_n(\theta(t')) - L_n(\theta(t))) \leq 0$.
Hence, if $C(t') - C(t) > 0$
then $L_n(\theta(t')) \leq L_n(\theta(t))$, which proves that
$L_n(\theta(t))$ is nonincreasing in $t$.

To prove (ii), we modify the proof of \eqref{eq:lip_bound} to accommodate
the case where $L_n(\theta)$ might not be differentiable.
Note that for any $t > 0$ and any
$g_{t'} \in \partial L_n(\theta(t'))$ and $g_t \in \partial L_n(\theta(t))$, we have
$\langle  g_{t'} - g_t \, ,
  \theta(t') - \theta(t) \rangle \geq 0$,
where $\partial L_n(\theta)$ denotes the \textit{subdifferential} of
$L_n(\cdot)$ at $\theta$.  Hence, for any
  $h_{t'} \in \partial f_{t'}(\theta(t')), h_t \in \partial f_{t'}(\theta(t))$
\begin{equation}
\langle h_{t'} - h_{t} \, ,
  \theta(t') - \theta(t) \rangle
  \geq  \|\theta(t') - \theta(t)\|_2^2
\end{equation}
Since $\bm 0 \in  \partial f_{t'}(\theta(t'))$ and
$-C(t') \theta(t) / C(t) + \theta(t)
\in \partial f_{t'}(\theta(t))$, substituting $h_{t'}$ with $\bm 0$ and
$h_t$ with $-C(t') \theta(t) / C(t) + \theta(t)$, we obtain that
\begin{equation}
 \langle C(t') \theta(t) / C(t) - \theta(t) \, ,
  \theta(t') - \theta(t) \rangle
  \geq  \|\theta(t') - \theta(t)\|_2^2 \, ,
\end{equation}
which implies that
\begin{equation*}
  \|\theta(t') - \theta(t)\|_2^2
  \leq \langle C(t') \theta(t) / C(t) - \theta(t) \, ,
  \theta(t') - \theta(t) \rangle
  \leq |C(t') / C(t) - 1| \|\theta(t)\|_2  \|\theta(t') - \theta(t)\|_2 \, ,
\end{equation*}
which proves \eqref{eq:lip_bound} when $L_n(\theta)$ might not be differentiable.
Using this, we have that for any $t > t'$ and $\theta(t) \neq \theta(t')$,
\begin{equation}
  \|\theta(t)\|_2 - \|\theta(t')\|_2
  \leq \|\theta(t') - \theta(t)\|_2
  \leq (C(t) - C(t'))  \|\theta(t)\|_2 / C(t) \, ,
\end{equation}
which implies that
\begin{equation}
  \|\theta(t)\|_2 / C(t) \leq \|\theta(t')\|_2 / C(t') \, .
\end{equation}
This also holds when $\theta(t) = \theta(t')$ because $C(t)$ is an increasing function.
This proves that part (ii).

Lastly, we prove part (iii). Denote by $\theta^\star$ the minimum $\ell_2$ norm
minimizer of $L_n(\theta)$.
Next, we show that $\theta(t)$ converges to $\theta^\star$ as
$t \rightarrow \infty$ if $\theta^\star$ is finite.
Note that $\bm 0 \in \partial L_n(\theta^\star)$ and
$\bm 0 \in C(t)\partial L_n(\theta(t)) + \theta(t)$.
As a result,
\begin{equation*}
  \bm 0 \in C(t)\left(\partial L_n(\theta(t)) - \partial L_n(\theta^\star) \right)
   +  \theta(t) \, ,
\end{equation*}
where $A - B$ denotes the set $\{a - b: a \in A \text{ and } b \in B\}$.
Multiplying $\theta(t) - \theta^\star$ on both sides, we obtain
that
\begin{equation*}
(\theta(t) - \theta^\star)^\top \theta(t)
\in  - C(t)(\theta(t) - \theta^\star)^\top
\left(\partial L_n(\theta(t)) - \partial L_n(\theta^\star) \right) \, ,
\end{equation*}
which implies that $(\theta(t) - \theta^\star)^\top \theta(t) \leq 0$.
Therefore, $\|\theta(t)\|_2^2 \leq (\theta^\star)^\top \theta(t)
  \leq \|\theta^\star\|_2 \|\theta(t)\|_2$, which implies that
$\|\theta(t)\|_2 \leq \|\theta^\star\|_2<\infty$ for any $t \geq 0$.
%
Denote by $\bar{\theta}$ the limit of
any converging subsequence $\theta(t_k)$, that is, $\bar{\theta} =
 \lim_{k \rightarrow \infty} \theta(t_k)$ for some
$t_k \rightarrow \infty$. Then,
$\|\bar{\theta}\|_2 = \lim_{k \rightarrow \infty} \|\theta(t_k)\|_2
\leq \|\theta^\star\|_2$.
Next, we show that $\bar{\theta}$ must also be a
minimizer of $L_n(\theta)$. To this end,
note that $L_n(\bar{\theta}) =
\lim_{k \rightarrow \infty} L_n(\theta(t_k))$
by using the continuity of
$L_n(\theta)$ in $\theta$. Moreover,
by optimality of $\theta(t_k)$,
\begin{equation}
  \label{eq:subopt_unpenalized}
  L_n(\theta(t_k)) \leq L_n(\theta(t_k)) + \frac{1}{2C(t_k)}
  \|\theta(t_k)\|_2^2 \leq L_n(\theta^\star)
  + \frac{1}{2C(t_k)} \|\theta^\star\|_2^2 \, .
\end{equation}
By letting $k \rightarrow \infty$
and using the fact that $L_n(\bar{\theta})
= \lim_{k \rightarrow \infty} L_n(\theta(t_k))$ due to continuity of $L_n(\theta)$,
we have that
\begin{equation*}
  L_n(\bar{\theta}) =
  \lim_{k \rightarrow \infty} L_n(\theta(t_k))
  \leq \lim_{k \rightarrow \infty}
  \left(L_n(\theta^\star) + \frac{1}{2C(t_k)} \|\theta^\star\|_2^2 \right)
  = L_n(\theta^\star) \, ,
\end{equation*}
where the last step uses the
assumption that $\|\theta^\star\|_2 < \infty$.
This proves that $\bar{\theta}$ must also be a minimizer of $L_n(\theta)$.

Now if $\bar{\theta} \neq \theta^\star$,
then their convex combination
$\frac{1}{2}(\bar{\theta} + \theta^\star)$ must also be
a minimizer of $L_n(\theta)$ due to the convexity of $L_n(\theta)$.
On the other hand, the convex combination
has strictly smaller norm than that of $\theta^\star$, because
 $\|\frac{1}{2}(\bar{\theta} + \theta^\star)\|_2 <
\frac{1}{2}(\|\bar{\theta}\| + \|\theta^\star\|_2)
\leq \|\theta^\star\|_2$. This contradicts with the
 definition of $\theta^\star$.
Hence, we must have $\lim_{k \rightarrow \infty}
\theta(t_k) = \bar{\theta} = \theta^\star$ for every converging subsequence
$\theta(t_k)$. Consequently, the sequence $\theta(t)$
must converge to $\theta^\star$.
This completes the proof of Corollary \ref{cor:solution_prop}.

\vskip .1in
\noindent
\textbf{Proof of Theorem \ref{thm:uniform_bd_general}. }
For any $t \in [t_k, t_{k+1}]$, we let $w_k = \frac{t_{k+1}-t}{t_{k+1} - t_k}$,
for $k = 0, 1, \ldots, N-1$. Then
$\tilde \theta(t) = w_k \theta_k + (1-w_k) \theta_{k+1}$.
By convexity of $f_t(\cdot)$, we have
$f_t(\tilde \theta(t)) \leq w_k f_t(\theta_k) + (1-w_k) f_t(\theta_{k+1})$.
Thus,
\begin{align}
\label{eq:thm2_tmp2}
   f_t(\tilde \theta(t)) - f_t(\theta(t))
  \leq & w_k(f_t(\theta_k) - f_t(\theta(t)) )
   + (1-w_k) (f_t(\theta_{k+1}) - f_t(\theta(t)) ) \, .
 \end{align}
For any $k = 1, \ldots, N-1$, the term $f_t(\theta_k) - f_t(\theta(t))$ in
\eqref{eq:thm2_tmp2} can be
bounded as follows:
 \begin{align*}
  f_t(\theta_k) & - f_t(\theta(t))
    \leq
   \frac{1}{2e^{-t}} \|\nabla f_t(\theta_k)\|_2^2
   = \frac{e^{t}}{2} \left \|\frac{1-e^{-t}}{1 - e^{-t_k}}\nabla f_{t_k}(\theta_k)
   + \frac{e^{-t} - e^{-t_k}}{1 - e^{-t_k}} \theta_k \right \|_2^2 \\
   & \leq   e^{t} \left( \frac{1-e^{-t}}{1 - e^{-t_k}} \right)^2
   \|\nabla f_{t_k}(\theta_k)\|_2^2 +
   e^{t} \left( \frac{e^{-t} - e^{-t_k}}{1 - e^{-t_k}} \right)^2 \|\theta_k\|_2^2 \\
  & =
  e^{t} \left( \frac{1-e^{-t}}{1 - e^{-t_k}} \right)^2 \|g_k\|_2^2
  + e^{t} \left( \frac{e^{-t} - e^{-t_k}}{1 - e^{-t_k}} \right)^2 \|\theta_k\|_2^2
  \, ,
 \end{align*}
 where the first inequality uses the fact that $f_t(\cdot)$ is $e^{-t}$-strongly
 convex and \eqref{eq:str_con_prop2}.
 Similarly, we can bound the term $f_t(\theta_{k+1}) - f_t(\theta(t))$  by
 \begin{equation*}
   e^{t} \left( \frac{1-e^{-t}}{1 - e^{-t_{k+1}}} \right)^2
  \|g_{k+1}\|_2^2
  + e^{t} \left( \frac{e^{-t} - e^{-t_{k+1}}}{1 - e^{-t_{k+1}}} \right)^2
  \|\theta_{k+1}\|_2^2
 \end{equation*}
 for any $k = 0, 1, \ldots, N-1$.
 Combining these two bounds, we have that
 \begin{align*}
   & w_k(f_t(\theta_k) - f_t(\theta(t)))
   + (1-w_k) (f_t(\theta_{k+1}) - f_t(\theta(t))) \\
   & \leq e^{t_{k+1}}
   \max \left\{\left( \frac{1-e^{-t_{k+1}}}{1 - e^{-t_k}} \right)^2  \|g_k\|_2^2, \,
  \|g_{k+1}\|_2^2 \right\} \\
   & \quad
   +  (e^{-t_k} - e^{-t_{k+1}})^2 \max \left\{
   \frac{e^{t_{k+1}} \|\theta_k\|_2^2 }{(1 - e^{-t_{k}})^2} , \,
   \frac{e^{t_k} \|\theta_{k+1}\|_2^2 }{(1 - e^{-t_{k+1}})^2} \right\} \, ,
 \end{align*}
for any $k = 1,\ldots, N-1$. This proves \eqref{eq:thm_fn_bound2}.

 When $k = 0$, the term $f_t(\theta_k) - f_t(\theta(t))$ in
\eqref{eq:thm2_tmp2} can be
bounded as follows
 \begin{equation*}
  f_t(\theta_0) - f_t(\theta(t)) =  f_t(\bm 0) - f_t(\theta(t)) \leq
  \frac{1}{2e^{-t}} \|\nabla f_{t}(\bm 0)\|_2^2 = \frac{e^{t} (1-e^{-t})^2}{2}
  \|\nabla L_n(\bm 0)\|_2^2
 \end{equation*}
 for any $0 \leq t < t_1$, where we have used \eqref{eq:str_con_prop2} in the above inequality.
Following a similar argument as before, we obtain that
\begin{eqnarray*}
   && w_0(f_t(\theta_0) - f_t(\theta(t)))
   + (1-w_0) (f_t(\theta_{1}) - f_t(\theta(t))) \\
   & \leq & \frac{e^{t_1} (1-e^{-t_1})^2}{2}
  \|\nabla L_n(\bm 0)\|_2^2 +
  \max\left( e^{t_1} \|g_1\|_2^2 , \,
    \|\theta_1\|_2^2 \right)
 \end{eqnarray*}
for any $t \in [0, t_1]$. This proves
\eqref{eq:thm_fn_bound1}.

Now we bound $f_t(\tilde \theta(t)) - f_t(\theta(t))$ when $t_N < t \leq t_{\max}$.
Toward this end,
notice that
\begin{align*}
  & f_t(\tilde \theta(t)) - f_t(\theta(t)) =
  f_t(\theta_N) - f_t(\theta(t)) \\
  = & \frac{1-e^{-t}}{1-e^{-t_N}}
  (f_{t_N}(\theta_{N}) - f_{t_N}(\theta(t_N)))
  + \frac{e^{-t_N} - e^{-t}}{2(1-e^{-t_N})} (\|\theta(t_N)\|_2^2 - \|\theta_N\|_2^2)
  + f_{t}(\theta(t_N)) - f_{t}(\theta(t)) \, .
\end{align*}
Next, we bound these three terms separately.
For the first term, by using \eqref{eq:str_con_prop2}, we have that
$f_{t_N}(\theta_{N}) - f_{t_N}(\theta(t_N)) \leq 2^{-1}e^{t_N} \|g_N\|_2^2$. Using this,
we obtain that
\begin{equation}
   \frac{1-e^{-t}}{1-e^{-t_N}}
  (f_{t_N}(\theta_{N}) - f_{t_N}(\theta(t_N)))
  \leq \frac{(1-e^{-t})e^{t_N}}{2(1-e^{-t_N})} \|g_N\|_2^2 \, .
\end{equation}
For the second term, note that
\begin{align*}
  &\|\theta(t_N)\|_2^2 - \|\theta_N\|_2^2
  = 2(\theta(t_N) - \theta_N)^\top \theta(t_N) -
   \|\theta(t_N) - \theta_N\|_2^2 \\
  & \leq
   (2\|\theta(t_N)\|_2 - \|\theta(t_N) - \theta_N\|_2)\|\theta(t_N) - \theta_N\|_2
   \leq 2\|\theta(t_N)\|_2\|\theta(t_N) - \theta_N\|_2 \\
   & \leq 2\|\theta(t_N)\|_2 e^{t_N} \|g_N\|_2 = 2 e^{t_N} \|\theta(t_N)\|_2\|g_N\|_2 \, .
\end{align*}
Thus, the second term can be bounded by
$\frac{1 - e^{-(t -t_N)}}{1-e^{-t_N}} \|\theta(t_N)\|_2 \|g_N\|_2$, which
can be further bounded using the Cauchy–Schwarz inequality:
\begin{equation*}
  \frac{1 - e^{-(t -t_N)}}{1-e^{-t_N}} \|\theta(t_N)\|_2 \|g_N\|_2
  \leq \frac{1}{2} \left(
  \frac{(1-e^{-t})e^{t_N}}{1-e^{-t_N}} \|g_N\|_2^2
  + \frac{(1 - e^{-(t -t_N)})^2}{(1-e^{-t_N})(1-e^{-t})e^{t_N}}
  \|\theta(t_N)\|_2^2\right) \, .
\end{equation*}
To bound the third term, by optimality of $\theta(t_N)$, we have
$f_{t_N}(\theta(t_N)) \leq f_{t_N}(\theta(t))$,
which in turn implies that
$L_n(\theta(t_N)) - L_n(\theta(t)) \leq .5 (e^{t_N} - 1)^{-1}
\left(\|\theta(t)\|_2^2 - \|\theta(t_N)\|_2^2\right)$.
Using this, the third term can be bounded as follows
\begin{align*}
  & f_{t}(\theta(t_N)) - f_{t}(\theta(t))
  = (1-e^{-t}) (L_n(\theta(t_N)) - L_n(\theta(t))) + \frac{e^{-t}}{2}
  \left(\|\theta(t_N)\|_2^2
  - \|\theta(t)\|_2^2\right) \\
  & \leq \frac{1-e^{-t}}{2(e^{t_N} - 1)} \left(\|\theta(t)\|_2^2
  - \|\theta(t_N)\|_2^2\right) +
   \frac{e^{-t}}{2}
  \left(\|\theta(t_N)\|_2^2
  - \|\theta(t)\|_2^2\right) \\
  & = \frac{1 - e^{-(t-t_N)}}{2(e^{t_N} - 1)}
  \left(\|\theta(t)\|_2^2
  - \|\theta(t_N)\|_2^2\right) \, .
\end{align*}
Combining the three bounds and using the fact that
\begin{equation*}
  \frac{1 - e^{-(t-t_N)}}{2(e^{t_N} - 1)}
  = \frac{(1 - e^{-(t -t_N)})^2}{2(1-e^{-t_N})(1-e^{-t})e^{t_N}}
  + \frac{1 - e^{-(t-t_N)}}{2(e^{t} - 1)}   \, ,
\end{equation*}
we obtain that
\begin{align*}
   \sup_{t_N < t \leq t_{\max}} & \left\{ f_t(\tilde \theta(t)) - f_t(\theta(t)) \right\}
   \leq  \sup_{t_N < t \leq t_{\max}} \left\{
  \frac{(1-e^{-t})e^{t_N}}{1-e^{-t_N}} \|g_N\|_2^2 +
  \frac{1 - e^{-(t-t_N)}}{2(e^{t_N} - 1)} \|\theta(t)\|_2^2
  \right\} \\
  & \leq  \frac{e^{t_N}(1-e^{-t_{\max}})}{1 - e^{-t_N}} \|g_N\|_2^2 +
   \sup_{t_N < t \leq t_{\max}}
  \frac{(1 - e^{-(t -t_N)})^2}{2(1-e^{-t_N})(1-e^{-t})e^{t_N}}
  \|\theta(t)\|_2^2  \\
  & \qquad + \sup_{t_N < t \leq t_{\max}}
  \frac{1 - e^{-(t-t_N)}}{2(e^{t} - 1)} \left(\|\theta(t)\|_2^2
  - \|\theta(t_N)\|_2^2\right) \, .
\end{align*}
Moreover, by \eqref{eq:lip_bound}, we have that
\begin{align*}
\|\theta(t)\|_2^2 - \|\theta(t_N)\|_2^2
& \leq
  \|\theta(t) - \theta(t_N)\|_2
  (\|\theta(t)\|_2 + \|\theta(t_N)\|_2) \\
& \leq
\frac{e^{t} - e^{t_N}}{e^{t} - 1}
  \|\theta(t)\|_2 (\|\theta(t)\|_2 + \|\theta(t_N)\|_2)
  \leq
  2 \frac{e^{t} - e^{t_N}}{e^{t} - 1}
  \|\theta(t)\|_2^2
\end{align*}
Combining the above two inequalities and using the fact that
\begin{equation*}
  \sup_{t_N < t \leq t_{\max}} \frac{(1 - e^{-(t-t_N)})^2}{1 - e^{-t}}
  = \frac{(1 - e^{-(t_{\max}-t_N)})^2}{1 - e^{-t_{\max}}} \, ,
\end{equation*}
we obtain that
\begin{align*}
  \sup_{t_N < t \leq t_{\max}} & \left\{ f_t(\tilde \theta(t)) - f_t(\theta(t)) \right\}
   \leq
   \frac{e^{t_N}(1-e^{-t_{\max}})}{1 - e^{-t_N}} \|g_N\|_2^2  +
   \frac{3(1 - e^{-(t_{\max} -t_N)})^2}{2(e^{t_N}-1)(1-e^{-t_{\max}})}
  \|\theta(t_{\max})\|_2^2 \, ,
\end{align*}
which implies \eqref{eq:thm_fn_bound3}.
This completes the proof of Theorem \ref{thm:uniform_bd_general}.

\vskip .1in
Next, we present a supporting lemma for the proof of Theorem \ref{thm:newton}.
\begin{lem}
\label{lm:thm3}
Under Assumption (A0),
   we have that
  \begin{equation}
  \label{eq:theta_norm_bds}
 \|\theta(t)\|_2 \leq (e^t - 1) \|\nabla L_n(\bm 0)\|_2
 \text{ and } \|\theta(t)\|_2 \leq
 (1-e^{-t}) \left(\|\nabla L_n(\bm 0)\|_2  +  \|\theta(t')\|_2 \right)
\end{equation}
for any $t' \geq t > 0$.
\end{lem}

\vskip .1in
\noindent
\textbf{Proof of Lemma \ref{lm:thm3}. }
Since $f_t(\cdot)$ is $e^{-t}$ strongly convex, using \eqref{eq:str_con_prop2}, we have
\begin{equation*}
  \|\theta(t) - \bm 0\|_2 \leq e^{t} \|\nabla f_t(\bm 0)\|_2 =
  (e^t - 1) \|\nabla L_n(\bm 0)\|_2 \, ,
\end{equation*}
which proves the first inequality $\|\theta(t)\|_2 \leq (e^{t} - 1)
\|\nabla L_n(\bm 0)\|_2$. Combining this with the fact that $\|\theta(t)\|_2 \leq
\|\theta(t')\|_2$, we have that
\begin{equation*}
  \|\theta(t)\|_2 \leq \min\left((e^{t} - 1)
\|\nabla L_n(\bm 0)\|_2, \,  \|\theta(t')\|_2\right) \leq
(1-e^{-t}) \left(\|\nabla L_n(\bm 0)\|_2  +  \|\theta(t')\|_2 \right) \, ,
\end{equation*}
which proves the second inequality in \eqref{eq:theta_norm_bds}.
This completes the proof of Lemma \ref{lm:thm3}.

\vskip .1in
\noindent
\textbf{Proof of Theorem \ref{thm:newton}. }
   Note that
\begin{align*}
  g_{k+1} &=
  (1 - e^{-t_{k+1}}) \nabla L_n(\theta_{k+1}) +
  e^{-t_{k+1}} \theta_{k+1}  \\
  & =  \underbrace{
  (1 - e^{-t_{k+1}})  \left( \nabla L_n(\theta_{k+1})
  - \nabla L_n(\theta_{k}) - \nabla^2 L_n(\theta_k) (\theta_{k+1} -
  \theta_{k}) \right)}_{\text{Part I}} + \\
   & \quad
   \underbrace{(1 - e^{-t_{k+1}}) \left( \nabla L_n(\theta_{k}) +
   \nabla^2 L_n(\theta_k) (\theta_{k+1} -
  \theta_{k}) \right) + e^{-t_{k+1}} \theta_{k+1}}_\text{Part II} \, .
\end{align*}
Moreover, based on the definition of $\theta_{k+1}$, we have that
\begin{equation}
  \left((1 - e^{-t_{k+1}}) \nabla^2 L_n(\theta_k) +
  e^{-t_{k+1}} I\right) (\theta_{k+1} - \theta_k)
  + (1 - e^{-\alpha_{k+1}}) \nabla L_n(\theta_k) + e^{-\alpha_{k+1}}g_k = 0
  \, .
\end{equation}
Combining this with the fact that $g_k =
(1 - e^{-t_{k}})  \nabla L_n(\theta_{k}) + e^{-t_{k}} \theta_k$, we obtain that
\begin{eqnarray*}
  \text{Part II}
  & = & \left( (1 - e^{-t_{k+1}})  \nabla^2 L_n(\theta_k)
  + e^{-t_{k+1}} I  \right) (\theta_{k+1} - \theta_k) +
  (1 - e^{-t_{k+1}})  \nabla L_n(\theta_{k}) + e^{-t_{k+1}} \theta_k \\
  &=&  (1 - e^{-t_{k+1}})  \nabla L_n(\theta_{k}) + e^{-t_{k+1}} \theta_k
  -  (1 - e^{-\alpha_{k+1}}) \nabla L_n(\theta_k) - e^{-\alpha_{k+1}}g_k \\
  &=&
   e^{-\alpha_{k+1}} \left( (1 - e^{-t_k})\nabla L_n(\theta_{k}) +
   e^{-t_k} \theta_k -g_k\right) =  \bm 0 \, .
\end{eqnarray*}
Hence, we have that
\begin{eqnarray}
  \|g_{k+1}\|_2
  & = & (1 - e^{-t_{k+1}})  \left\|  \nabla L_n(\theta_{k+1})
  - \nabla L_n(\theta_{k}) - \nabla^2 L_n(\theta_k) (\theta_{k+1} -
  \theta_{k}) \right\|_2 \nonumber \\
  & \leq & \beta (1 - e^{-t_{k+1}}) (\theta_{k+1} - \theta_{k})^\top
  \left[\nabla^2 L_n(\theta_k)\right]^{\gamma_1}
  (\theta_{k+1} - \theta_{k}) \, ,
  \label{eq:grad_inintial_bd}
\end{eqnarray}
where the last inequality uses \eqref{eq:hessian_cond}
in Assumption (A1), provided
that
\begin{equation*}
  (\theta_{k+1} - \theta_k)^\top [\nabla^2 L_n(\theta_k)]^{\gamma_2}
 (\theta_{k+1} - \theta_k) \leq \beta^{-2} \, ,
\end{equation*}
which is to be verified later by induction.
Next, we define
\begin{equation}
  H_{k+1} = (1 - e^{-t_{k+1}}) \nabla^2 L_n(\theta_k) + e^{-t_{k+1}} I
  \text{ and } J_{k+1} = H_{k+1}^{-1} \left[\nabla^2 L_n(\theta_k)\right]^{\gamma_1}
   H_{k+1}^{-1} \, .
\end{equation}
Notice that $\theta_{k+1} - \theta_{k}
= - H_{k+1}^{-1} \left((1 - e^{-\alpha_{k+1}}) \nabla L_n(\theta_k)
+ e^{-\alpha_{k+1}}g_k \right)$. Combining this
with $g_k = (1 - e^{-t_k})\nabla L_n(\theta_k) + e^{-t_k} \theta_k$,
we obtain that,
\begin{align}
\theta_1 & =  - (1 - e^{-\alpha_1}) H_1^{-1} \nabla L_n(\bm 0),
\label{eq:theta1_newton} \\
  \theta_{k+1} - \theta_{k}
  & =  - H_{k+1}^{-1} \left((1 - e^{-\alpha_{k+1}}) \nabla L_n(\theta_k)
    + e^{-\alpha_{k+1}}g_k \right)  \nonumber \\
  & =  - H_{k+1}^{-1} \left( \frac{1 - e^{-t_{k+1}}}{1 - e^{-t_k}} g_k
  - \frac{e^{-t_k} - e^{-t_{k+1}}}{1 - e^{-t_k}} \theta_k
  \right)
  \text{ for any }
  k \geq 1 \, .  \label{eq:thetak_iter_difference}
\end{align}
Combining \eqref{eq:grad_inintial_bd} and \eqref{eq:thetak_iter_difference},
and using the fact that $\| a + b \|_2^2 \leq
2 (\| a\|_2^2 + \|b \|_2^2)$ for any vectors $a$ and $b$,
we have that
\begin{align}
\|g_1\|_2 & \leq \beta (1 - e^{-\alpha_1}) \theta_1^\top
[\nabla^2 L_n(\bm 0)]^{\gamma_1} \theta_1
\label{eq:deri_grad_k1} \\
  \|g_{k+1}\|_2
   & \leq  2 \beta \lambda_{\max}(J_{k+1}) (1 - e^{-t_{k+1}})
\left\{ \frac{(1 - e^{-t_{k+1}})^2}{(1 - e^{-t_k})^2} \|g_k \|_2^2
   + \frac{(e^{-t_k} - e^{-t_{k+1}})^2}{(1 - e^{-t_k})^2}\|\theta_k\|_2^2  \right\}
   \label{eq:gk_bound1}
\end{align}
for any $k \geq 1$. Using \eqref{eq:gk_bound1} and the fact that
\begin{equation}
 \lambda_{\text{max}}(J_{k+1}) \leq
  \sup_{\lambda: \lambda \geq 0}
   \frac{\lambda^{\gamma_1}}{((1-e^{-t_{k+1}})\lambda + e^{-t_{k+1}})^2}
   = \frac{e^{(2 - \gamma_1)t_{k+1}}}{4 (1 - e^{-t_{k+1}})^{\gamma_1}}
    (2-\gamma_1)^{2-\gamma_1} \gamma_1^{\gamma_1}
    \label{eq:eigen_bd}
\end{equation}
for any $0 \leq \gamma_1 \leq 2$ and $k \geq 0$,
 we obtain
that for any $k \geq 1$,
\begin{align}
  \|g_{k+1}\|_2
   &\leq
   \frac{\beta h(\gamma_1)e^{(2-\gamma_1)t_{k+1}}}{2(1 - e^{-t_{k+1}})^{\gamma_1-1}}
   \Bigg\{ \frac{(1-e^{-t_{k+1}})^2}{(1 - e^{-t_k})^2}  \|g_k\|_2^2
   + \frac{(e^{-t_k} - e^{-t_{k+1}})^2}{(1-e^{-t_k})^2} \|\theta_k\|_2^2 \Bigg\} \, ,
   \label{eq:deri_grad}
\end{align}
where $h(\gamma_1) = (2-\gamma_1)^{2-\gamma_1} \gamma_1^{\gamma_1}$.
Throughout the proof, we shall treat $h(\gamma)$ as an absolute constant as
$1 \leq h(\gamma) \leq 4$ for $0 \leq \gamma \leq 2$.
We then use induction to show that
\begin{align}
\label{eq:grad_bound_theta_star_finite}
   \|g_{k+1}\|_2 & \leq
   \frac{c_0 \beta h(\gamma_1)e^{(2-\gamma_1)t_{k+1}}}{2(1 - e^{-t_{k+1}})^{\gamma_1-1}}
   \frac{(e^{-t_k} - e^{-t_{k+1}})^2}{(1-e^{-t_k})^2} \|\theta_k\|_2^2
\end{align}
for any $k \geq 1$ and $c_0 = 26/25$.
To this end, in view of \eqref{eq:deri_grad},
we only need to show that
\begin{equation*}
  \frac{1-e^{-t_{k+1}}}{1 - e^{-t_k}}  \|g_k\|_2
    \leq (c_0 - 1)^{1/2}
   \frac{(e^{-t_k} - e^{-t_{k+1}})}{(1-e^{-t_k})} \|\theta_k\|_2  \, ,
\end{equation*}
or equivalently,
\begin{equation}
\label{eq:ineq_induction_step_to_check}
  \|g_k\|_2
    \leq (c_0 - 1)^{1/2}
    \frac{(e^{-t_k} - e^{-t_{k+1}})}{(1-e^{-t_{k+1}})} \|\theta_k\|_2
\end{equation}
for any $k \geq 1$.
To verify this,
our plan is to show that (i)
inequality \eqref{eq:ineq_induction_step_to_check}
holds for $k = 1$ using the bound in
 \eqref{eq:deri_grad_k1} for $\|g_1\|_2$; and (ii)
inequality \eqref{eq:ineq_induction_step_to_check} holds for
$\|g_k\|_2$ if the bound \eqref{eq:grad_bound_theta_star_finite} holds
for $\|g_k\|_2$ and inequality \eqref{eq:ineq_induction_step_to_check} holds
for $\|g_{k-1}\|_2$.

When $k = 1$, we note that
\begin{align*}
\left \|[\nabla^2 L_n(\bm 0)]^{\gamma_1} \theta_1 \right\|_2
& =  (1 - e^{-\alpha_1})
\left \|[\nabla^2 L_n(\bm 0)]^{\gamma_1} H_1^{-1} \nabla L_n(\bm 0) \right\|_2
\\
& =  (1 - e^{-\alpha_1}) \sqrt{ [\nabla L_n(\bm 0)]^\top
H_1^{-1} [\nabla^2 L_n(\bm 0)]^{2\gamma_1} H_1^{-1} \nabla L_n(\bm 0) }\\
& \leq
\begin{cases}
   2^{-1} \sqrt{h(2\gamma_1)}
    e^{(1-\gamma_1)\alpha_1} (1-e^{-\alpha_1})^{1-\gamma_1}
  \|\nabla L_n(\bm 0)\|_2
  & \text{ if } \gamma_1 \leq 1; \\
 \min \left( \nu^{\gamma_1 - 1}, \,
 \nu^{\gamma_1} e^{-\alpha_1}(1- e^{-\alpha_1})
 \right)
 \|\nabla L_n(\bm 0)\|_2 & \text{ if } \gamma_1 > 1 \, .
\end{cases}
\end{align*}
where $\nu$ denotes the largest eigenvalue of $\nabla^2 L_n(\bm 0)$, and
we have used \eqref{eq:eigen_bd}.

Combining this with \eqref{eq:deri_grad_k1}, we obtain that
\begin{align}
\|g_1\|_2 & \leq \beta (1 - e^{-\alpha_1}) \theta_1^\top
[\nabla^2 L_n(\bm 0)]^{\gamma_1} \theta_1
\leq \beta (1 - e^{-\alpha_1}) \left \|[\nabla^2 L_n(\bm 0)]^{\gamma_1} \theta_1 \right\|_2
 \|\theta_1\|_2  \nonumber \\
& \leq
\begin{cases}
   \beta  e^{(1-\gamma_1)\alpha_1} (1-e^{-\alpha_1})^{2-\gamma_1}
  \|\nabla L_n(\bm 0)\|_2  \|\theta_1\|_2
  & \text{ if } \gamma_1 \leq 1; \\
  \beta \min \left( \nu^{\gamma_1 - 1}, \,
 \nu^{\gamma_1} e^{-\alpha_1}(1- e^{-\alpha_1})
 \right)  (1-e^{-\alpha_1}) \|\nabla L_n(\bm 0)\|_2 \|\theta_1\|_2
 & \text{ if } \gamma_1 > 1
\end{cases} \nonumber \\
& =  \frac{C_1}{15} \beta e^{-\alpha_1}
(e^{\alpha_1} - 1)^{\max(2-\gamma_1, 1)} \|\nabla L_n(\bm 0)\|_2 \|\theta_1\|_2   \, ,
\label{eq:g1_bd_thm3}
\end{align}
which can be upper bounded by
\begin{equation}
  (c_0 - 1)^{1/2} \frac{(e^{-t_1} - e^{-t_{2}})}{(1-e^{-t_{2}})} \|\theta_1\|_2
  = \frac{(e^{-t_1} - e^{-t_{2}})}{5(1-e^{-t_{2}})} \|\theta_1\|_2  \, ,
\end{equation}
if we choose $c_0 = 26/25$,
because \eqref{eq:step_size_condition_A}, where we have used the fact that
$e^{t_1} (e^{-t_1} - e^{-t_{2}}) (1-e^{-t_{2}})^{-1} \geq 3^{-1}$.

When $k \geq 2$, we next verify \eqref{eq:ineq_induction_step_to_check} when the bound in
\eqref{eq:grad_bound_theta_star_finite} holds for $\|g_k\|_2$ and
inequality \eqref{eq:ineq_induction_step_to_check} holds for $\|g_{k-1}\|_2$.
First using \eqref{eq:thetak_iter_difference}, we have that
\begin{align*}
  \|\theta_{k-1}\|_2 & \leq \|\theta_k\|_2 + \frac{1-e^{-t_{k}}}{1-e^{-t_{k-1}}}
  \left( e^{-t_k} + \frac{e^{-t_{k-1}}-e^{-t_{k}}}{1-e^{-t_{k-1}}}\right)^{-1} \|g_{k-1}\|_2 \nonumber
  \\
  & \leq \|\theta_k\|_2 +
  \frac{1-e^{-t_{k}}}{1-e^{-t_{k-1}}}
  \left( e^{-t_k} + \frac{e^{-t_{k-1}}-e^{-t_{k}}}{1-e^{-t_{k-1}}}\right)^{-1}
  (c_0 - 1)^{1/2}
    \frac{(e^{-t_{k-1}} - e^{-t_{k}})}{(1-e^{-t_{k}})} \|\theta_{k-1}\|_2 \nonumber
    \\
& = \|\theta_k\|_2 + (c_0 - 1)^{1/2} \frac{1-e^{-\alpha_{k}}}{1-e^{-t_{k}}} \|\theta_{k-1}\|_2 \, ,
\end{align*}
which implies that
\begin{equation}
\label{eq:thetak_upper_bd}
  \|\theta_{k-1}\|_2 \leq \frac{\|\theta_k\|_2}
  {1 - (c_0 - 1)^{1/2} \frac{1-e^{-\alpha_{k}}}{1-e^{-t_{k}}} } \, .
\end{equation}
Using this, we have that
\begin{align*}
   \|g_k\|_2 & \leq
   \frac{c_0 \beta h(\gamma_1)e^{(2-\gamma_1)t_{k}}}{2(1 - e^{-t_{k}})^{\gamma_1-1}}
   \frac{(e^{-t_{k-1}} - e^{-t_{k}})^2}{(1-e^{-t_{k-1}})^2} \|\theta_{k-1}\|_2^2
   \\
   & \leq \frac{c_0 \beta h(\gamma_1)e^{(2-\gamma_1)t_{k}}}{2(1 - e^{-t_{k}})^{\gamma_1-1}}
   \frac{(e^{-t_{k-1}} - e^{-t_{k}})^2}{(1-e^{-t_{k-1}})^2} \|\theta_{k-1}\|_2
    \frac{\|\theta_k\|_2}
  {1 - (c_0 - 1)^{1/2} \frac{1-e^{-\alpha_{k}}}{1-e^{-t_{k}}} } \\
    & \leq (c_0 - 1)^{1/2}
    \frac{(e^{-t_k} - e^{-t_{k+1}})}{(1-e^{-t_{k+1}})} \|\theta_k\|_2 \, ,
\end{align*}
provided that
\begin{equation}
\label{eq:thm3_tmp5}
  12 \sqrt{2}  \beta h(\gamma_1)
\frac{c_0(c_0 - 1)^{-1/2}}{1 - (c_0 - 1)^{1/2} \frac{1-e^{-\alpha_{k}}}{1-e^{-t_{k}}}}
  \frac{e^{t_k}(e^{\alpha_k} - 1)}{(e^{t_k} - 1)^{\gamma_1}}
   \|\theta_{k-1}\|_2 \le 1
\end{equation}
for any $k \geq 2$,
where we have used the fact that
\begin{equation*}
  \frac{(1-e^{-t_{k+1}})(1-e^{-t_{k}})(e^{\alpha_k} - 1)}
  {(1-e^{-t_{k-1}})^2(1 - e^{-\alpha_{k+1}})}
  \leq 24 \sqrt{2} \, ,
\end{equation*}
because $\alpha_{k+1} \geq \alpha_k / 2$,
$\alpha_k \leq 2 \alpha_{k-1}$, and $\alpha_{k-1} \leq \ln(2)$
for any $k \geq 1$ by assumption.
Note that \eqref{eq:thm3_tmp5} can be ensured by the
second inequality in \eqref{eq:step_size_conditions} if we choose $c_0 = 26/25$.

It remains to check Assumption (A1) through bounding
$(\theta_{k+1} - \theta_k)^\top [\nabla^2 L_n(\theta)]^{\gamma_2}
 (\theta_{k+1} - \theta_k)$, which can be achieved through
similar arguments used in the derivations of \eqref{eq:deri_grad}
ignoring the term $\beta (1 - e^{-t_{k+1}})$.
Recall that for any $0 \leq \gamma_2 \leq 2$
\begin{equation}
\label{eq:eigen_bd_gamma2}
\lambda_{\max}(H_{k+1}^{-1} [\nabla^2 L_n(\theta_k)]^{\gamma_2}
H_{k+1}^{-1}) \leq  \sup_{\lambda: \lambda \geq 0}
   \frac{\lambda^{\gamma_2}}{((1-e^{-t_{k+1}})\lambda + e^{-t_{k+1}})^2}
   = \frac{h(\gamma_2)e^{(2-\gamma_2)t_{k+1}}}{4 (1 - e^{-t_{k+1}})^{\gamma_2}}
   \, .
\end{equation}
Then replacing $\lambda_{\max}(J_{k+1})$ in \eqref{eq:deri_grad} with
the above bound,
we obtain that for any $k \geq 1$
\begin{align*}
  & (\theta_{k+1} - \theta_k)^\top [\nabla^2 L_n(\theta_k)]^{\gamma_2}
   (\theta_{k+1} - \theta_k) \\
     &\leq
     \frac{h(\gamma_2)e^{(2-\gamma_2)t_{k+1}}}{2(1 - e^{-t_{k+1}})^{\gamma_2}}
     \left(\frac{(1-e^{-t_{k+1}})^2}{(1 - e^{-t_k})^2}  \|g_k\|_2^2
     +\frac{(e^{-t_k} - e^{-t_{k+1}})^2}{(1-e^{-t_k})^2} \|\theta_k\|_2^2 \right) \\
  &\leq  \frac{h(\gamma_2) e^{(2-\gamma_2)t_{k+1}} }{(1 - e^{-t_{k+1}})^{\gamma_2}}
 \frac{c_0 (1 - e^{-\alpha_{k+1}})^2\|\theta_k\|_2^2}{(e^{t_k} -1)^2} \leq
  \frac{h(\gamma_2)}{4\beta^2 h^2(\gamma_1)}
\leq \frac{1}{\beta^2}\, ,
\end{align*}
provided that
\begin{equation}
\label{eq:thm3_tmp6}
   6 \sqrt{c_0} \beta h(\gamma_1)
   \frac{e^{t_{k+1}}(e^{\alpha_{k+1}} - 1)}
   {(e^{t_{k+1}} - 1)^{1 + \gamma_2 / 2}}
   \|\theta_k\|_2
   \leq 1
\end{equation}
for any $k \geq 1$,  which can be ensured by the
second inequality in \eqref{eq:step_size_conditions}. Here we have used
the fact that
$(1-e^{-t_{k+1}}) / (1-e^{-t_k}) \leq 3$
if $\alpha_{k+1} \leq 2 \alpha_k; k \geq 2$.

Moreover, when $k = 1$, using the first equation in \eqref{eq:thetak_iter_difference}
and the eigenvalue bound in \eqref{eq:eigen_bd_gamma2},
we obtain that
\begin{align*}
  & (\theta_{k+1} - \theta_k)^\top [\nabla^2 L_n(\theta_k)]^{\gamma_2}
   (\theta_{k+1} - \theta_k)
      =  \theta_1^\top [\nabla^2 L_n(\bm 0)]^{\gamma_2}
   \theta_1 \\
   & = (1-e^{-\alpha_1})^2 (\nabla L_n(\bm 0))^\top H_1^{-1}
   [\nabla^2 L_n(\bm 0)]^{\gamma_2} H_1^{-1} \nabla L_n(\bm 0)
   \leq \frac{h(\gamma_2)e^{(2-\gamma_2)\alpha_1}}{4 (1 - e^{-\alpha_1})^{\gamma_2}}
   (1-e^{-\alpha_1})^2 \|\nabla L_n(\bm 0)\|_2^2  \\
   & = 4^{-1} h(\gamma_2) (e^{\alpha_1} - 1)^{2-\gamma_2} \|\nabla L_n(\bm 0)\|_2^2
   \leq \frac{1}{\beta^2} \, ,
\end{align*}
provided that $\beta (e^{\alpha_1} - 1)^{1-\gamma_2/2}
   \|\nabla L_n(\bm 0)\|_2 \leq 1$,
which can be ensured by the first condition in
\eqref{eq:step_size_conditions}. This completes the proof of
\eqref{eq:grad_bound_theta_star_finite}.
Finally,
the bound in \eqref{eq:grad_bound_unified}
follows from
\eqref{eq:ineq_induction_step_to_check} and Lemma \ref{lm:thetak_theta_true_bd},
\begin{align*}
 \|g_k\|_2 & \leq  (c_0 - 1)^{1/2}
   \frac{(e^{-t_k} - e^{-t_{k+1}})}{(1-e^{-t_k})} \|\theta_k\|_2
   \leq  \frac{e^{-t_k} - e^{-t_{k} - 2\alpha_{k}}}{5(1-e^{-t_k})} \|\theta_k\|_2
   \leq \frac{2(1 - e^{-\alpha_{k}})}{5(e^{t_k}-1)} \|\theta_k\|_2  \\
   & \leq \frac{(1 - e^{-\alpha_{k}})}{2(e^{t_k}-1)} \|\theta(t_k)\|_2 \, ,
\end{align*}
where  we have used the fact that $\alpha_{k+1} \leq 2 \alpha_k; k \geq 1$.
This completes the
 proof of Theorem \ref{thm:newton}.

\vskip .1in
\noindent
\textbf{Proof of Corollary \ref{cor:newton}. }
By using \eqref{eq:theta1_newton}, we have that
$\|\theta_1\|_2  =  (1 - e^{-\alpha_1}) \|H_1^{-1} \nabla L_n(\bm 0)\|_2
\leq (e^{\alpha_1} - 1)\|\nabla L_n(\bm 0)\|_2$.
Using this and \eqref{eq:g1_bd_thm3},
it follows that,
\begin{align*}
  \|g_1\|_2 & \leq
\frac{C_1}{15}  \beta e^{-\alpha_1}
(e^{\alpha_1} - 1)^{\max(1,2-\gamma_1)} \|\nabla L_n(\bm 0)\|_2 \|\theta_1\|_2
\leq \frac{C_1}{15} \beta
(e^{\alpha_1} - 1)^{\max(2,3-\gamma_1)} \|\nabla L_n(\bm 0)\|_2^2  \, ,
\end{align*}
which proves the first bound in \eqref{eq:grad_bound_unified_cor}.

Next, we turn to the proof of the second bound in \eqref{eq:grad_bound_unified_cor}.
Using \eqref{eq:grad_bound_theta_star_finite} with $c_0 = 26/25$,
\eqref{eq:thetak_theta_true_bd} in
Lemma \ref{lm:thetak_theta_true_bd}, and
the fact that $(1-e^{-t_{k+1}}) / (1-e^{-t_k}) \leq 3$
when $\alpha_{k+1} \leq 2 \alpha_k; k \geq 1$, we obtain that
\begin{align*}
 \|g_{k+1}\|_2 & \leq
   \frac{c_0 \beta h(\gamma_1)e^{(2-\gamma_1)t_{k+1}}}{2(1 - e^{-t_{k+1}})^{\gamma_1-1}}
   \frac{(e^{-t_k} - e^{-t_{k+1}})^2}{(1-e^{-t_k})^2} \|\theta_k\|_2^2 \\
   & \leq \frac{52 \beta e^{(2-\gamma_1)t_{k+1}}}{25(1 - e^{-t_{k+1}})^{\gamma_1-1}}
   \frac{3^2 (e^{-t_k} - e^{-t_{k+1}})^2}{(1-e^{-t_{k+1}})^2}
   \left(\frac{5}{4} \right)^2 \|\theta(t_k)\|_2^2 \\
   & \leq \frac{30 \beta e^{-\gamma_1 t_{k+1}}(e^{\alpha_{k+1}} - 1)^2}{(1-e^{-t_{k+1}})^{\gamma_1-1}}
    \frac{\|\theta(t_k)\|_2^2 }{(1-e^{-t_{k+1}})^2} \\
  & \leq \frac{30 \beta e^{-\gamma_1 t_{k+1}}(e^{\alpha_{k+1}} - 1)^2}{(1-e^{-t_{k+1}})^{\gamma_1-1}}
    \left(\|\theta(t_{k})\|_2 + \|\nabla L_n(\bm 0)\|_2\right)^2
\end{align*}
for any $k \geq 1$, where the last inequality uses
Lemma \ref{lm:thm3}.
This proves the second bound in \eqref{eq:grad_bound_unified_cor}.
This completes the proof of Corollary \ref{cor:newton}.

\vskip .1in
\noindent
\textbf{Proof of Theorem \ref{thm:global_bd_newton}. }
Using \eqref{eq:thm_fn_bound1} and \eqref{eq:ineq_induction_step_to_check},
we have that
\begin{align}
\label{eq:thm4_seg1}
  \sup_{t \in [0, t_{1}]}
   \left\{ f_t(\tilde \theta(t)) - f_t(\theta(t)) \right\} & \leq
    \max \left((e^{\alpha_1}  - 1)^2
  \|\nabla L_n(\bm 0)\|_2^2, \,   \|\theta_1\|_2^2
  \right) \, .
\end{align}

Using \eqref{eq:ineq_induction_step_to_check}, we have that
\begin{align*}
   e^{t_{k+1}} \left( \frac{1-e^{-t_{k+1}}}{1 - e^{-t_k}} \right)^2  \|g_k\|_2^2
   & \leq e^{t_{k+1}} \left( \frac{1-e^{-t_{k+1}}}{1 - e^{-t_k}} \right)^2
   \left( (c_0 - 1)^{1/2}
    \frac{(e^{-t_k} - e^{-t_{k+1}})}{(1-e^{-t_{k+1}})} \|\theta_k\|_2  \right)^2 \\
   & \leq (e^{-t_k} - e^{-t_{k+1}})^2 \frac{e^{t_{k+1}} \|\theta_k\|_2^2 }{(1 - e^{-t_{k}})^2}
\end{align*}
and
\begin{align*}
  e^{t_{k+1}}  \|g_{k+1}\|_2^2 &  \leq
  e^{t_{k+1}}  \left( (c_0 - 1)^{1/2}
    \frac{(e^{-t_{k+1}} - e^{-t_{k+2}})}{(1-e^{-t_{k+2}})} \|\theta_{k+1}\|_2  \right)^2 \\
  & \leq 4^{-1} \frac{(e^{\alpha_{k+2}} - 1)^2}{(e^{\alpha_{k+1}} - 1)^2}
  e^{-2\alpha_{k+2} + \alpha_{k+1}}
  (e^{-t_k} - e^{-t_{k+1}})^2
 \frac{e^{t_k} \|\theta_{k+1}\|_2^2 }{(1 - e^{-t_{k+1}})^2}
 \\
 & \leq (e^{-t_k} - e^{-t_{k+1}})^2
 \frac{e^{t_k} \|\theta_{k+1}\|_2^2 }{(1 - e^{-t_{k+1}})^2} \, ,
\end{align*}
where we have used the fact that $\alpha_{k+2} \leq 2 \alpha_{k+1}$ and
\begin{equation*}
  \frac{(e^{\alpha_{k+2}} - 1)^2}{(e^{\alpha_{k+1}} - 1)^2}
  e^{-2\alpha_{k+2} + \alpha_{k+1}}  \leq
  (e^{\alpha_{k+1}} + 1)^2 e^{-3\alpha_{k+1}} \leq 4 \, .
\end{equation*}
Combining, we obtain that
\begin{equation*}
   e^{t_{k+1}}
   \max \left\{\left( \frac{1-e^{-t_{k+1}}}{1 - e^{-t_k}} \right)^2  \|g_k\|_2^2, \,
  \|g_{k+1}\|_2^2 \right\}
  \leq (e^{-t_k} - e^{-t_{k+1}})^2 \max \left\{
   \frac{e^{t_{k+1}} \|\theta_k\|_2^2 }{(1 - e^{-t_{k}})^2} , \,
   \frac{e^{t_k} \|\theta_{k+1}\|_2^2 }{(1 - e^{-t_{k+1}})^2} \right\}
\end{equation*}
Combining this with \eqref{eq:thm_fn_bound2} in Theorem \ref{thm:uniform_bd_general},
it follows that
\begin{align*}
  & \sup_{t \in [t_k, t_{k+1}]}
   \left\{ f_t(\tilde \theta(t)) - f_t(\theta(t)) \right\} \leq
   2 (e^{-t_k} - e^{-t_{k+1}})^2 \max \left\{
   \frac{e^{t_{k+1}} \|\theta_k\|_2^2 }{(1 - e^{-t_{k}})^2} , \,
   \frac{e^{t_k} \|\theta_{k+1}\|_2^2 }{(1 - e^{-t_{k+1}})^2} \right\} \, .
\end{align*}
Therefore,
\begin{align}
  \max_{1 \leq k \leq N - 1} & \sup_{t \in [t_k, t_{k+1}]}
   \left\{ f_t(\tilde \theta(t)) - f_t(\theta(t)) \right\} \nonumber \\
    & \leq  \max_{1 \leq k \leq N - 1} 2
  (e^{-t_k} - e^{-t_{k+1}})^2 \max \left\{
   \frac{e^{t_{k+1}} \|\theta_k\|_2^2 }{(1 - e^{-t_{k}})^2} , \,
   \frac{e^{t_k} \|\theta_{k+1}\|_2^2 }{(1 - e^{-t_{k+1}})^2} \right\} \nonumber \\
   & \leq 2 \max_{1 \leq k \leq N} \left\{
   \frac{e^{-t_{k}} \|\theta_k\|_2^2}{(1 - e^{-t_{k}})^2}
   \max\left( e^{-\alpha_{k+1}}(e^{\alpha_{k+1}} - 1)^2,
    e^{-\alpha_{k}}(e^{\alpha_{k}} - 1)^2\right)
   \right\} \nonumber \\
   & \leq 8
   \max_{1 \leq k \leq N} \left\{
   \frac{e^{-t_{k}} (e^{\alpha_{k+1}} - 1)^2\|\theta_k\|_2^2 }{(1 - e^{-t_{k}})^2}
   \right\} \, .
   \label{eq:thm4_tmp8}
\end{align}
Combining this with \eqref{eq:thm4_seg1},
we have that
\begin{align*}
   \sup_{0 \leq t \leq t_{\max}} &
  \left\{ f_t(\tilde \theta(t)) - f_t(\theta(t)) \right\}
   \leq
  \sup_{0 \leq t \leq t_{N}}   \left\{ f_t(\tilde \theta(t)) - f_t(\theta(t)) \right\}
  \nonumber  \\
  & \le 8 \max \left\{(e^{\alpha_1} - 1)^2 \|\nabla L_n(\bm 0)\|_2^2, \,
\max_{1 \leq k \leq N} \left(
   \frac{e^{-t_{k}} (e^{\alpha_{k+1}} - 1)^2\|\theta_k\|_2^2}{(1 - e^{-t_{k}})^2}
   \right)
  \right\} \, ,
\end{align*}
if $t_{N-1} \leq t_{\max} \leq t_N$ for some $N \geq 1$.
This proves \eqref{eq:newton_uniform_general_bd2}.

Lastly, when $0 < t_N \leq t_{\max}$,
using \eqref{eq:ineq_induction_step_to_check}, we obtain that
\begin{align*}
   \frac{e^{t_N}\|g_N\|_2^2}{1 - e^{-t_N}} & \leq
    \frac{e^{t_N}}{1 - e^{-t_N}} \left(  (c_0 - 1)^{1/2}
    \frac{(e^{-t_N} - e^{-t_{N+1}})}{(1-e^{-t_{N+1}})} \|\theta_N\|_2  \right)^2 \\
    & \leq  \frac{(1 - e^{-\alpha_{N+1}})^2}{4(1-e^{-t_{N+1}})^2(e^{t_{N}} - 1)}
     \|\theta_N\|_2^2 \leq \frac{\|\theta_N\|_2^2}{4(e^{t_{N}} - 1)} \, .
\end{align*}

Combining this with
\eqref{eq:thm_fn_bound3} in
 Theorem \ref{thm:uniform_bd_general}, we obtain that when $t_N \leq t_{\max}$,
 \begin{align}
  & \sup_{t_N < t \leq t_{\max}}
  \left\{ f_t(\tilde \theta(t)) - f_t(\theta(t)) \right\} \leq
   \frac{\|\theta_N\|_2^2}{4(e^{t_{N}} - 1)}
   + \frac{3\|\theta(t_{\max})\|_2^2}{2(e^{t_N}-1)} \, .
   \label{eq:thm4_tmp9}
\end{align}
Combining \eqref{eq:thm4_tmp8} and \eqref{eq:thm4_tmp9}, we obtain that
when $t_{\max} = \infty$
\begin{align*}
    \sup_{0 \leq t \leq t_{\max}} &  \left\{ f_t(\tilde \theta(t)) - f_t(\theta(t)) \right\}
  \leq
  \max\Bigg\{ 8 (e^{\alpha_{1}} - 1)^2 \|\nabla L_n(\bm 0)\|_2^2, \, \\
  & 8 \max_{1 \leq k \leq N-1} e^{-t_{k}}
  \left( \frac{e^{\alpha_{k+1}} - 1}{1 - e^{-t_{k}}} \right)^2
     \|\theta_k\|_2^2, \,
\frac{2\max(\|\theta(t_{\max})\|_2^2, \|\theta_N\|_2^2)}{(e^{t_N}-1)} \Bigg\} \, ,
\end{align*}
which implies  \eqref{eq:newton_uniform_general_bd1}.
This completes the proof of Theorem \ref{thm:global_bd_newton}.

\vskip .1in
Next, we present a supporting lemma to be used in the proof of Theorem \ref{thm:newton_global_bd}.
\begin{lem}
\label{lm:thetak_theta_true_bd}
  Under the assumptions in Theorem \ref{thm:newton}, we have that
  \begin{align}
   & \|\theta_1\|_2 \leq (e^{\alpha_1} - 1) \| \nabla L_n(\bm 0)\|_2
   \text{ and }
   \label{eq:theta1_bd} \\
   &  \frac{1}{1 +  (c_0 - 1)^{1/2}}
  \leq \frac{\|\theta_k\|_2}{\|\theta(t_{k})\|_2}
  \leq \frac{1}{1 -   (c_0 - 1)^{1/2}} \, ,
  \label{eq:thetak_theta_true_bd}
  \end{align}
  where $c_0 = 26/25$.
\end{lem}
\noindent
\textbf{Proof of Lemma \ref{lm:thetak_theta_true_bd}. }
For the first inequality, using \eqref{eq:theta1_newton}, we have that
\begin{align*}
  \|\theta_1\|_2 &= \|(1 - e^{-\alpha_1}) H_1^{-1} \nabla L_n(\bm 0)\|_2
  \leq \lambda_{\max}(H_1^{-1}) (1 - e^{-\alpha_1}) \| \nabla L_n(\bm 0)\|_2  \\
  & \leq e^{\alpha_1}(1 - e^{-\alpha_1}) \| \nabla L_n(\bm 0)\|_2
  = (e^{\alpha_1} - 1) \| \nabla L_n(\bm 0)\|_2,
\end{align*}
which proves \eqref{eq:theta1_bd}.

We next prove \eqref{eq:thetak_theta_true_bd}.
Using \eqref{eq:ineq_induction_step_to_check} and \eqref{eq:str_con_prop2},
we have that  $\|\theta_k - \theta(t_k)\|_2
\leq e^{t_k} \|g_k\|_2$ and
\begin{align*}
  \|\theta_k\|_2 & \leq
  \|\theta(t_k)\|_2 +
  \|\theta(t_k) - \theta_k\|_2
  \leq \|\theta(t_k)\|_2 + e^{t_k} \|g_k\|_2 \\
  & \leq \|\theta(t_k)\|_2 +
  (c_0 - 1)^{1/2} \frac{(1 - e^{-\alpha_{k+1}})}{(1-e^{-t_{k+1}})} \|\theta_k\|_2 \, ,
\end{align*}
which implies that
\begin{equation*}
  \|\theta_k\|_2 \leq
  \frac{\|\theta(t_k)\|_2}
  {1 - (c_0 - 1)^{1/2} \frac{(1 - e^{-\alpha_{k+1}})}{(1-e^{-t_{k+1}})}} \, .
\end{equation*}
Similarly, we also have
\begin{equation*}
  \|\theta(t_{k})\|_2 \leq \|\theta_k\|_2 +
  \|\theta(t_{k}) - \theta_k\|_2  \leq
  \|\theta_k\|_2 +
   (c_0 - 1)^{1/2} \frac{(1 - e^{-\alpha_{k+1}})}{(1-e^{-t_{k+1}})} \|\theta_k\|_2 \, .
\end{equation*}
Combining, we obtain that
\begin{equation*}
  \frac{1}{1 + (c_0 - 1)^{1/2} \frac{(1 - e^{-\alpha_{k+1}})}{(1-e^{-t_{k+1}})}}
  \leq \frac{\|\theta_k\|_2}{\|\theta(t_{k})\|_2}
  \leq \frac{1}{1 -  (c_0 - 1)^{1/2} \frac{(1 - e^{-\alpha_{k+1}})}{(1-e^{-t_{k+1}})}} \, ,
\end{equation*}
which implies \eqref{eq:thetak_theta_true_bd}.
This completes the proof of Lemma \ref{lm:thetak_theta_true_bd}.

\vskip 0.1in
\noindent
\textbf{Proof of Theorem \ref{thm:newton_global_bd}. }
We first verify $\alpha_{1}$ and $\alpha_{k+1}$ satisfy
the conditions in \eqref{eq:step_size_conditions}.
By the definitions of $\alpha_{1}$ and $\alpha_{k+1}$,
we only need to prove that $\alpha_{k+1} \geq \alpha_k / 2$.

If $\alpha_{k+1} = \alpha_{\max} = 1/10$ or $2\alpha_k$, then trivially
$\alpha_{k+1} \geq \alpha_k / 2$.
Hence, we only need to consider the case
$\alpha_{k+1} = A_k \text{ or } B_k$, where
\begin{align*}
  A_k & := \ln \left( 1 + \frac{e^{t_k/2}(e^{\alpha_1} - 1)\|\nabla L_n(\bm 0)\|_2(1-e^{-t_k})}
  {\|\theta_k\|_2}
  \right) \, , \\
  B_k & :=  \ln\left( 1 + \left(C_2\beta e^{t_{k}}
  \max\left((e^{t_{k}} - 1)^{-\gamma_1},
  (e^{t_{k}} - 1)^{-1-\gamma_2/2}\right)
  \|\theta_k\|_2\right)^{-1} \right) \, .
\end{align*}
It is easy to check that $A_k \leq B_k$ if and only if
\begin{equation}
\label{eq:cor2_tmp1}
  C_2 \beta \|\nabla L_n(\bm 0)\|_2 (e^{\alpha_1} - 1)
 e^{t_k/2} \max\left((e^{t_{k}} - 1)^{1-\gamma_1},
  (e^{t_{k}} - 1)^{-\gamma_2/2}\right)  \leq 1 \, .
\end{equation}
We first show that $A_k \leq B_k$ when $e^{t_k} \leq 2$.
Equivalently, we need to show that the above inequality \eqref{eq:cor2_tmp1}
holds if $e^{t_k} \leq 2$.
To this end,
we consider two cases: (i) $\gamma_1 \geq 1$; and (ii)
$\gamma_1 < 1$.
It is easy to see that for any $s \geq 0$,
 function $x / (x^2 - 1)^s$ with $e^{\alpha_1 / 2} \leq x \leq \sqrt{2}$
achieves its maximum at the two boundary points, that is,
\begin{equation}
\label{eq:cor_tmp_ineq}
  x / (x^2 - 1)^s \leq \max\left(\sqrt{2}, e^{\alpha_1 / 2} / (e^{\alpha_1} - 1)^s\right) \, .
\end{equation}
For case (i), note that when $e^{t_k} \leq 2$,
\begin{align*}
 & e^{t_k/2} \max\left((e^{t_{k}} - 1)^{1-\gamma_1},
  (e^{t_{k}} - 1)^{-\gamma_2/2}\right)
  = \frac{e^{t_k/2}}{(e^{t_{k}} - 1)^{\max(\gamma_1 - 1, \gamma_2/2})} \\
  & \leq \max\left(\sqrt{2}, e^{\alpha_1 / 2} / (e^{\alpha_1} - 1)^{\max(\gamma_1 - 1, \gamma_2/2)}
  \right)
\end{align*}
where we have used \eqref{eq:cor_tmp_ineq}.
 Hence, inequality \eqref{eq:cor2_tmp1} holds if
\begin{equation*}
  C_2 e^{\alpha_1 / 2}  \beta
 \|\nabla L_n(\bm 0)\|_2
  (e^{\alpha_1} - 1)^{\min(2-\gamma_1, 1-\gamma_2 / 2)}  \leq 1
  \text{ and } \sqrt{2} C_2 \beta
 \|\nabla L_n(\bm 0)\|_2
  (e^{\alpha_1} - 1) \leq 1 \, ,
\end{equation*}
both of which can be ensured by \eqref{eq:newton_stepsize_cond_ideal_alpha1}.

Similarly, for case (ii), note that when $e^{t_k} \leq 2$,
\begin{align*}
 & e^{t_k/2} \max\left((e^{t_{k}} - 1)^{1-\gamma_1},
  (e^{t_{k}} - 1)^{-\gamma_2/2}\right)
  = \frac{e^{t_k/2}}{(e^{t_{k}} - 1)^{\gamma_2/2}}
  \leq \max\left(\sqrt{2}, e^{\alpha_1 / 2} / (e^{\alpha_1} - 1)^{\gamma_2/2}
  \right) \, ,
\end{align*}
where we have used \eqref{eq:cor_tmp_ineq}. Hence, inequality \eqref{eq:cor2_tmp1} holds if
\begin{equation*}
  C_2 e^{\alpha_1 / 2} \beta
 \|\nabla L_n(\bm 0)\|_2
  (e^{\alpha_1} - 1)^{1-\gamma_2 / 2}  \leq 1
  \text{ and } \sqrt{2} C_2 \beta
 \|\nabla L_n(\bm 0)\|_2
  (e^{\alpha_1} - 1) \leq 1 \, ,
\end{equation*}
both of which can be ensured by \eqref{eq:newton_stepsize_cond_ideal_alpha1}.
This completes the proof that $A_k \leq B_k$ when $e^{t_k} \leq 2$.

Since $e^{\alpha_1} \leq 2$, we only need to consider the case $\alpha_{2} = A_1$ when $k = 1$.
To show that $e^{\alpha_{2}} - 1 \geq e^{\alpha_1 / 2}$, we note that
\begin{align*}
\frac{e^{\alpha_{2}} - 1}{e^{\alpha_1 / 2} - 1} & =
  \frac{e^{A_1} - 1}{e^{\alpha_1 / 2} - 1} =
  \frac{e^{t_1 / 2}(e^{\alpha_1} - 1)\|\nabla L_n(\bm 0)\|_2(1-e^{-t_1})}
  {\|\theta_1\|_2 (e^{\alpha_1 / 2} - 1)}  \\
  & \geq \frac{e^{\alpha_1 / 2}(e^{\alpha_1} - 1)\|\nabla L_n(\bm 0)\|_2(1-e^{-t_1})}
  { (e^{\alpha_1} - 1)\|\nabla L_n(\bm 0)\|_2 (e^{\alpha_1 / 2} - 1)}
  = e^{-\alpha_1 / 2}  (e^{\alpha_1 / 2} + 1) > 1
  \, ,
\end{align*}
where we have used \eqref{eq:theta1_bd} in Lemma \ref{lm:thetak_theta_true_bd}.
This proves that $\alpha_2 \geq \alpha_1 / 2$.

For $k \geq 2$,
we consider two cases
(i) $\alpha_{k+1} = A_k$;
and (ii) $\alpha_{k+1} = B_k$.
For case (i), using the fact that $\alpha_k \leq A_{k-1}$, we have
\begin{align}
 \frac{e^{\alpha_{k+1}} - 1}{e^{\alpha_{k}/2} - 1}  & =
  (e^{\alpha_{k}/2} + 1) \frac{e^{\alpha_{k+1}} - 1}{e^{\alpha_{k}} - 1}
  \geq (e^{\alpha_{k}/2} + 1) \frac{e^{A_k} - 1}{e^{A_{k-1}} - 1}  \\
  & = (e^{\alpha_k / 2} + 1) e^{\alpha_k / 2}
  \frac{(e^{\alpha_1} - 1)\|\nabla L_n(\bm 0)\|_2(1-e^{-t_k})}
  {\|\theta_k\|_2}
  \frac{\|\theta_{k-1}\|_2}
  {(e^{\alpha_1} - 1)\|\nabla L_n(\bm 0)\|_2(1-e^{-t_{k-1}})} \nonumber \\
  & = (e^{-\alpha_k / 2} + 1)
   \frac{(e^{t_k} - 1)\|\theta_{k-1}\|_2}
  {(e^{t_{k-1}} - 1)\|\theta_{k}\|_2} \, .
  \label{eq:cor2_tmp2}
\end{align}
Applying Lemma \ref{lm:thetak_theta_true_bd}, we have that
\begin{align}
   \frac{(e^{t_k} - 1)\|\theta_{k-1}\|_2}
  {(e^{t_{k-1}} - 1)\|\theta_{k}\|_2} & \geq
   \frac{(e^{t_k} - 1)\|\theta(t_{k-1})\|_2}
  {(e^{t_{k-1}} - 1)\|\theta(t_{k})\|_2}
  \frac{1 -  (c_0 - 1)^{1/2} \frac{(1 - e^{-\alpha_{k+1}})}{(1-e^{-t_{k+1}})}}
  {1 +  (c_0 - 1)^{1/2} \frac{(1 - e^{-\alpha_{k+1}})}{(1-e^{-t_{k+1}})}}  \nonumber \\
  & \geq \frac{1 -  (c_0 - 1)^{1/2} \frac{(1 - e^{-\alpha_{k+1}})}{(1-e^{-t_{k+1}})}}
  {1 +  (c_0 - 1)^{1/2} \frac{(1 - e^{-\alpha_{k+1}})}{(1-e^{-t_{k+1}})}}
  \geq \frac{1 -  (c_0 - 1)^{1/2}}
  {1 +  (c_0 - 1)^{1/2}}\, , \label{eq:cor_ratio_bd}
\end{align}
where the last inequality uses part (ii) of Corollary \ref{cor:solution_prop}.
Combining this with \eqref{eq:cor2_tmp2}, we obtain that
\begin{equation}
  \frac{e^{\alpha_{k+1}} - 1}{e^{\alpha_{k}/2} - 1} \geq
  (e^{-\alpha_k / 2} + 1) \frac{1 -  (c_0 - 1)^{1/2}}
  {1 +  (c_0 - 1)^{1/2}} \geq 1 \, ,
\end{equation}
because $\alpha_k \leq 10^{-1}$ and $c_0 = 26 / 25$.
This proves case (i).


For case (ii),
we have $\alpha_{k+1} = B_k$ and $\alpha_k \leq B_{k-1}$.
Since we have shown that $\alpha_{k+1} = A_k$ when $e^{t_k} > 2$,
we must have that $e^{t_k} \leq 2$ and $e^{t_{k-1}} > 2 e^{-\alpha_k}$.
Using these, we have that
\begin{align*}
\frac{e^{\alpha_{k+1}} - 1}{e^{\alpha_{k}/2} - 1}  & =
  (e^{\alpha_{k}/2} + 1) \frac{e^{\alpha_{k+1}} - 1}{e^{\alpha_{k}} - 1}
  \geq
  (e^{\alpha_{k}/2} + 1) \frac{e^{B_k} - 1}{e^{B_{k-1}} - 1} \\
  &
  = (e^{\alpha_k / 2} + 1) \frac{e^{t_{k-1}}
  \max\left((e^{t_{k-1}} - 1)^{-\gamma_1},
  (e^{t_{k-1}} - 1)^{-1-\gamma_2/2}\right)
  \|\theta_{k-1}\|_2}
  {e^{t_{k}}
  \max\left((e^{t_{k}} - 1)^{-\gamma_1},
  (e^{t_{k}} - 1)^{-1-\gamma_2/2}\right)
  \|\theta_k\|_2}  \\
  & \geq  (e^{\alpha_k / 2} + 1) \frac{
  \max\left((e^{t_{k-1}} - 1)^{1-\gamma_1},
  (e^{t_{k-1}} - 1)^{-\gamma_2/2}\right)}
  {e^{\alpha_{k}} \max\left((e^{t_{k}} - 1)^{1-\gamma_1},
  (e^{t_{k}} - 1)^{-\gamma_2/2}\right)}
   \frac{1 -  (c_0 - 1)^{1/2}}
  {1 +  (c_0 - 1)^{1/2}}\, .
\end{align*}
where we have used \eqref{eq:cor_ratio_bd}.
Now if $e^{t_{k-1}} \geq 2$, then
\begin{align*}
  \frac{e^{\alpha_{k+1}} - 1}{e^{\alpha_{k}/2} - 1}
  & \geq (e^{\alpha_k / 2} + 1) \frac{
  (e^{t_{k-1}} - 1)^{\max(1-\gamma_1, \, -\gamma_2/2)}}
  {e^{\alpha_{k}}
  (e^{t_{k}} - 1)^{\max(1-\gamma_1, \, -\gamma_2/2)}}
  \frac{1 -  (c_0 - 1)^{1/2}}
  {1 +  (c_0 - 1)^{1/2}} \\
  & \geq (e^{\alpha_k / 2} + 1) \frac{(e^{t_{k-1}} - 1)}
  {e^{\alpha_{k}}
  (e^{t_{k}} - 1) } \frac{1 -  (c_0 - 1)^{1/2}}
  {1 +  (c_0 - 1)^{1/2}}
  \\
  & \geq
  \frac{(e^{\alpha_k / 2} + 1) e^{-\alpha_{k}} }
  {e^{\alpha_k} + (e^{\alpha_{k}} - 1)/(e^{t_{k-1}} - 1)}
    \frac{1 -  (c_0 - 1)^{1/2}}
  {1 +  (c_0 - 1)^{1/2}}\\
  &  \geq
  \frac{(e^{\alpha_k / 2} + 1) e^{-\alpha_{k}} }
  {2 e^{\alpha_k}  - 1}  \frac{1 -  (c_0 - 1)^{1/2}}
  {1 +  (c_0 - 1)^{1/2}}
  \geq 1 \, ,
\end{align*}
where we have used the fact that the last inequality holds when
$\alpha_k \leq 10^{-1}$ and $c_0 = 26/25$.
If $e^{t_{k-1}} < 2$, we have $e^{t_k} < 2 e^{\alpha_k}$ and
\begin{align*}
  \frac{e^{\alpha_{k+1}} - 1}{e^{\alpha_{k}/2} - 1}
  & \geq (e^{\alpha_k / 2} + 1) \frac{
  (e^{t_{k-1}} - 1)^{\min(1-\gamma_1, \, -\gamma_2/2)}}
  {e^{\alpha_{k}}
  (e^{t_{k}} - 1)^{\max(1-\gamma_1, \, -\gamma_2/2)}}
  \frac{1 -  (c_0 - 1)^{1/2}}
  {1 +  (c_0 - 1)^{1/2}} \\
  & \geq (e^{\alpha_k / 2} + 1) \frac{(2e^{-\alpha_{k}} - 1)}
  {e^{\alpha_{k}} (2e^{\alpha_{k}} - 1) }
  \frac{1 -  (c_0 - 1)^{1/2}}
  {1 +  (c_0 - 1)^{1/2}}  \geq 1 \, ,
\end{align*}
where we have used the fact that the last inequality holds when
$\alpha_k \leq 10^{-1}$ and $c_0 = 26/25$.
This completes the proof of $\alpha_{k+1} \geq \alpha_k / 2$.

Next, we show that the algorithm terminates after a finite number of steps.
We first show that $t_k$ diverges.
To this end, using \eqref{eq:theta_norm_bds} and \eqref{eq:thetak_theta_true_bd},
we have
\begin{align*}
  & \frac{e^{t_k/2}(e^{\alpha_1} - 1)\|\nabla L_n(\bm 0)\|_2(1-e^{-t_k})}
  {\|\theta_k\|_2}  \geq
  \frac{e^{t_k/2}(e^{\alpha_1} - 1)\|\nabla L_n(\bm 0)\|_2(1-e^{-t_k})}
  {\|\theta(t_k)\|_2 / (1-(c_0-1)^{1/2})}
  \\
  & \geq \frac{e^{t_k/2}(e^{\alpha_1} - 1)\|\nabla L_n(\bm 0)\|_2(1-e^{-t_k})(1-(c_0-1)^{1/2})}
  {(e^{t_k} - 1) \|\nabla L_n(\bm 0)\|_2}
  \geq 2^{-1}(e^{\alpha_1} - 1)e^{-t_k/2}  \, ,
\end{align*}
which implies that
\begin{equation}
  A_k \geq \ln(1+2^{-1}(e^{\alpha_1} - 1)e^{-t_k/2}) \geq
  \frac{(e^{\alpha_1} - 1)e^{-t_k/2}}{2+(e^{\alpha_1} - 1)e^{-t_k/2}}
  =  \frac{(e^{\alpha_1} - 1)}{2e^{t_k/2}+(e^{\alpha_1} - 1)} \, .
\end{equation}
Moreover, $B_k \geq A_k$ when $e^{t_k} < 2$,
and when $e^{t_k} \geq 2$, we have
\begin{align*}
   & e^{t_{k}} \max\left((e^{t_{k}} - 1)^{-\gamma_1},
  (e^{t_{k}} - 1)^{-1-\gamma_2/2}\right)
  \|\theta_k\|_2 \\
  &  \leq 2 e^{t_{k}} \max\left((e^{t_{k}} - 1)^{1-\gamma_1},
  (e^{t_{k}} - 1)^{-\gamma_2/2}\right)  \|\nabla L_n(\bm 0)\|_2 \\
   & \leq 2  e^{2 t_{k}}\|\nabla L_n(\bm 0)\|_2 \, ,
\end{align*}
which implies that when $e^{t_k} \geq 2$
\begin{align*}
  B_k & \geq
  \ln(1+(C_1 \beta e^{2 t_{k}}\|\nabla L_n(\bm 0)\|_2)^{-1})
  \geq \frac{(2 C_1 \beta e^{2 t_{k}}\|\nabla L_n(\bm 0)\|_2)^{-1}}
  {1+ (2 C_1 \beta e^{2 t_{k}}\|\nabla L_n(\bm 0)\|_2)^{-1}}  \\
  & = \frac{1}{1 + (2 C_1 \beta e^{2 t_{k}}\|\nabla L_n(\bm 0)\|_2)} \, .
\end{align*}
Thus,
\begin{align*}
  B_k & \geq  \min\left\{ \frac{(e^{\alpha_1} - 1)}{2e^{t_k/2}+(e^{\alpha_1} - 1)}, \,
  \frac{1}{1 + (2 C_1 \beta e^{2 t_{k}}\|\nabla L_n(\bm 0)\|_2)}
  \right\} \, .
\end{align*}
Combining we have that
\begin{equation}
  \alpha_{k+1} \geq \min\left(\alpha_{\max}, 2\alpha_k,
  \frac{(e^{\alpha_1} - 1)}{2 e^{t_k/2}+(e^{\alpha_1} - 1)} \, ,
  \frac{1}{1 + 2 C_1 \beta e^{2 t_{k}}\|\nabla L_n(\bm 0)\|_2}\right) \, .
\end{equation}
Now we prove the divergence of $t_k$ by contradiction.
Suppose that $t_k$ does not diverge. Then there must exist a constant $T$ such that
$t_k < T$ for all $k$. However, now we have
\begin{equation*}
   \alpha_{k+1}  \geq \min\left(\alpha_{\max}, 2\alpha_k,
  \frac{(e^{\alpha_1} - 1)}{2 e^{T/2}+(e^{\alpha_1} - 1)} \, ,
  \frac{1}{1 + 2 C_1 \beta e^{2 T}\|\nabla L_n(\bm 0)\|_2}\right)  \, ,
 \end{equation*}
 which implies that $\alpha_{k}$ is lower bounded by a positive constant
 when $k$ is large enough, implying that
$t_k$ should diverge. This is a contradiction.
Hence, $t_k$ diverges.

Now we are ready to show that the algorithm
must terminate after a finite number of iterations.
If $t_{\max} < \infty$, then $t_k \geq t_{\max}$ must hold
for large enough $k$ as $t_k$ diverges.
If $t_{\max} = \infty$, then we have
that $\theta(t_{\max}) = \theta^\star$ is finite by assumption.
Therefore,
the termination criterion in
\eqref{eq:termination_criterion} should also be met when $N$ is large
enough, because $t_N$ diverges and
\begin{align*}
  &  \frac{\max(\|\theta(t_{\max})\|_2^2, \|\theta_N\|_2^2)}{(e^{t_N}-1)}
  \leq \frac{2\max(\|\theta(t_{\max})\|_2^2, \|\theta(t_N)\|_2^2)}{(e^{t_N}-1)}
   \le \frac{2 \|\theta^\star\|_2^2}{(e^{t_N}-1)}
      \rightarrow 0 \text{ as } N \rightarrow \infty \, .
\end{align*}

Finally, we are ready to prove \eqref{eq:cor2_bd_global} after the algorithm is terminated.
Upon termination when $k+1 = N$, we have one of
the two conditions in \eqref{eq:termination_criterion} must hold.
If $t_N > t_{\max}$, it is easy to see that
the step sizes defined in \eqref{eq:newton_stepsize_cond_ideal_alpha1} and
\eqref{eq:newton_stepsize_cond_ideal_alphak}
satisfy all the assumptions in Theorem \ref{thm:global_bd_newton},
by using \eqref{eq:newton_uniform_general_bd2} in Theorem \ref{thm:global_bd_newton}
and the definition
of $\alpha_k$, we have that
\begin{align*}
   \sup_{0 \leq t \leq t_{\max}} &  \left\{ f_t(\tilde \theta(t)) - f_t(\theta(t)) \right\}
  \lesssim  (e^{\alpha_{1}} - 1)^2 \|\nabla L_n(\bm 0)\|_2^2 \leq \epsilon  \, .
\end{align*}
If the second inequality in \eqref{eq:termination_criterion} holds, then
\begin{align*}
  \frac{2\max(\|\theta(t_{\max})\|_2^2, \|\theta_N\|_2^2)}{(e^{t_N}-1)}
  & \leq \frac{2\|\theta(t_{\max})\|_2^2}{(e^{t_N}-1)}
  \leq  \frac{2\|\theta_N\|_2^2}{(e^{t_N}-1)}
  \frac{\|\theta(t_{\max})\|_2^2}{\|\theta_N\|_2^2} \\
  & \leq  \frac{2 \|\theta(t_{\max})\|_2^2}{\|\theta(t_N)\|_2^2}
  (e^{\alpha_{1}} - 1)^2 \|\nabla L_n(\bm 0)\|_2^2 \, .
\end{align*}
Combining this with \eqref{eq:newton_uniform_general_bd1} in Theorem \ref{thm:global_bd_newton},
we obtain that
\begin{align*}
   \sup_{0 \leq t \leq t_{\max}} &  \left\{ f_t(\tilde \theta(t)) - f_t(\theta(t)) \right\}
  \lesssim
  \frac{\|\theta(t_{\max})\|_2^2}{\|\theta(t_N)\|_2^2}
   (e^{\alpha_{1}} - 1)^2 \|\nabla L_n(\bm 0)\|_2^2 \leq \epsilon  \, .
\end{align*}
This completes the proof of Theorem \ref{thm:newton_global_bd}.


\vskip 0.1in
We next present a supporting lemma for the proof of Theorem \ref{thm:complexity_newton} and
\ref{thm:complexity_gd}.
\begin{lem}
\label{lm:compleixty}
For any $\epsilon \in (0, 1]$, let $(e^{\alpha_1} - 1)^2 = c_1^2 \epsilon$ and
$\alpha_{k+1} = \ln(1 + c_2 e^{t_k/2}(e^{\alpha_1} - 1))$, where $c_1, c_2 > 0$ are
constants and
$t_k = \sum_{i = 1}^k \alpha_i$. Define
$k^\star := \max\{k: t_k \leq \ln(\epsilon^{-1}) \}$.
Then,
\begin{equation}
\label{eq:kstar_bd_newton}
k^\star < \frac{1+c_1 c_2+\sqrt{1+c_1 c_2}}
{c_1 c_2\sqrt{\epsilon} \sqrt{1+c_1\sqrt{\epsilon}}}, \,
\sum_{k=1}^{k^\star}e^{t_k}(t_k+1)
\leq
\frac{(1+2c_1c_2+\sqrt{1+c_1c_2})(1 + \ln (\epsilon^{-1}))}{c_1c_2 \epsilon}  \, .
\end{equation}
\end{lem}

\vskip 0.1in
\noindent
\textbf{Proof of Lemma \ref{lm:compleixty}. }
It is easy to see that $\alpha_k$ is strictly increasing for $k \geq 2$
and by definition of $k^\star$, we have that
\begin{equation}
\alpha_{k+1} \leq \ln (1+c_1 c_2) \text{ for any } 1 \leq k \leq k^\star
\, ,
\end{equation}
which implies that
$e^{\alpha_{k+1}/2} \leq (1+c_1 c_2)^{1/2}$ for any $1 \leq k \leq k^\star$.
Moreover, since $e^{\alpha_{k+1}}=1+c_1 c_2 \sqrt{\epsilon}e^{t_k/2}$, we have that
\begin{align*}
e^{-t_k/2}-e^{-t_{k+1}/2} & =  \frac{e^{\alpha_{k+1}/2}-1}{e^{t_{k+1}/2}}\
= \frac{e^{-t_{k+1}/2} (e^{\alpha_{k+1}}-1)}{e^{\alpha_{k+1}/2}+1}
=  \frac{e^{-\alpha_{k+1}/2} c_1 c_2 \sqrt{\epsilon}}{e^{\alpha_{k+1}/2}+1}
=\frac{c_1 c_2\sqrt{\epsilon}}{e^{\alpha_{k+1}}+e^{\alpha_{k+1}/2}}
\end{align*}
for any $k \geq 1$. Therefore, for any $1 \leq k\leq k^\star$,
\begin{align*}
e^{-t_1/2}-e^{-t_{k^\star+1}/2}
& = \sum_{i = 1}^{k^\star} \left(e^{-t_i/2}-e^{-t_{i+1}/2}\right)
= \sum_{i = 1}^{k^\star} \frac{c_1 c_2\sqrt{\epsilon}}{e^{\alpha_{i+1}}+e^{\alpha_{i+1}/2}}
\geq \frac{k^\star  c_1c_2\sqrt{\epsilon} }{1+c_1c_2+\sqrt{1+c_1c_2}},  \\
e^{-t_k/2}-e^{-t_{k^\star}/2}
& = \sum_{i = k}^{k^\star-1} \left(e^{-t_i/2}-e^{-t_{i+1}/2}\right)
= \sum_{i = k}^{k^\star-1} \frac{c_1 c_2\sqrt{\epsilon}}{e^{\alpha_{i+1}}+e^{\alpha_{i+1}/2}}
\geq \frac{ (k^\star-k)c_1c_2\sqrt{\epsilon} }{1+c_1c_2+\sqrt{1+c_1c_2}},
\end{align*}
which implies the first inequality in \eqref{eq:kstar_bd_newton}
and
\begin{equation}
e^{-t_k/2} \geq e^{-t_{k^\star}/2}+
\frac{ (k^\star-k)c_1c_2\sqrt{\epsilon} }{1+c_1c_2+\sqrt{1+c_1c_2}}
\geq \sqrt{\epsilon}+\frac{c_1c_2\sqrt{\epsilon} }{1+c_1c_2+\sqrt{1+c_1c_2}} (k^\star-k)
\end{equation}
by using the fact that $t_{k^\star} \leq \ln(\epsilon^{-1})$.
Therefore, for any $1 \leq k \leq k^\star$,
\begin{equation}
\label{eq:bound_exp_t}
e^{t_k}\leq \frac{\epsilon^{-1}}{(1+C(k^\star-k))^2} \, ,
\end{equation}
where $C=\frac{c_1c_2}{1+c_1c_2+\sqrt{1+c_1c_2}}<1$.
Now we are ready to prove the second inequality in \eqref{eq:kstar_bd_newton}. By
\eqref{eq:bound_exp_t} and the fact that $t_{k^\star} \leq \ln(\epsilon^{-1})$,
it follows that
\begin{eqnarray*}
\sum_{k=1}^{k^\star} e^{t_k}t_k
&\leq& e^{t_{k^\star}}t_{k^\star}  +
\sum_{k=1}^{k^\star-1} \epsilon^{-1} \frac{1}{(1+C(k^\star-k))^2} \ln
\left( \epsilon^{-1} \frac{1}{(1+C(k^\star-k))^2}\right)\\
&\leq&
\epsilon^{-1} \ln(\epsilon^{-1})
+ \int_1^{k^\star} \epsilon^{-1} \frac{1}{(1+C(k^\star-x))^2}
\ln \left( \epsilon^{-1} \frac{1}{(1+C(k^\star-x))^2}\right) dx\\
&=& \epsilon^{-1} \ln(\epsilon^{-1})
+ \frac{1}{2C\sqrt{\epsilon}}\int_a^b t e^{t/2} dt = 
\epsilon^{-1} \ln(\epsilon^{-1}) + \frac{1}{C\sqrt{\epsilon}}
( (b-2)e^{b/2}-(a-2)e^{a/2} ) 
\end{eqnarray*}
where $a=\ln (\epsilon^{-1})-2\ln(1+C(k^\star-1))$ and
$b=\ln (\epsilon^{-1})$. Using \eqref{eq:bound_exp_t} with $k = 1$ and
 the fact that function $(t-2)e^{t/2}$
is increasing over $t \geq 0$, we have that
\begin{align*}
& a  =\ln (\epsilon^{-1})-2\ln(1+C(k^\star-1))\geq t_1 >0
\text{ and } (a-2)e^{a/2}  \geq -2 \, .
\end{align*}
Hence,
\begin{align*}
&\frac{ (b -2)e^{b/2}-(a-2)e^{a/2} }{C\sqrt{\epsilon}}
\leq \frac{(b -2)e^{b/2}}{C\sqrt{\epsilon}} +
\frac{2}
{C\sqrt{\epsilon}} =
\frac{\epsilon^{-1}(\ln (\epsilon^{-1})-2)}{C}
+\frac{2}{C\sqrt{\epsilon}}
\leq \frac{\epsilon^{-1}\ln (\epsilon^{-1})}{C} \, ,
\end{align*}
provided that $\epsilon<1$.  Combining, we obtain that
\begin{equation*}
  \sum_{k=1}^{k^\star} e^{t_k}t_k \leq
  \epsilon^{-1} \ln(\epsilon^{-1}) +
  \frac{\epsilon^{-1} \ln(\epsilon^{-1})  }{C}
 = \frac{(C+1)\epsilon^{-1}\ln (\epsilon^{-1})}{C} \, .
\end{equation*}
Similarly, by using \eqref{eq:bound_exp_t}, we have that
\begin{eqnarray*}
\sum_{k=1}^{k^\star} e^{t_k} &\leq&
\epsilon^{-1} +
\sum_{k=1}^{k^\star-1} \epsilon^{-1}
\frac{1}{(1+C(k^\star-k))^2}
\leq \epsilon^{-1} + \int_{1}^{k^\star}
\frac{\epsilon^{-1}}{(1+C(k^\star-x))^2} \, dx\\
&=& \epsilon^{-1} +
\frac{\epsilon^{-1}}{C} \left(1 - \frac{1}{1+C(k^\star - 1)} \right)
\leq \frac{(C+1)\epsilon^{-1}}{C} \, .
\end{eqnarray*}
Consequently, we have that
\begin{equation*}
\sum_{k=1}^{k^\star} e^{t_k} (t_k+1)
\leq \frac{(C+1)\epsilon^{-1}(1 + \ln (\epsilon^{-1}))}{C}
= \frac{(1+2c_1c_2+\sqrt{1+c_1c_2}) \epsilon^{-1}(1 + \ln (\epsilon^{-1}))}{c_1c_2} \, .
\end{equation*}
This completes the proof of Lemma \ref{lm:compleixty}.

\vskip 0.1in
\noindent
\textbf{Proof of Theorem \ref{thm:complexity_newton}. }
We first consider the case where $t_{\max} = \infty$ and $\gamma_1 \geq 1$.
In view of the termination criterion \eqref{eq:termination_criterion},
the algorithm will be terminated when $t_k = \mathcal O(\ln(\epsilon^{-1}))$.
We define $N = \max\{k: t_k \leq \ln(\epsilon^{-1}) \}$.
By applying Lemma \ref{lm:thm3} and \ref{lm:thetak_theta_true_bd} with $c_0 = 26/25$,
we have
\begin{equation}
  \frac{\|\theta_k\|_2}{1 - e^{-t_k}}
  \leq \frac{5}{4}\frac{\|\theta(t_k)\|_2}{1 - e^{-t_k}}
  \leq \frac{5}{4}\left(\|\theta(t_{\max})\|_2 + \|\nabla L_n(\bm 0)\|_2\right) \, .
\end{equation}
Therefore,
\begin{align*}
  e^{A_k} - 1 & = \frac{e^{t_k/2}(e^{\alpha_1} - 1)\|\nabla L_n(\bm 0)\|_2(1-e^{-t_k})}
  {\|\theta_k\|_2} \geq \frac{4e^{t_k/2}(e^{\alpha_1} - 1)\|\nabla L_n(\bm 0)\|_2}
  {5(\|\theta(t_{\max})\|_2 + \|\nabla L_n(\bm 0)\|_2)}
  \, , \\
  e^{B_k} - 1 & =  \left(C_2\beta e^{t_{k}}
  \max\left((e^{t_{k}} - 1)^{-\gamma_1},
  (e^{t_{k}} - 1)^{-1-\gamma_2/2}\right)
  \|\theta_k\|_2\right)^{-1} \\
  & \geq \left(C_2\beta
  \max\left((e^{t_{k}} - 1)^{1-\gamma_1},
  (e^{t_{k}} - 1)^{-\gamma_2/2}\right)
  (\|\theta(t_{\max})\|_2 + \|\nabla L_n(\bm 0)\|_2)\right)^{-1} \\
  & \geq \left(C_2\beta
  (\|\theta(t_{\max})\|_2 + \|\nabla L_n(\bm 0)\|_2)\right)^{-1} \, ,
\end{align*}
which implies that,
\begin{equation*}
  \alpha_{k+1} \geq \ln(1 + \nu_1 e^{t_k/2}(e^{\alpha_1} - 1))
\end{equation*}
when $\alpha_{k+1} \leq \min(\alpha_{\max}, \, \ln(1 + \nu_2))$,
where we treat
\begin{equation*}
  \nu_1 = \frac{4\|\nabla L_n(\bm 0)\|_2}
  {5(\|\theta(t_{\max})\|_2 + \|\nabla L_n(\bm 0)\|_2)}
  \text{ and }
  \nu_2 = \left(C_2\beta
  (\|\theta(t_{\max})\|_2 + \|\nabla L_n(\bm 0)\|_2)\right)^{-1}
\end{equation*}
as problem-dependent constants.
Then, applying Lemma \ref{lm:compleixty}, we have that
\begin{equation*}
  N \leq \mathcal O\left(\epsilon^{-1/2} +
  \ln(\epsilon^{-1}) / \min(\alpha_{\max}, \ln(1+\nu_2))\right)
  = \mathcal O\left(\epsilon^{-1/2}\right) \, ,
\end{equation*}
if we treat $\beta$, $\|\theta(t_{\max})\|_2$, and $\|\nabla L_n(\bm 0)\|_2$
as constants.
When $t_{\max} < \infty$ and $\gamma_1 \geq 1$, the algorithm terminates at $N$ if
$t_N > t_{\max}$. Since $\alpha_k$ is increasing, it follows that
$N \leq t_{\max} / \alpha_1 = \mathcal O\left(\epsilon^{-1/2}\right)$.
This completes the proof of Theorem \ref{thm:complexity_newton}.

\vspace{.1in}
\noindent
\textbf{Proof of Theorem \ref{thm:gradient_convergence}. }
It is easy to verify that $f_{t_k}(\theta)$ is $m_k$-strongly convex
with $L_k$-Lipschitz gradient, where
 $m_k = m (1-e^{-t_{k}}) + e^{-t_{k}}$ and
$L_k = L (1-e^{-t_{k}}) + e^{-t_{k}}$.
By standard analysis of gradient descent for strongly convex and smooth functions
 \cite[see, e.g., Theorem 2.1.14 of][]{nesterov1998}, we have that
\begin{eqnarray}
  \label{eq:gd_convergence_theta}
\|\theta_{k+1} - \theta(t_{k+1})\|_2 &\leq&
\left(1 - \frac{2m_{k+1}L_{k+1}}{m_{k+1}+L_{k+1}} \eta_{k+1}\right)^{n_{k+1}}
\|\theta_{k} - \theta(t_{k+1})\|_2 \, ,
\end{eqnarray}
where $\eta_{k+1} \leq \frac{2}{m_{k+1}+L_{k+1}}$.
Similar to the derivation of \eqref{eq:lip_bound},
 we obtain that
\begin{equation}
  \label{eq:lip_bound_strongly_convex}
  \|\theta(t') - \theta(t)\|_2
  \leq \frac{|C(t') - C(t)|}{C(t) + m C(t) C(t')} \|\theta(t)\|_2 \, ,
\end{equation}
for any $t' < t$,
because $C(t')L_n(\theta) + \frac{1}{2} \|\theta\|_2^2$ is
$(1 + C(t')m)$-strongly convex.
Using this with $t' = t_{k}, t = t_{k+1}$,
and applying the triangular inequality, we obtain that
\begin{eqnarray*}
 &&  \|\theta_{k} - \theta(t_{k+1})\|_2
  \leq \|\theta_{k} - \theta(t_{k})\|_2
  + \|\theta(t_{k}) - \theta(t_{k+1})\|_2 \nonumber \\
& = &\|\theta_{k} - \theta(t_{k})\|_2 +
 \frac{(e^{t_{k+1}} - e^{t_k})}{(e^{t_{k+1}} - 1)(1 + m (e^{t_{k}} - 1))}
\|\theta(t_{k+1})\|_2\\
& \leq &
\|\theta_{k} - \theta(t_{k})\|_2 +
 \frac{(e^{\alpha_{k+1}} - 1)}{(e^{t_{k+1}} - 1)m_k}\|\theta(t_{k+1})\|_2
\, .
\end{eqnarray*}
Combining this with \eqref{eq:gd_convergence_theta}, we get
\begin{align}
  \|\theta_{k+1} - \theta(t_{k+1})\|_2 & \leq
  \left(1 - \frac{2m_{k+1}L_{k+1}}{m_{k+1}+L_{k+1}} \eta_{k+1}\right)^{n_{k+1}}
  \nonumber \\
  & \quad \quad \quad \left(
  \|\theta_{k} - \theta(t_{k})\|_2 +
 \frac{(e^{\alpha_{k+1}} - 1)\|\theta(t_{k+1})\|_2}{(e^{t_{k+1}} - 1)m_k}
 \right) \, .\label{eq:grad_induction_bd}
\end{align}
Next we use induction to show that
\begin{equation}
\label{eq:thm:grad_tmp0}
   \|\theta_{k} - \theta(t_{k})\|_2 \leq
  2 \left(1 - \frac{2m_{k}L_{k}}{m_{k}+L_{k}} \eta_{k}\right)^{n_{k}}
  \frac{(e^{\alpha_{k}} - 1)\|\theta(t_{k})\|_2}{(e^{t_{k}} - 1)m_{k-1}} \, .
\end{equation}
Suppose that \eqref{eq:thm:grad_tmp0} holds for $\theta_k$, then using
\eqref{eq:grad_induction_bd} and \eqref{eq:thm:grad_tmp0}, it follows that
\eqref{eq:thm:grad_tmp0} holds for $\theta_{k+1}$ if
\begin{equation}
\label{eq:thm_grad_tmp1}
 2 \left(1 - \frac{2m_{k}L_{k}}{m_{k}+L_{k}} \eta_{k}\right)^{n_{k}}
   \frac{(e^{\alpha_{k}} - 1)\|\theta(t_{k})\|_2}{(e^{t_{k}} - 1)m_{k-1}}
 \leq  \frac{(e^{\alpha_{k+1}} - 1)\|\theta(t_{k+1})\|_2}{(e^{t_{k+1}} - 1)m_k}
\end{equation}
for any $k \geq 1$. Next we show that \eqref{eq:thm_grad_tmp1} can be ensured by
the conditions in \eqref{eq:grad_cond_unified}. First,
using the fact that
\begin{equation}
  \frac{m_{k}}{m_{k-1}}
  = \frac{m(1-e^{-t_{k}}) + e^{-t_{k}}}{m(1-e^{-t_{k-1}}) + e^{-t_{k-1}}}
  \leq
  \begin{cases}
    m_1 & \text{ when } k = 1 \\
    \frac{1-e^{-t_{k}}}{1-e^{-t_{k-1}}} & \text{ for any } k \geq 2 \, ,
  \end{cases}
\end{equation}
and
\begin{align*}
  \frac{(e^{t_{k+1}} - 1)(e^{\alpha_{k}} - 1)(1-e^{-t_{k}})}
  {(e^{t_k} - 1)(e^{\alpha_{k+1}} - 1)(1-e^{-t_{k-1}})}
\leq 12 \text{ for any } k \geq 2, \,
 \frac{(e^{t_{2}} - 1)(e^{\alpha_{1}} - 1)}
  {(e^{t_1} - 1)(e^{\alpha_{2}} - 1)}
\leq 5 \, ,
\end{align*}
 when $\alpha_k \leq 2 \alpha_{k-1}$, $\alpha_{k+1} \geq \alpha_k / 2$,
  and $\alpha_k \leq \ln(2)$ for $k \geq 1$,
 it follows that
a sufficient condition for \eqref{eq:thm_grad_tmp1} is
\begin{equation*}
n_1 \geq \frac{\log(10m_1)}
{-\log\left(1 - \frac{2m_1 L_1}{m_1+L_1} \eta_{1}\right)}
\text{ and }
n_{k} \geq \frac{\log(24)}
{-\log\left(1 - \frac{2m_{k}L_{k}}{m_{k}+L_{k}} \eta_{k}\right)}
\end{equation*}
for any $k \geq 2$.

Next, using the fact that $f_{t_k}(\cdot)$ has $L_k$-Lipschitz gradient,
we have
\begin{equation*}
  \|g_k\|_2 \leq L_k \|\theta_k - \theta(t_{k})\|_2
  \leq 2 L_k  \left(1 - \frac{2m_{k}L_{k}}{m_{k}+L_{k}} \eta_{k}\right)^{n_{k}}
  \frac{(e^{\alpha_{k}} - 1)\|\theta(t_{k})\|_2}{(e^{t_{k}} - 1)m_{k-1}} \, .
\end{equation*}
This completes the proof of Theorem \ref{thm:gradient_convergence}.

\vskip .1in
\noindent
\textbf{Proof of Theorem \ref{thm:uniform_grad_bd}. }
Since $C(t)L_n(\theta) + \frac{1}{2} \|\theta\|_2^2$ is
$(1 + C(t)m)$-strongly convex, using \eqref{eq:str_con_prop2}, we have that
\begin{equation}
\label{eq:thm5_tmp3}
  \|\theta(t) - \bm 0\|_2 \leq \frac{\|C(t)\nabla L_n(\bm 0)\|_2}{1+C(t)m} \, ,
\end{equation}
which implies that,
\begin{equation}
\label{eq:theta_bd_strong_convex}
  \|\theta(t_k)\|_2 \leq
   \frac{1-e^{-t_k}}{m_k} \|\nabla L_n(\bm 0)\|_2\, .
\end{equation}
By Theorem \ref{thm:gradient_convergence}, since
\begin{equation}
  n_{k} \geq \frac{\log(24) + \max(0, \log(L_k / m_{k-1}))}
{-\log\left(1 - \frac{2m_{k}L_{k}}{m_{k}+L_{k}} \eta_{k}\right)} \, ,
\end{equation}
it follows that
\begin{equation}
\label{eq:thm5_tmp1}
  \|g_k\|_2 \leq 2 \left(1 - \frac{2m_{k}L_{k}}{m_{k}+L_{k}} \eta_{k}\right)^{n_{k}}
  \frac{(e^{\alpha_{k}} - 1)\|\theta(t_{k})\|_2}{(e^{t_{k}} - 1)m_{k-1}}
  \leq (12)^{-1}
  \frac{(e^{\alpha_{k}} - 1)\|\theta(t_{k})\|_2}{(e^{t_{k}} - 1)}  \, .
\end{equation}
Similar to \eqref{eq:thetak_theta_true_bd}, we can show that
\begin{equation}
\label{eq:grad_thetak_bd}
  \frac{1}{1 + (12)^{-1}}
  \leq \frac{\|\theta_k\|_2}{\|\theta(t_k)\|_2}
  \leq \frac{1}{1 - (12)^{-1}} \, .
\end{equation}
Using this, Lemma \ref{lm:thm3}, and \eqref{eq:thm_fn_bound1},
we have that
\begin{align*}
  \sup_{t \in [0, t_{1}]} &
   \left\{ f_t(\tilde \theta(t)) - f_t(\theta(t)) \right\}  \leq
  \max\left( e^{t_1} \|g_1\|_2^2 , \,
    \|\theta_1\|_2^2 \right) +
   \frac{e^{t_1} (1-e^{-t_1})^2}{2}
  \|\nabla L_n(\bm 0)\|_2^2   \\
   & \leq   \frac{e^{t_1} (1-e^{-t_1})^2}{2}
  \|\nabla L_n(\bm 0)\|_2^2 +  (3/2)  \|\theta(t_1)\|_2^2 \leq
   2 (e^{\alpha_1} - 1)^2 \|\nabla L_n(\bm 0)\|_2^2 \, .
\end{align*}
Moreover, using \eqref{eq:thm5_tmp1}, \eqref{eq:grad_thetak_bd}, and
Corollary \ref{cor:solution_prop}, we obtain that
\begin{align*}
   & e^{t_{k+1}} \left( \frac{1-e^{-t_{k+1}}}{1 - e^{-t_k}} \right)^2  \|g_k\|_2^2 \leq
   \frac{e^{t_{k+1}}}{144}
   \left( \frac{1-e^{-t_{k+1}}}{1 - e^{-t_k}} \right)^2
   \left(\frac{(e^{\alpha_{k}} - 1)\|\theta(t_{k})\|_2}{(e^{t_{k}} - 1)}\right)^2
   \\
   & \leq \frac{e^{\alpha_{k+1}}}{144}
    \frac{(e^{\alpha_{k}} - 1)^2(1-e^{-t_{k+1}})^2}{(e^{\alpha_{k+1}} - 1)^2(1-e^{-t_{k}})^2}
   e^{-t_{k}} \left( \frac{e^{\alpha_{k+1}} - 1}{1 - e^{-t_k}} \right)^2
    \|\theta(t_{k})\|_2^2
    \leq (20)^{-1} e^{-t_{k}} \left( \frac{e^{\alpha_{k+1}} - 1}{1 - e^{-t_k}} \right)^2
    \|\theta_k\|_2^2
   \\
   & e^{t_{k+1}} \|g_{k+1}\|_2^2  \leq  \frac{e^{t_{k+1}}}{144}
   \left(\frac{(e^{\alpha_{k + 1}} - 1)\|\theta(t_{k + 1})\|_2}{(e^{t_{k + 1}} - 1)}\right)^2
   \leq \frac{e^{\alpha_{k+1}}}{144}
    e^{-t_{k}} \left( \frac{e^{\alpha_{k+1}} - 1}{1 - e^{-t_k}} \right)^2
    \|\theta(t_{k})\|_2^2 \\
    &  \quad \quad \quad
    \leq (50)^{-1}
    e^{-t_{k}} \left( \frac{e^{\alpha_{k+1}} - 1}{1 - e^{-t_k}} \right)^2
    \|\theta_k\|_2^2 \, ,
\end{align*}
and
\begin{align*}
  & (e^{-t_k} - e^{-t_{k+1}})^2 \max \left\{
   \frac{e^{t_{k+1}} \|\theta_k\|_2^2 }{(1 - e^{-t_{k}})^2} , \,
   \frac{e^{t_k} \|\theta_{k+1}\|_2^2 }{(1 - e^{-t_{k+1}})^2} \right\} \\
   & \leq
    \left(1-(12)^{-1}\right)^{-2}
    (e^{-t_k} - e^{-t_{k+1}})^2 \max \left\{
   \frac{e^{t_{k+1}} \|\theta(t_k)\|_2^2 }{(1 - e^{-t_{k}})^2} , \,
   \frac{e^{t_k} \|\theta(t_{k+1})\|_2^2 }{(1 - e^{-t_{k+1}})^2} \right\} \\
  & \leq  \left(1-(12)^{-1}\right)^{-2}
  e^{-t_{k}} \left( \frac{e^{\alpha_{k+1}} - 1}{1 - e^{-t_k}} \right)^2
    \|\theta(t_{k})\|_2^2 \\
  & \leq
    \left(1-(12)^{-1}\right)^{-2}  \left(1+(12)^{-1}\right)^{2}
  e^{-t_{k}} \left( \frac{e^{\alpha_{k+1}} - 1}{1 - e^{-t_k}} \right)^2
    \|\theta_k\|_2^2 \, ,
\end{align*}
where we have used the fact that
$e^{\alpha_{k+1}} (e^{\alpha_{k}} - 1) (1-e^{-t_{k+1}})
/ (e^{\alpha_{k+1}} - 1) (1 - e^{-t_{k}}) \leq (\sqrt{2} + 1)^2$.
Combining these with \eqref{eq:thm_fn_bound2},
we have that
\begin{align*}
   & \sup_{t \in [t_k, t_{k+1}]}
    \left\{ f_t(\tilde \theta(t)) - f_t(\theta(t)) \right\}  \leq
   e^{t_{k+1}}
   \max \left\{\left( \frac{1-e^{-t_{k+1}}}{1 - e^{-t_k}} \right)^2  \|g_k\|_2^2, \,
  \|g_{k+1}\|_2^2 \right\} \nonumber \\
   & \quad \quad \quad \quad
   +  (e^{-t_k} - e^{-t_{k+1}})^2 \max \left\{
   \frac{e^{t_{k+1}} \|\theta_k\|_2^2 }{(1 - e^{-t_{k}})^2} , \,
   \frac{e^{t_k} \|\theta_{k+1}\|_2^2 }{(1 - e^{-t_{k+1}})^2} \right\} \\
    & \leq
    \left( (20)^{-1} +  (1-(12)^{-1})^{-2}  (1+(12)^{-1})^{2}   \right)
    e^{-t_{k}} \left( \frac{e^{\alpha_{k+1}} - 1}{1 - e^{-t_k}} \right)^2
    \|\theta_k\|_2^2 \\
    & \leq 2 e^{-t_{k}} \left( \frac{e^{t_{k+1}-t_k} - 1}{1 - e^{-t_k}} \right)^2
    \|\theta_k\|_2^2
\end{align*}
for any $k \geq 1$.

Lastly, using \eqref{eq:thm_fn_bound3}, \eqref{eq:thm5_tmp1},
\eqref{eq:thm4_tmp9}, and
part (ii) of Corollary \ref{cor:solution_prop}, we have
\begin{align*}
  \sup_{t_N < t \leq t_{\max}}
  & \left\{ f_t(\tilde \theta(t)) - f_t(\theta(t)) \right\}
 \leq
  \frac{e^{t_N}(1 - e^{-t_{\max}})}{1 - e^{-t_N}} \|g_N\|_2^2 +
  \frac{3}{2(e^{t_N}-1)} \|\theta(t_{\max})\|_2^2
   \nonumber \\
    & \leq \frac{e^{t_N}(1 - e^{-t_{\max}})}{144(1 - e^{-t_N})} \left( \frac{e^{\alpha_N}-1}
  {e^{t_N} - 1} \right)^2
   \|\theta(t_N)\|_2^2 +
  \frac{3}{2(e^{t_N}-1)} \|\theta(t_{\max})\|_2^2 \nonumber  \\
  & \leq   \frac{2\|\theta(t_{\max})\|_2^2}{e^{t_N} - 1} \, .
\end{align*}
Combining the three bounds,
we obtain that when $t_N \leq t_{\max}$,
\begin{align*}
   \sup_{0 \leq t \leq t_{\max}}  & \left\{ f_t(\tilde \theta(t)) - f_t(\theta(t)) \right\}
    \leq 2
    \max \Bigg( (e^{\alpha_1} - 1)^2 \|\nabla L_n(\bm 0)\|_2^2 \, ,  \\
    &
     \max_{ 1 \leq k \leq N-1}
   e^{-t_{k}} \frac{(e^{\alpha_{k+1}} - 1)^2}{(1 - e^{-t_k})^2}
    \|\theta_k\|_2^2 \, ,
    \frac{\|\theta(t_{\max})\|_2^2}{e^{t_N} - 1}   \Bigg) \, ,
\end{align*}
and when $t_{N-1} \leq t_{\max} < t_N$ for some $N \geq 1$.
\begin{align*}
   & \sup_{0 \leq t \leq t_{\max}}   \left\{ f_t(\tilde \theta(t)) - f_t(\theta(t)) \right\}
     \leq
    \sup_{0 \leq t \leq t_N}  \left\{ f_t(\tilde \theta(t)) - f_t(\theta(t)) \right\}  \\
    & \leq 2 \max \Bigg( (e^{\alpha_1} - 1)^2 \|\nabla L_n(\bm 0)\|_2^2 \, ,
     \max_{ 1 \leq k \leq N-1}
   e^{-t_{k}} \frac{(e^{\alpha_{k+1}} - 1)^2}{(1 - e^{-t_k})^2}
    \|\theta_k\|_2^2  \Bigg) \, .
\end{align*}
This completes the proof of Theorem \ref{thm:uniform_grad_bd}.

\vskip .1in
\noindent
\textbf{Proof of Theorem \ref{thm:gd_bd}. }
We first show that $\alpha_{k+1} \geq \alpha_k / 2$.
If $\alpha_{k+1} = 5^{-1}$ or $2\alpha_k$, then trivially
$\alpha_{k+1} \geq \alpha_k / 2$.
Now we assume that $\alpha_{k+1} = A_k$, where
\begin{align*}
  A_k & := \ln \left( 1 +
  \frac{ \epsilon^{1/2} e^{t_k/2} (1-e^{-t_k})}
  {\|\theta_k\|_2} \right) \, .
\end{align*}
First,
using \eqref{eq:thm5_tmp1}, \eqref{eq:grad_thetak_bd}, and
the fact that $\|\theta_k - \theta(t_k)\|_2
\leq e^{t_k} \|g_k\|_2$, we obtain that
\begin{align*}
  \|\theta_k\|_2 & \leq
  \|\theta(t_k)\|_2 +
  \|\theta(t_k) - \theta_k\|_2
  \leq \|\theta(t_k)\|_2 + e^{t_k} \|g_k\|_2
  \leq \|\theta(t_k)\|_2 + e^{t_k}
  \frac{(e^{\alpha_{k}} - 1)}{12(e^{t_{k}} - 1)}\|\theta(t_k)\|_2 \, ,
\end{align*}
which implies that
\begin{equation}
\label{eq:thetak_bd}
  \|\theta_k\|_2 \leq
  \|\theta(t_k)\|_2 \left(1 + \frac{(e^{\alpha_{k}} - 1)}{12(1 - e^{-t_{k}})} \right) \, .
\end{equation}

When $k = 1$, using \eqref{eq:thetak_bd} and the fact that $e^{\alpha_1} - 1
\leq \sqrt{\epsilon} / \|\nabla L_n(\bm 0)\|_2$, we obtain that
\begin{align*}
 \frac{e^{\alpha_{2}} - 1}{e^{\alpha_{1}/2} - 1}  & =
  (e^{\alpha_{1}/2} + 1) \frac{e^{\alpha_{2}} - 1}{e^{\alpha_{1}} - 1}
  \geq (e^{\alpha_{1}/2} + 1) \frac{e^{A_1} - 1}{e^{\alpha_1} - 1} =
  (e^{\alpha_1 / 2} + 1) e^{\alpha_1 / 2}
  \frac{\sqrt{\epsilon} (1-e^{-t_1})}
  {\|\theta_1\|_2(e^{\alpha_1} - 1)} \\
  & \geq (e^{\alpha_1 / 2} + 1) e^{\alpha_1 / 2}
  \frac{\sqrt{\epsilon} (1-e^{-t_1})}
  {\|\theta(t_1)\|_2 (e^{\alpha_1} - 1)}
  \left(1 + \frac{(e^{\alpha_{1}} - 1)}{12(1 - e^{-t_{1}})} \right)^{-1} \\
  & \geq
  \frac{(e^{-\alpha_1 / 2} + 1) \sqrt{\epsilon} } {\|\nabla L_n(\bm 0)\|_2 (e^{\alpha_1} - 1)}
   \left(1 + e^{\alpha_{1}} / 12 \right)^{-1}
   \geq \frac{3}{4}
   (e^{-\alpha_1 / 2} + 1)(1 + e^{\alpha_{1}} / 12)^{-1} \geq 1 \, ,
\end{align*}
provided that $\alpha_1 \leq \ln(2)$.
This implies that $\alpha_2 \geq \alpha_1 / 2$.

When $k \geq 2$, note that $\alpha_k \leq A_{k-1}$ and
\begin{align}
 \frac{e^{\alpha_{k+1}} - 1}{e^{\alpha_{k}/2} - 1}  & =
  (e^{\alpha_{k}/2} + 1) \frac{e^{\alpha_{k+1}} - 1}{e^{\alpha_{k}} - 1}
  \geq (e^{\alpha_{k}/2} + 1) \frac{e^{A_k} - 1}{e^{A_{k-1}} - 1}  \nonumber \\
  & = (e^{\alpha_k / 2} + 1) e^{\alpha_k / 2}
  \frac{\epsilon^{1/2}(1-e^{-t_k})}
  {\|\theta_k\|_2}
  \frac{\|\theta_{k-1}\|_2}
  {\epsilon^{1/2}(1-e^{-t_{k-1}})} \nonumber \\
  & \geq (e^{-\alpha_k / 2} + 1) \frac{(e^{t_k} - 1)\|\theta_{k-1}\|_2}
  {(e^{t_{k-1}} - 1)\|\theta_{k}\|_2} \, .
  \label{eq:cor_general_tmp1}
\end{align}
Now similar to \eqref{eq:thetak_bd}, we obtain that
\begin{equation*}
  \|\theta(t_{k})\|_2 \leq \|\theta_k\|_2 +
  \|\theta(t_{k}) - \theta_k\|_2 \leq
  \|\theta_k\|_2 +
  e^{t_k} \|g_k\|_2 \leq
  \|\theta_k\|_2 +
   \frac{(e^{\alpha_{k}} - 1)}{12(1 - e^{-t_{k}})}\|\theta(t_k)\|_2 \, ,
\end{equation*}
we have that
\begin{equation}
\label{eq:thetak_thetat_bd_GD}
  \left( 1 -  \frac{(e^{\alpha_{k}} - 1)}{12(1 - e^{-t_{k}})} \right)
  \leq \frac{\|\theta_k\|_2}{\|\theta(t_{k})\|_2}
  \leq \left( 1 +  \frac{(e^{\alpha_{k}} - 1)}{12(1 - e^{-t_{k}})} \right) \, .
\end{equation}
Hence,
\begin{equation*}
   \frac{(e^{t_k} - 1)\|\theta_{k-1}\|_2}
  {(e^{t_{k-1}} - 1)\|\theta_{k}\|_2} \geq
   \frac{(e^{t_k} - 1)\|\theta(t_{k-1})\|_2}
  {(e^{t_{k-1}} - 1)\|\theta(t_{k})\|_2}
  \frac{1 -  \frac{(e^{\alpha_{k}} - 1)}{12(1 - e^{-t_{k}})}}
  {1 +  \frac{(e^{\alpha_{k}} - 1)}{12(1 - e^{-t_{k}})}}
  \geq \frac{1 -  \frac{(e^{\alpha_{k}} - 1)}{12(1 - e^{-t_{k}})}}
  {1 +  \frac{(e^{\alpha_{k}} - 1)}{12(1 - e^{-t_{k}})}}
\end{equation*}
where the last inequality uses part (ii) of Corollary \ref{cor:solution_prop}.
Combining this with \eqref{eq:cor_general_tmp1}, we obtain that
\begin{equation}
\label{eq:cor_general_tmp2}
  \frac{e^{\alpha_{k+1}} - 1}{e^{\alpha_{k}/2} - 1}
  \geq (e^{-\alpha_k / 2} + 1)
  \frac{1 -  \frac{(e^{\alpha_{k}} - 1)}{12(1 - e^{-t_{k}})}}
  {1 +  \frac{(e^{\alpha_{k}} - 1)}{12(1 - e^{-t_{k}})}} \, .
\end{equation}
Therefore, to prove $\alpha_{k+1} \geq \alpha_{k} / 2$, it suffices to show that
the RHS of the above inequality is no smaller than $1$. To this end,
using the fact that $\alpha_{k} \leq \ln(2)$ for all $k \geq 1$, we have
\begin{align*}
  \frac{(e^{\alpha_{k}} - 1)}{(1 - e^{-t_{k}})}
  & = \frac{e^{\alpha_{k}}(1 - e^{-\alpha_{k}})}{1 - e^{-t_{k}}}
  \leq e^{\alpha_k} \leq 2
\end{align*}
for any $k \geq 1$.
Combining this with \eqref{eq:cor_general_tmp2}, we have that
\begin{equation*}
   \frac{e^{\alpha_{k+1}} - 1}{e^{\alpha_{k}/2} - 1} \geq (e^{-\alpha_k / 2} + 1)
  \frac{1 -  \frac{(e^{\alpha_{k}} - 1)}{12(1 - e^{-t_{k}})}}
  {1 +  \frac{(e^{\alpha_{k}} - 1)}{12(1 - e^{-t_{k}})}} \geq
  (1/\sqrt{2} + 1)
  \frac{1 -  6^{-1}}
  {1 +  6^{-1}} > 1 \, ,
\end{equation*}
which proves that $\alpha_{k+1} \geq \alpha_{k} / 2$.
Hence, $\alpha_k$ satisfy all the conditions in Theorem \ref{thm:uniform_grad_bd}.

Next, we show that the algorithm will terminate in finite steps.
We first show that $t_k$ diverges.
To this end, using \eqref{eq:theta_norm_bds} and \eqref{eq:grad_thetak_bd},
we have
\begin{align*}
  & \frac{\epsilon^{1/2} e^{t_k/2}(1-e^{-t_k})}
  {\|\theta_k\|_2}  \geq
  \frac{e^{t_k/2}(e^{\alpha_1} - 1)\|\nabla L_n(\bm 0)\|_2(1-e^{-t_k})}
  {\|\theta(t_k)\|_2 / (1-(12)^{-1})}
  \\
  & \geq \frac{e^{t_k/2}(e^{\alpha_1} - 1)\|\nabla L_n(\bm 0)\|_2(1-e^{-t_k})}
  {2(e^{t_k} - 1) \|\nabla L_n(\bm 0)\|_2}
  \geq 2^{-1}(e^{\alpha_1} - 1)e^{-t_k/2}  \, ,
\end{align*}
which implies that
\begin{equation}
  A_k \geq \ln(1+2^{-1}(e^{\alpha_1} - 1)e^{-t_k/2}) \geq
  \frac{(e^{\alpha_1} - 1)e^{-t_k/2}}{2+(e^{\alpha_1} - 1)e^{-t_k/2}}
  =  \frac{(e^{\alpha_1} - 1)}{2e^{t_k/2}+(e^{\alpha_1} - 1)} \, .
\end{equation}
Thus,
\begin{equation}
  \alpha_{k+1} \geq \min\left(\alpha_{\max}, 2\alpha_k,
  \frac{(e^{\alpha_1} - 1)}{2 e^{t_k/2}+(e^{\alpha_1} - 1)}\right) \, .
\end{equation}
Now we prove the divergence of $t_k$ by contradiction.
Suppose that $t_k$ does not diverge. Then there must exist a constant $T$ such that
$t_k < T$ for all $k$. However, now we have
\begin{equation*}
   \alpha_{k+1}  \geq \min\left(\alpha_{\max}, 2\alpha_k,
  \frac{(e^{\alpha_1} - 1)}{2 e^{T/2}+(e^{\alpha_1} - 1)} \right)  \, ,
 \end{equation*}
 which implies that $\alpha_{k}$ is lower bounded by a positive constant
 when $k$ is large enough, implying that
$t_k$ should diverge. This is a contradiction.
Hence, $t_k$ diverges.

Now we are ready to show that the algorithm
must terminate after a finite number of iterations.
If $t_{\max} < \infty$, then $t_k \geq t_{\max}$ must hold
for large enough $k$ as $t_k$ diverges.
If $t_{\max} = \infty$, then we have
that $\theta(t_{\max}) = \theta^\star$ is finite by assumption.
Therefore,
the termination criterion in
\eqref{eq:termination_criterion} should also be met when $N$ is large
enough, because $t_N$ diverges and
\begin{align*}
  &  \frac{\max(\|\theta(t_{\max})\|_2^2, \|\theta_N\|_2^2)}{(e^{t_N}-1)}
  \leq \frac{2\max(\|\theta(t_{\max})\|_2^2, \|\theta(t_N)\|_2^2)}{(e^{t_N}-1)}
   \le \frac{2 \|\theta^\star\|_2^2}{(e^{t_N}-1)}
      \rightarrow 0 \text{ as } N \rightarrow \infty \, .
\end{align*}

Finally, we are ready to prove \eqref{eq:GD_thm_bd_global} upon termination.
Using
\eqref{eq:grad_uniform_general_bd_strong2} and \eqref{eq:grad_uniform_general_bd_strong1}
in Theorem \ref{thm:uniform_grad_bd}, and the definition of $\alpha_1$ and
$\alpha_{k+1}$, we have that
\begin{align*}
  &  e^{-t_{k}} \left( \frac{e^{\alpha_{k+1}} - 1}{1 - e^{-t_{k}}} \right)^2 \|\theta_k\|_2^2
   \leq (e^{\alpha_{1}} - 1)^2 \|\nabla L_n(\bm 0)\|_2^2 \leq \epsilon
   \text{ and } \\
   & \frac{\|\theta(t_{\max})\|_2^2}
   {(e^{t_N} - 1)}  \leq \frac{2\|\theta_N\|_2^2}
   {(e^{t_N} - 1)} \frac{\|\theta(t_{\max})\|_2^2}{2\|\theta_N\|_2^2}
   \leq \epsilon \frac{\|\theta(t_{\max})\|_2^2}{\|\theta(t_N)\|_2^2} \, .
\end{align*}
for any $k \geq 1$ when the algorithm is terminated, and after termination,
\begin{equation*}
   \sup_{0 \leq t \leq t_{\max}}
  \left\{ f_t(\tilde \theta(t)) - f_t(\theta(t)) \right\}
  \leq
  \begin{cases}
    \frac{2\|\theta(t_{\max})\|_2^2}{\|\theta(t_N)\|_2^2} \epsilon &
    \text{ when } t_{\max} \geq t_N \\
    2 \epsilon & \text{ when } t_{\max} < t_N
  \end{cases}\, ,
\end{equation*}
which implies \eqref{eq:GD_thm_bd_global}.
This completes the proof of Theorem \ref{thm:gd_bd}.

\vskip .1in
\noindent
\textbf{Proof of Theorem \ref{thm:complexity_gd}. }
We first bound the number of gradient steps at each iteration.
Since $m = 0$, we have that $m_k = e^{-t_k}$.
  By assumption, we have that $\eta_k = \eta = \mathcal O(\min(1,L^{-1}))$.
  Then the upper bounds on $n_1$ and $n_{k+1}$ in \eqref{eq:grad_stepsize_relaxed}
  can be bounded further as follows
    \begin{align*}
  & \frac{\log(24) + \max(0, \log(L_{k+1}/m_{k}))}
  {-\log\left(1 - \frac{2m_{k+1}L_{k+1}}{m_{k+1}+L_{k+1}} \eta_{k+1} \right)}
   \leq \frac{\log(24) + \max(0, \log(L_{k+1}/m_{k}))}
  {\frac{2m_{k+1}L_{k+1}}{m_{k+1}+L_{k+1}} \eta_{k+1}} \\
  & \leq \frac{\log(24) + \max(0, \log(L_{k+1}/m_{k}))}
  {m_{k+1} \eta}
  \leq \frac{\log(24) + t_k + \log(L_{k+1})}{m_{k+1} \eta}
  = \mathcal O\left(e^{t_{k+1}} (t_{k+1} + 1)\right) \, ,
  \end{align*}
  where we have used $L_k \geq m_k$ for any $k \geq 1$.
  Similarly, we also we have
  \begin{align*}
    & \frac{\log(10m_1 L_1)}
  {-\log\left(1 - \frac{2m_{1}L_{1}}{m_{1}+L_{1}} \eta_{1} \right)}
  \lesssim  e^{t_1}  \, ,
  \end{align*}
  where we have used the fact that $-\log(1-x) \geq x$,
  $L_k \leq \max(L, 1)$, and treated $L$ as a constant.

Next,
By applying Lemma \ref{lm:thm3} and  \eqref{eq:thetak_thetat_bd_GD},
we have
\begin{equation}
  \frac{\|\theta_k\|_2}{1 - e^{-t_k}}
  \leq \frac{7}{6}\frac{\|\theta(t_k)\|_2}{1 - e^{-t_k}}
  \leq \frac{7}{6}\left(\|\theta(t_{\max})\|_2 + \|\nabla L_n(\bm 0)\|_2\right) \, .
\end{equation}
Therefore, when $\alpha_{k+1} \leq \ln(2)$, we have that
\begin{align*}
  e^{\alpha_{k+1}} - 1 & = \frac{e^{t_k/2}(e^{\alpha_1} - 1)\|\nabla L_n(\bm 0)\|_2(1-e^{-t_k})}
  {\|\theta_k\|_2} \geq \frac{6e^{t_k/2}(e^{\alpha_1} - 1)\|\nabla L_n(\bm 0)\|_2}
  {7(\|\theta(t_{\max})\|_2 + \|\nabla L_n(\bm 0)\|_2)}
  \, ,
\end{align*}
which implies that
\begin{equation*}
  \alpha_{k+1} \geq \ln(1 + \nu_1 e^{t_k/2}(e^{\alpha_1} - 1))
\end{equation*}
when $\alpha_{k+1} \leq \ln(2)$,
where we treat
\begin{equation*}
  \nu_1 = \frac{6\|\nabla L_n(\bm 0)\|_2}
  {7(\|\theta(t_{\max})\|_2 + \|\nabla L_n(\bm 0)\|_2)}
\end{equation*}
as problem-dependent constants.

Now we are ready to derive the bound for the number of gradient steps.
When $t_{\max} = \infty$,  in view of the termination criterion
\eqref{eq:termination_criterion_GD},
the algorithm will be terminated when
$t_k = \mathcal O \left( \ln(\epsilon^{-1}) \right)$.
Therefore, the total number of gradient steps can be bounded as
\begin{equation*}
  \sum_{k = 1}^{k^\star} n_k \lesssim
  \sum_{k=1}^{k^\star} e^{t_{k}} (t_{k} + 1)
  \lesssim  \epsilon^{-1} (1 + \ln(\epsilon^{-1}))\, ,
\end{equation*}
where $k^\star = \max\{k: t_k \leq \ln(\epsilon^{-1}) \}$ and we have used
 Lemma \ref{lm:compleixty}. Hence, the total number of gradient steps is
at most $\mathcal O\left(\epsilon^{-1} \ln(\epsilon^{-1}) \right)$.

 When $t_{\max} < \infty$, then in view of the termination criterion
\eqref{eq:termination_criterion_GD}, the algorithm will be terminated when
$t_k > t_{\max}$. Therefore, the total number of gradient steps can be bounded as
\begin{align*}
  & \sum_{k = 1}^{k^\star} n_k + \max(0, t_{\max} - t_{k^\star}) / \ln(1 + \nu_1)
   \lesssim
  \sum_{k=1}^{k^\star} e^{t_{k}} (t_{k} + 1)
  + 1 / \ln(1 + \nu_1) \lesssim  \epsilon^{-1} (1 + \ln(\epsilon^{-1})) \, ,
\end{align*}
where again we have used
Lemma \ref{lm:compleixty}. In this case, the total number of gradient steps is also
at most $\mathcal O\left(\epsilon^{-1} \ln(\epsilon^{-1}) \right)$. This completes
the proof of Theorem \ref{thm:complexity_gd}.

\vskip .1in
\noindent
\textbf{Proof of Corollary \ref{cor:GD_bound}. }
We first show that the bounds in Theorem \ref{thm:uniform_grad_bd} continue to hold.
  The proof is similar to that of Theorem \ref{thm:uniform_grad_bd} with some slight modifications. In particular, upon termination of the gradient descent method
  at $t_k$, we have \eqref{eq:gd_termination_criterion_at_tk}. Replacing the bound
  \eqref{eq:thm5_tmp1} in the proof of Theorem \ref{thm:uniform_grad_bd} by
  \eqref{eq:gd_termination_criterion_at_tk}, and following a similar argument,
  we obtain that
  \begin{multline*}
    \sup_{0 \leq t \leq t_{\max}}   \left\{ f_t(\tilde \theta(t)) - f_t(\theta(t)) \right\}
    \\
    \lesssim   \max \Bigg( (e^{\alpha_1} - 1)^2 \|\nabla L_n(\bm 0)\|_2^2 \, ,
     \max_{ 1 \leq k \leq N-1}
e^{-t_{k}} \left( \frac{e^{\alpha_{k+1}} - 1}{1 - e^{-t_{k}}} \right)^2 \|\theta_k\|_2^2
 \Bigg) \, .
\end{multline*}
when $t_N \geq t_{\max}$ for some $N$; and
  \begin{multline*}
    \sup_{0 \leq t \leq t_{\max}}   \left\{ f_t(\tilde \theta(t)) - f_t(\theta(t)) \right\}
    \lesssim  \max \Bigg( (e^{\alpha_1} - 1)^2 \|\nabla L_n(\bm 0)\|_2^2 \, ,  \\
     \max_{ 1 \leq k \leq N-1}
   e^{-t_{k}} \left( \frac{e^{\alpha_{k+1}} - 1}{1 - e^{-t_{k}}} \right)^2 \|\theta_k\|_2^2 \, ,
     \frac{\|\theta(t_{\max})\|_2^2}
   {(e^{t_N} - 1)}  \Bigg)
\end{multline*}
when $t_N \leq t_{\max}$ and $\|\theta(t_{\max})\|_2 < \infty$.
Then the bound \eqref{eq:GD_cor_bd_global} follows from these bounds, following the
proof of Theorem \ref{thm:gd_bd}. This completes the proof of Corollary \ref{cor:GD_bound}.

\vskip .1in
\noindent
\textbf{Proof of Proposition \ref{pro:function_examples}. }
  We verify condition \eqref{eq:hessian_cond} for the functions in
Table \ref{tab:function_examples} separately.

\vskip .1in
\noindent
\textbf{Log-barrier function. }  Note that
 $L^\prime_n(\theta)=-\theta^{-1}$
and $L_n''(\theta)=\theta^{-2}$.
Consider $\gamma_1= 3/2$, $\gamma_2=1$ and $\beta\geq 2$. Then
 \eqref{eq:Lipschitz_local_neighbor} reduces to
 $|\delta|\leq \theta \beta^{-1}$.
It is easy to see the right hand side
of \eqref{eq:hessian_cond} reduces to
$\beta\delta^2\theta^{-3}$. Under the condition
$|\delta|\leq \theta \beta^{-1}$,
the left hand side of \eqref{eq:hessian_cond} can be bounded as
follows
\begin{equation*}
 \left|-\frac{1}{\theta+\delta}+\theta^{-1}-
\frac{\delta}{\theta^2}\right|
= \frac{\delta^2}{\theta^2}\left|\frac{1}{\theta+\delta}\right|  =
\frac{\delta^2}{\theta^3}\left|\frac{1}{1+\delta\theta^{-1}}\right|
\leq \frac{\delta^2}{\theta^3}\frac{\beta}{\beta-1} \leq
 \beta \frac{\delta^2}{\theta^3} \, .
\end{equation*}
Thus, $-\ln(\theta)$ satisfies the condition
with $\gamma_1=\frac{3}{2}$ and $\gamma_2=1$.

\vskip 0.1in
\noindent
\textbf{Entropy-barrier function. }  Note that the first and second derivatives
are $L_n'(\theta)=\ln(\theta)+1-\theta^{-1}$ and
  $L_n''(\theta)=\theta^{-1}+\theta^{-2}$, respectively.
Consider $\gamma_1=\frac{3}{2}$, $\gamma_2=1$ and any $\beta$
   satisfying  $\frac{\beta}{(\beta-1)^2}\leq 1$. Then
 \eqref{eq:Lipschitz_local_neighbor}
 reduces to $|\delta|\leq \theta \beta^{-1}$.
The right hand side of \eqref{eq:hessian_cond} can be bounded as
\begin{equation*}
\beta \delta^2 (\theta^{-1}+\theta^{-2})^{3/2}=
\beta \frac{\delta^2}{\theta^2} \frac{(1+\theta)^{3/2}}{\theta}
\geq \beta \frac{\delta^2}{\theta^2} (\theta^{-1}+3/2) \, .
\end{equation*}
By Taylor's expansion,
there exists $s\in (0,1)$ such that
\begin{eqnarray*}
&& \left|\ln(\theta+\delta)-\frac{1}{\theta+\delta}-
\ln(\theta)+\theta^{-1}-(\theta^{-1}+\theta^{-2})\delta\right| \\
&=&
\left|\ln(1+\frac{\delta}{\theta})+\frac{\delta}{\theta(\theta+\delta)}
-\frac{\delta}{\theta}(1+\theta^{-1})\right| =
 \left|-\frac{\delta^2}{2\theta^2(1+s \delta\theta^{-1})^2}-\frac{\delta}{\theta^2}
 +\frac{\delta}{\theta(\theta+\delta)}\right|\\
&=&  \frac{\delta^2}{\theta^2}\left(\frac{1}{\theta(1+\delta\theta^{-1})}+
\frac{1}{2(1+s\delta\theta^{-1})^2}\right) \leq
  \frac{\delta^2}{\theta^2}\left(\theta^{-1} \frac{1}{1-\beta^{-1}}+
 \frac{1}{2(1-\beta^{-1})^2}\right)\\
&=&  \frac{\delta^2}{\theta^2} (\frac{\beta}{\beta-1})^2
 \left(\theta^{-1}(1-\frac{1}{\beta})+\frac{1}{2}\right) \leq
 \beta \frac{\delta^2}{\theta^2} (\theta^{-1}+3/2)  \frac{\beta}{(\beta-1)^2}\\
&\leq& \beta \frac{\delta^2}{\theta^2} (\theta^{-1}+3/2)
 \leq \beta \delta^2 (\theta^{-1}+\theta^{-2})^{3/2} \, ,
\end{eqnarray*}
which implies that $\theta \ln(\theta)-\ln(\theta)$ satisfies
\eqref{eq:hessian_cond} with $\gamma_1=3/2$ and $\gamma_2=1$.

\vskip 0.1in
\noindent
\textbf{Logistic function. }
Note that
 $L_n'(\theta)=-\frac{1}{1+e^{\theta}}$ and $L_n''(\theta)
 =\frac{e^{\theta}}{(1+e^{\theta})^2}$. Consider $\gamma_1=1$, $\gamma_2=0$ and any
 $\beta$ satisfying $e^{1/\beta} \leq 2\beta$ (e.g.,  $\beta=2$).
 Then \eqref{eq:Lipschitz_local_neighbor} reduces to $|\delta|\leq \beta^{-1}$.
The right hand side of \eqref{eq:hessian_cond} is
$\beta \delta^2 e^{\theta} (1+e^{\theta})^{-2}$. By Taylor's expansion,
 there exists $s\in (0,1)$ such that
\begin{equation*}
\left|-\frac{1}{1+e^{\theta+\delta}}+\frac{1}{1+e^{\theta}}-\frac{e^{\theta}}{(1+e^{\theta})^2}\delta\right|
=  \frac{\delta^2}{2}
\left|\frac{e^{\theta+s\delta}-1}{e^{\theta+s\delta}+1}\right|
\frac{e^{\theta+s\delta}}{(e^{\theta+s\delta}+1)^2}
\leq \frac{\delta^2}{2}\frac{e^{\theta+s\delta}}
{(e^{\theta+s\delta}+1)^2} \, .
\end{equation*}
Let $q=e^{s\delta}$. Since $|\delta| \leq \beta^{-1}$, $s\in (0,1)$,
we have $q\in (e^{-\frac{1}{\beta}}, e^{\frac{1}{\beta}})$. Therefore,
\begin{eqnarray*}
\frac{q e^{\theta}}{(q e^{\theta}+1)^2} &=&
q \left(\frac{q e^{\theta}+1}{e^{\theta}+1}\right)^{-2}
\frac{ e^{\theta}}{(e^{\theta}+1)^2}
=  q\left(q+\frac{1-q}{e^{\theta}+1}\right)^{-2}
\frac{ e^{\theta}}{(e^{\theta}+1)^2}\\
&\leq& \max(q,q^{-1}) \frac{ e^{\theta}}{(e^{\theta}+1)^2}
\leq  e^{\frac{1}{\beta}} \frac{ e^{\theta}}{(e^{\theta}+1)^2} \, .
\end{eqnarray*}
Consequently, for any $|\delta|\leq 1/\beta$ with $\beta$ satisfying $e^{ 1/\beta}
\leq 2\beta$, we have
\begin{equation*}
\frac{\delta^2}{2}\frac{e^{\theta+s\delta}}{(e^{\theta+s\delta}+1)^2}
\leq \frac{\delta^2}{2}e^{\frac{1}{\beta}}\frac{ e^{\theta}}{(e^{\theta}+1)^2}
\leq \beta \delta^2 \frac{e^{\theta}}{(1+e^{\theta})^2} \, ,
\end{equation*}
which implies that $\ln(1+e^{-\theta})$ satisfies
\eqref{eq:hessian_cond} with $\gamma_1=1$ and $\gamma_2=0$.

\vskip 0.1in
\noindent
\textbf{Exponential function. } Note that
 $L_n'(\theta)=-e^{\theta}$ and $L_n''(\theta)=e^{-\theta}$.
 Consider $\gamma_1=1$, $\gamma_2=0$, and $\beta\geq \frac{1}{2}$.
 Then \eqref{eq:Lipschitz_local_neighbor} reduces to
  $|\delta|\leq \beta^{-1}$.
It is easy to see the right hand side of inequality
\eqref{eq:hessian_cond} is $\beta e^{-\theta} \delta^2$.
Note that $e^x\geq 1+x$ and $e^x-1-x-\beta x^2\leq 0$ for any
$\beta \geq 1/2$.
 Thus under the condition that $|\delta|\leq \beta^{-1}$,
 the left hand side of inequality \eqref{eq:hessian_cond}
  satisfies
\begin{equation*}
\left|-e^{-(\theta+\delta)}+e^{-\theta}-e^{-\theta}\delta\right|
= e^{\theta} (e^{-\delta}-1+\delta) \leq e^{\theta} \beta \delta^2 \, ,
\end{equation*}
which implies that $e^{\theta}$
satisfies the condition with $\gamma_1=1$ and $\gamma_2=0$.

\vskip 0.1in
\noindent
\textbf{Square function. } Note that
 $L_n'(\theta)=2\theta$ and $L_n''(\theta)=2$.
Since for any $\delta\in \mathbb R$,
the left hand side of inequality \eqref{eq:hessian_cond}
satisfies
\begin{equation*}
\left|L_n^\prime(\theta+\delta)-L_n^\prime(\theta)-L_n^{\prime\prime}(\theta)\delta \right|=0
\end{equation*}
we conclude that $\theta^2$ satisfies inequality \eqref{eq:hessian_cond} with any
$0 \leq \gamma_1 < 2$, and $0 \leq \gamma_2< 2$.
This completes of the proof of Proposition
\ref{pro:function_examples}.

\vskip .1in
\noindent
\textbf{Proof of Proposition \ref{pro:logistic_loss}. }
We first prove two claims to be used later in the proof.
\begin{claim}
\label{claim_1:multi}
Suppose that function $f: \mathbb R\mapsto \mathbb R$
satisfies Assumption 1 with $\gamma_1=1$, $\gamma_2=0$ and $\beta$.
Then for any $p$-dimensional vector $a$
and any scalar $b$,
the function $g(x):=f(a^\top x+b)$ also
 satisfies Assumption 1 with $\gamma_1=1$,
 $\gamma_2=0$, and $\beta\|a\|_2$.
\end{claim}

\noindent
\textbf{Proof of Claim  \ref{claim_1:multi}. }
 Suppose that
Assumption (A1) holds for $f(\cdot)$
with $\gamma_1=1$, $\gamma_2=0$, and $\beta$.
Then, for any $d \in \mathbb R^{p}$ with
$\|d\|_2\leq (\beta\|a\|_2)^{-1}$, we have that
$|a^\top d|^2 \leq \|a\|_2^2\|d\|_2^2\leq \beta^{-2}$.
By Assumption (A1), we have, for any $s\in \mathbb R$, that
\begin{equation*}
\left|f'(s+a^\top d)-f'(s)-f''(s)  a^\top d\right|\leq
\beta f''(s)  (a^\top d)^2 \, .
\end{equation*}
Moreover, note that
$\nabla g(x)=f'(a^\top x+b) a$ and
$\nabla^2 g(x)= f''(a^\top x+b) a a^\top$.
Consequently, for any $x\in \mathbb R^p$, we have that
\begin{align*}
& \|\nabla g(x+d)-\nabla g(x)-\nabla^2 g(x)d \|_2 \\
&=\|(f'(a^\top x+b+a^\top d)-f'(a^\top x+b)-f''(a^\top x+b) a^\top d ) a\|_2\\
&=\left|f'(a^\top x+b+a^\top d)-f'(a^\top x+b)-f''(a^\top x+b) a^\top d  \right|\cdot \|a\|_2\\
&\leq \beta f''(a^\top x+b)  (a^\top d)^2 \|a\|_2
= \beta \|a\|_2  d^{\top} \nabla^2 g(x) d \, ,
\end{align*}
which implies that $g(\cdot)$
satisfies Assumption 1 with $\gamma_1=1$, $\gamma_2=0$, and
$\beta \|a\|_2$.

\begin{claim}
\label{claim_2:multi}
Suppose that functions $f_i$ with
 $f_i: \mathbb R^p\mapsto \mathbb R$ satisfies
 Assumption (A1) with $\gamma_1=1$, $\gamma_2=0$, and
 $\beta = \beta_i$; $i = 1,\ldots, n$.
 Then $\alpha_1 f_1+\alpha_2 f_2 \ldots +\alpha_n f_n$ also
 satisfies Assumption (A1)
 with $\gamma_1=1$, $\gamma_2=0$, and $\beta =
 \max_{1 \leq i \leq n} \beta_i$, where
 $\alpha_1, \ldots ,\alpha_n \geq 0$.
\end{claim}

\noindent
\textbf{Proof of Claim \ref{claim_2:multi}. }
For any $d$ satisfying $\|d\| \leq 1/\beta$, we have $\|d\| \leq 1/\beta_i$  for all $i=1, \ldots, n$.
By Assumption (A1), this implies
\begin{equation}
\left \|\nabla f_i(s+d)-\nabla f_i(s)-\nabla^2 f_i(s)d
\right \|_2 \leq \beta_i
 d^{\top}\nabla^2 f_i(s)d
 \leq \beta d^{\top}\nabla^2 f_i(s)d,
\end{equation}
for any vector $s \in \mathbb R^p$ and $i=1, \ldots, n$.
Thus,
\begin{align*}
& \left \|
\sum_{i=1}^n \alpha_i \nabla f_i(s+d)- \sum_{i=1}^n
\alpha_i \nabla f_i(s)-\sum_{i=1}^n \alpha_i \nabla^2 f_i(s)d
\right \|_2  \\
&\leq
\sum_{i=1}^n \alpha_i \left \|\nabla f_i(s+d)-\nabla f_i(s)
-\nabla^2 f_i(s)d \right \|_2 \\
& \leq
 \sum_{i=1}^n \alpha_i \beta d^{\top} \nabla^2 f_i(s)d
 =\beta d^{\top} \left(\sum_{i=1}^n\alpha_i \nabla^2 f_i(s) \right) d \, .
\end{align*}
Consequently, the function
 $\sum_{i=1}^n \alpha_i f_i$ also satisfies Assumption (A1)
 with $\gamma_1=1$, $\gamma_2=0$, and $\beta = \max_{1 \leq i \leq n} \beta_i$.

Now we are ready to prove the main result.
 It has been shown in Proposition \ref{pro:function_examples}
  that the logistic regression loss function satisfies
Assumption (A1) with $\gamma_1=1$, $\gamma_2=0$
and $\beta=2$. By using Claim \ref{claim_1:multi}, it follows that
 $\log(1+e^{-Y_i X_i^\top \theta})$ also
satisfies Assumption (A1) with $\gamma_1=1$, $\gamma_2=0$ and
 $\beta=2 \max_{1\leq i \leq n} \|X_i\|_2$ for $i = 1,\ldots,n$.
Moreover, it follows from Claim \ref{claim_2:multi} that
the logistic regression empirical loss
$n^{-1}\sum_{i=1}^n\log(1+e^{-Y_i X_i^\top \theta})$
satisfies Assumption (A1) with $\gamma_1=1$, $\gamma_2=0$,
 and $\beta=2 \max_{1\leq i \leq n} \|X_i\|_2$.
 This completes the proof of Proposition \ref{pro:logistic_loss}.

\section{Approximation-error bounds for the ODE methods}
\label{sec:appendix_ode}
In this section,
we follow the classical global approximation error analysis of
ordinary differential equation, which studies the ODE
$\theta^\prime(t) = F(t,\theta(t))$.
In particular,
we focus on Euler's method
and second-order Runge-Kutta method,
which have been studied extensively in the numerical ODE literature
 \citep{hairer2008solving, butcher2016numerical}.
Both methods belong to the more general class of the so-called
one-step method \citep{hairer2008solving}, for which
Lipschitz continuity of function $F(t,\theta)$ plays an important
 role in quantifying the approximation error. Here we present a result which is a direct
 application of Theorem 3.4 of \citet{hairer2008solving}.

\begin{thm}
  (Theorem 3.4 of \cite{hairer2008solving})
\label{thm:one-step method}
Assume that $L_n(\theta)$ is $M$-Lipschitz continuous
and $m$-strongly convex.
Moreover, assume that the gradient of $L_n(\theta)$ is $L$-Lipschitz continuous and
the Hessian of $L_n(\theta)$ is $S$-Lipschitz continuous.
We have that
\begin{equation}
\label{eq:ode_error_bd}
\|\theta_k - \theta(t_k)\|_2 \leq \frac{C \alpha^p}{L^{\star}}(e^{L^{\star} t_k}-1) \, ,
\end{equation}
where $\alpha$ is the step size, $t_k = k \alpha$, $C$ is an absolute constant,
and $L^{\star} = \frac{MS}{\min\{m, 1\}^2}+\frac{L}{\min\{m, 1\}}$.
\end{thm}

Note that the approximation error is a power function of the
step size $\alpha$. The power $p$ is often referred to as the order
the corresponding approximation method.
As we can see from the above Theorem, Euler's method defined
in \eqref{eq:euler_approx_iterates} and the special
case of Runge-Kutta method defined in \eqref{eq:Runge-Kutta}
are first-order
method and second-order method, respectively.
In both cases, we can control
global error in finite interval by adjusting step size $\alpha$.

We also point out that the upper bound in \eqref{eq:ode_error_bd} gets
 worse as $k \rightarrow \infty$, which is
less desirable compared with the approximation error
bounds derived for the other two path-following methods.
This is likely due to the generality of problem class considered in
Theorem 3.4 of \cite{hairer2008solving}.
Indeed, some preliminary empirical studies suggest that the second-order
Runge-Kutta is practically comparable to the Newton method in terms of
approximation error.
A more refined theoretical upper bound may hold for the particular
ODE we consider here, although we choose not to pursue this due to space limit.

\vspace{.3cm}
\noindent
\textbf{Proof of Theorem \ref{thm:one-step method}. }
Applying Theorem 3.4 of \cite{hairer2008solving}, it suffices to
show that $F(t, \theta(t))$ is Lipschitz continuous
with respect to $\theta(t)$, based on which we could bound
the global error directly. For any $t > 0$,
 $\theta_1$ and $\theta_2$,
 let $\Delta F=F(t,\theta_1)-F(t,\theta_2)$. Note that
\begin{eqnarray*}
\|\Delta F\|_2 &=& \left \|
[(1-e^{-t})\nabla^2 L_n(\theta_1)+e^{-t} I]^{-1}\nabla L_n(\theta_1)
-
[(1-e^{-t})\nabla^2 L_n(\theta_2)+e^{-t} I]^{-1}\nabla L_n(\theta_2)
 \right\|_2\\
&\leq& \underbrace{\left \|\left[((1-e^{-t})\nabla^2 L_n(\theta_1)
+e^{-t} I)^{-1}-((1-e^{-t})\nabla^2 L_n(\theta_2)+e^{-t} I)^{-1}
\right] \nabla L_n(\theta_2) \right\|_2}_\text{Part I}+\\
&&
\underbrace{\left \|((1-e^{-t})\nabla^2 L_n(\theta_1)+e^{-t} I)^{-1}
(\nabla L_n(\theta_1)-\nabla L_n(\theta_2))\right\|_2}_\text{Part II} \, .
\end{eqnarray*}
Since $L_n(\theta)$ is $m$-strongly convex and $\nabla L_n(\theta)$
is $L$-Lipschitz continuous, we have that for any $\theta$, $\nabla^2 f_t(\theta)=(1-e^{-t})
  \nabla^2 L_n(\theta)+e^{-t} I$ satisfies that
\begin{equation*}
[(1-e^{-t})\cdot L +e^{-t}]^{-1} I  \preceq [\nabla^2 f_t(\theta)]^{-1}
\preceq [(1-e^{-t}) m +e^{-t}]^{-1} I \, .
\end{equation*}
Therefore,
\begin{eqnarray*}
\|(\nabla^2 f_t(\theta_1))^{-1}-(\nabla^2 f_t(\theta_2))^{-1}\|_2
&=&
\|(\nabla^2 f_t(\theta_1))^{-1} (\nabla^2 f_t(\theta_2)-\nabla^2 f_t(\theta_1))
(\nabla^2 f_t(\theta_2))^{-1}\|_2\\
&\leq&\|(\nabla^2 f_t(\theta_1))^{-1}\|_2 \|\nabla^2 f_t(\theta_2)-\nabla^2
 f_t(\theta_1)\|_2 \|(\nabla^2 f_t(\theta_2))^{-1}\|_2\\
&\leq& [(1-e^{-t}) m +e^{-t}]^{-2} (1-e^{-t})
 \|\nabla^2 L_n(\theta_1)-\nabla^2 L_n(\theta_1)\|_2 \\
&\leq& [(1-e^{-t}) m +e^{-t}]^{-2} (1-e^{-t})
S\|\theta_1-\theta_2\|_2 \, .
\end{eqnarray*}
Moreover, since $L_n(\theta)$ is $M$-Lipschitz continuous and convex, we have that
\begin{equation}
\|\nabla L_n(\theta)\|_2^2 \leq
|L_n(\theta+\nabla L_n(\theta))-L_n(\theta)|\leq M\|\nabla L_n(\theta)\|_2 \, ,
\end{equation}
which implies that  $\|\nabla L_n(\theta)\|_2\leq M$.
Consequently, we can bound Part I as follows
\begin{eqnarray*}
\text{Part I} &\leq& \|[\nabla^2 f_t(\theta_1)]^{-1}
-[\nabla^2 f_t(\theta_2)]^{-1}\|_2\|\nabla L_n(\theta_2)\|_2\\
&\leq& M\|[\nabla^2 f_t(\theta_1)]^{-1}-[\nabla^2 f_t(\theta_2)]^{-1}\|_2\\
&\leq&
M [(1-e^{-t}) m +e^{-t}]^{-2} (1-e^{-t}) S\|\theta_1-\theta_2\|_2 \, .
\end{eqnarray*}
For part II, we have that
\begin{eqnarray*}
\text{Part II}&\leq&
\|[(1-e^{-t})\nabla^2 L_n(\theta_1)+e^{-t} I]^{-1}\|_2\|
\nabla L_n(\theta_1)-\nabla L_n(\theta_2)\|_2\\
&\leq& [(1-e^{-t}) m +e^{-t}]^{-1} L \|\theta_1-\theta_2\|_2 \, .
\end{eqnarray*}
Combining the above two bounds, it follows that
\begin{equation*}
\|\Delta F\|_2 \leq \left(M [(1-e^{-t}) m +e^{-t}]^{-2} (1-e^{-t}) S
+[(1-e^{-t}) m +e^{-t}]^{-1} L\right) \|\theta_1-\theta_2\|_2 \, .
\end{equation*}
Let $S^*=M [(1-e^{-t}) m +e^{-t}]^{-2} (1-e^{-t}) S
 +[(1-e^{-t}) m +e^{-t}]^{-1} L$. It can be shown that
\begin{eqnarray*}
S^*\leq L^\star := \frac{MS}{\min\{m, 1\}^2}+\frac{L}{\min\{m, 1\}} \, .
\end{eqnarray*}
Hence, we have that
$\|\Delta F\|_2 \leq L^\star \|\theta_1-\theta_2\|_2$ for any $m>0$,
which implies that
 $F(t,\theta)$ is $L^\star$-Lipschitz continuous
  with respect to $\theta$. This completes the proof of
Theorem \ref{thm:one-step method}.

\section{Additional experiments}
\label{sec:appendix_add_exp}
In this section, we provide some additional simulation results for ridge regression.
In particular, we compare the proposed methods based on Newton and gradient descent updates
against glmnet in terms of both runtime and approximation error,
under the setting of ridge regression.
In our simulation, the data $\{(X_i, Y_i)\}_{i=1}^n$ are generated from
the usual linear regression model
$Y_i = X_i^\top \theta^\star + \tilde \epsilon$, where
$\tilde \epsilon \sim N(0, \sigma^2)$,
$\theta^\star = (1/\sqrt{p},\ldots, 1/\sqrt{p})^\top$,
and $X_1, \ldots, X_n$ are IID samples from $N_p(0, I_{p\times p})$.
We consider two different scenarios with
$\sigma^2=1/4$ and $\sigma^2=4$.
Moreover, for each scenario, we consider three different problem dimensions:
$(n,p)=(1000, 500)$, $(n,p)=(1000,1000)$, and $(n,p)=(1000, 2000)$.

Again, we use the global approximation error
$\sup_{0 \leq t \leq t_{\max}} \{ f_t(\tilde \theta(t)) - f_t(\theta(t)) \}$
to assess the accuracy for the approximate solution path
$\tilde \theta(t)$,
where $\tilde \theta(t)$ is the linear interpolation
of the iterates $\theta_k$ generated by each method.
Moreover,
we sample $N$ points $s_1, \ldots, s_N$
uniformly from $(0, t_{\max})$ and use
$\max_{1\leq i\leq N}\{ f_{s_i}(\tilde \theta({s_i})) - f_{s_i}(\theta({s_i})) \}$
as an approximation of
$\sup_{0 \leq t \leq t_{\max}} \{ f_t(\tilde \theta(t)) - f_t(\theta(t)) \}$.
Here $\theta({s_i})$ is the exact solution at $s_i$ and can be computed explicitly.
In our simulations, we use $N=100$ and $t_{\max}=10$.

\begin{figure}[ht!]
\begin{subfigure}{0.5\textwidth}
\includegraphics[width=\linewidth]{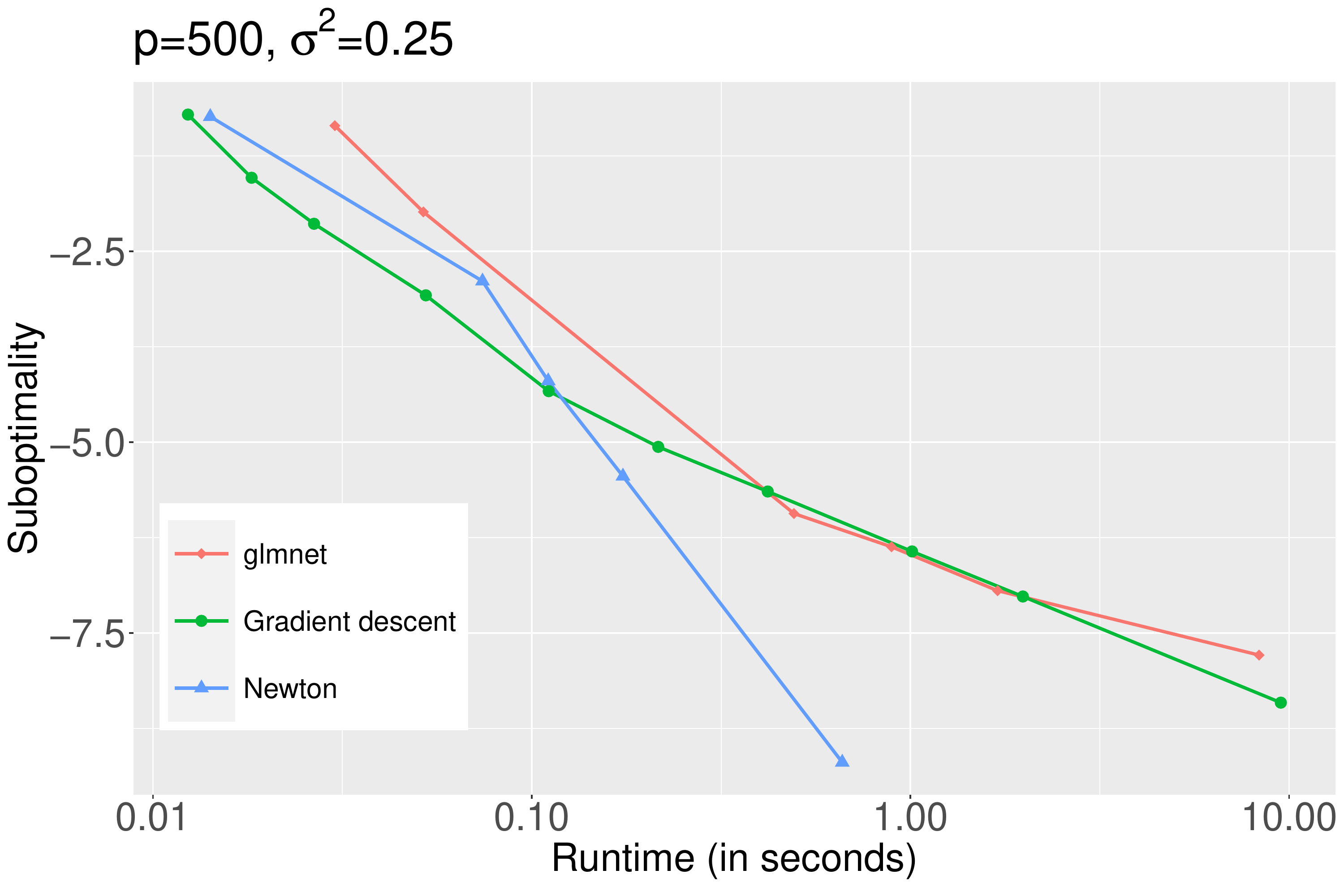}
\end{subfigure}
\hspace*{\fill}
\begin{subfigure}{0.5\textwidth}
\includegraphics[width=\linewidth]{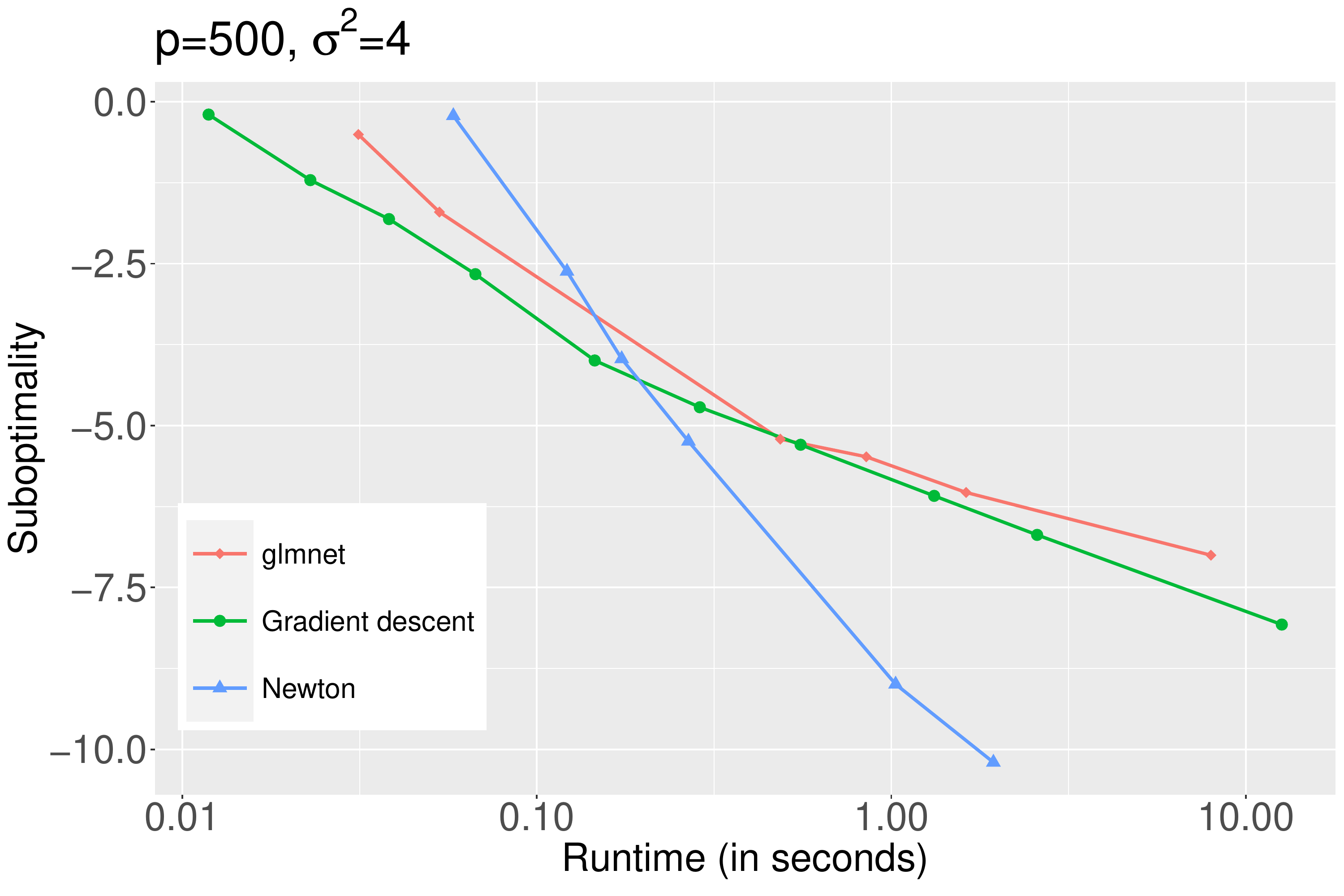}
\end{subfigure}

\medskip
\begin{subfigure}{0.5\textwidth}
\includegraphics[width=\linewidth]{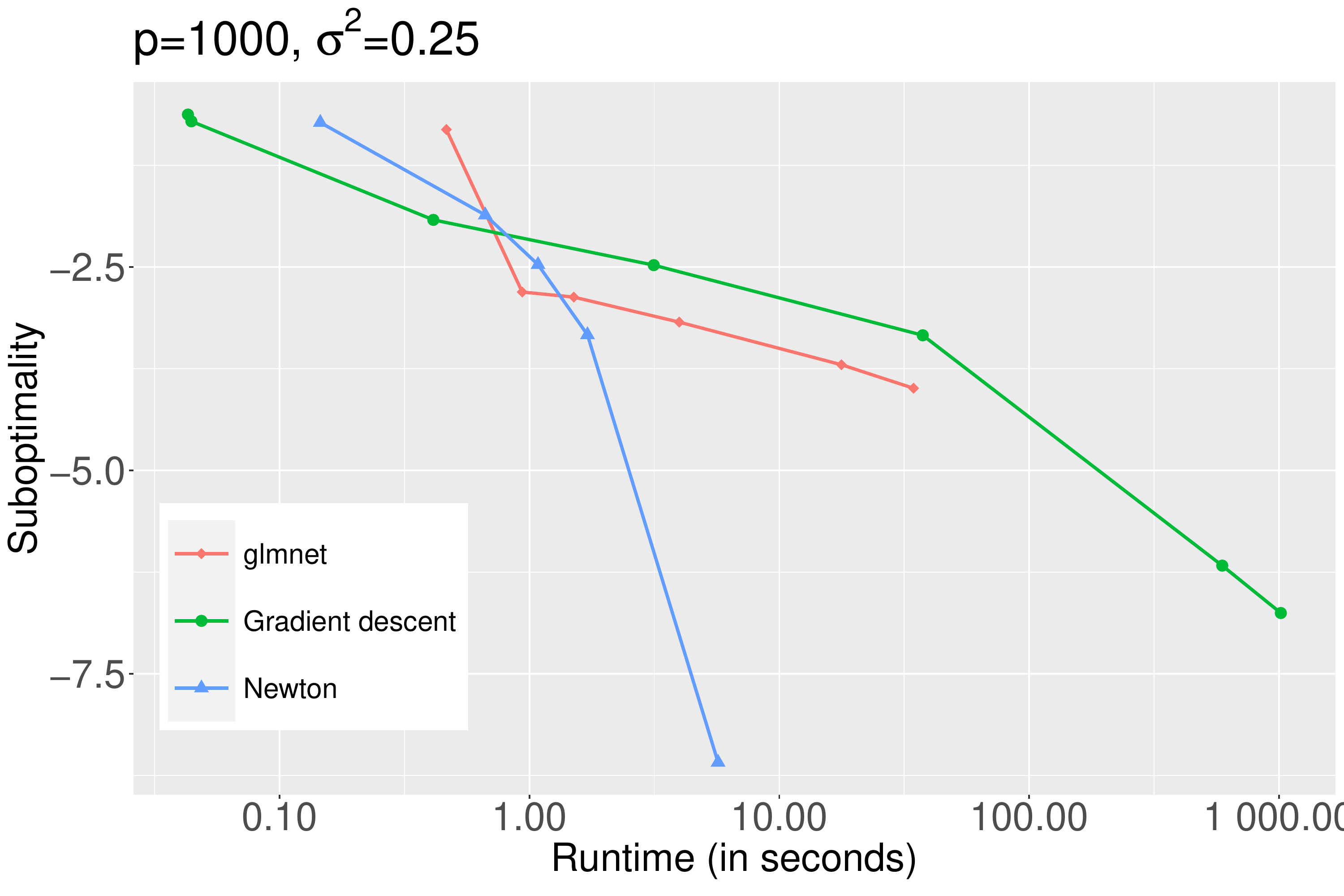}
\end{subfigure}
\hspace*{\fill}
\begin{subfigure}{0.5\textwidth}
\includegraphics[width=\linewidth]{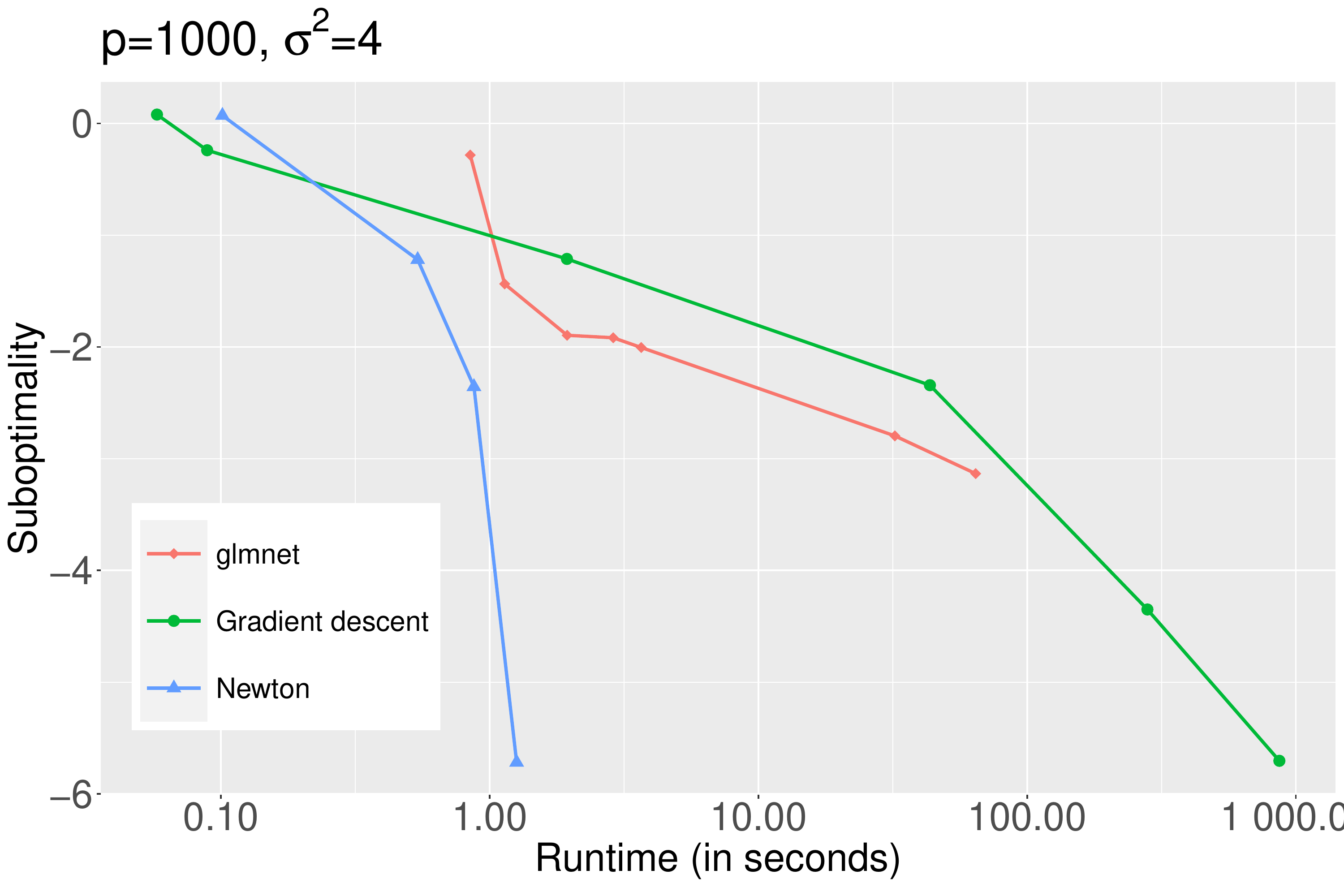}
\end{subfigure}

\medskip
\begin{subfigure}{0.5\textwidth}
\includegraphics[width=\linewidth]{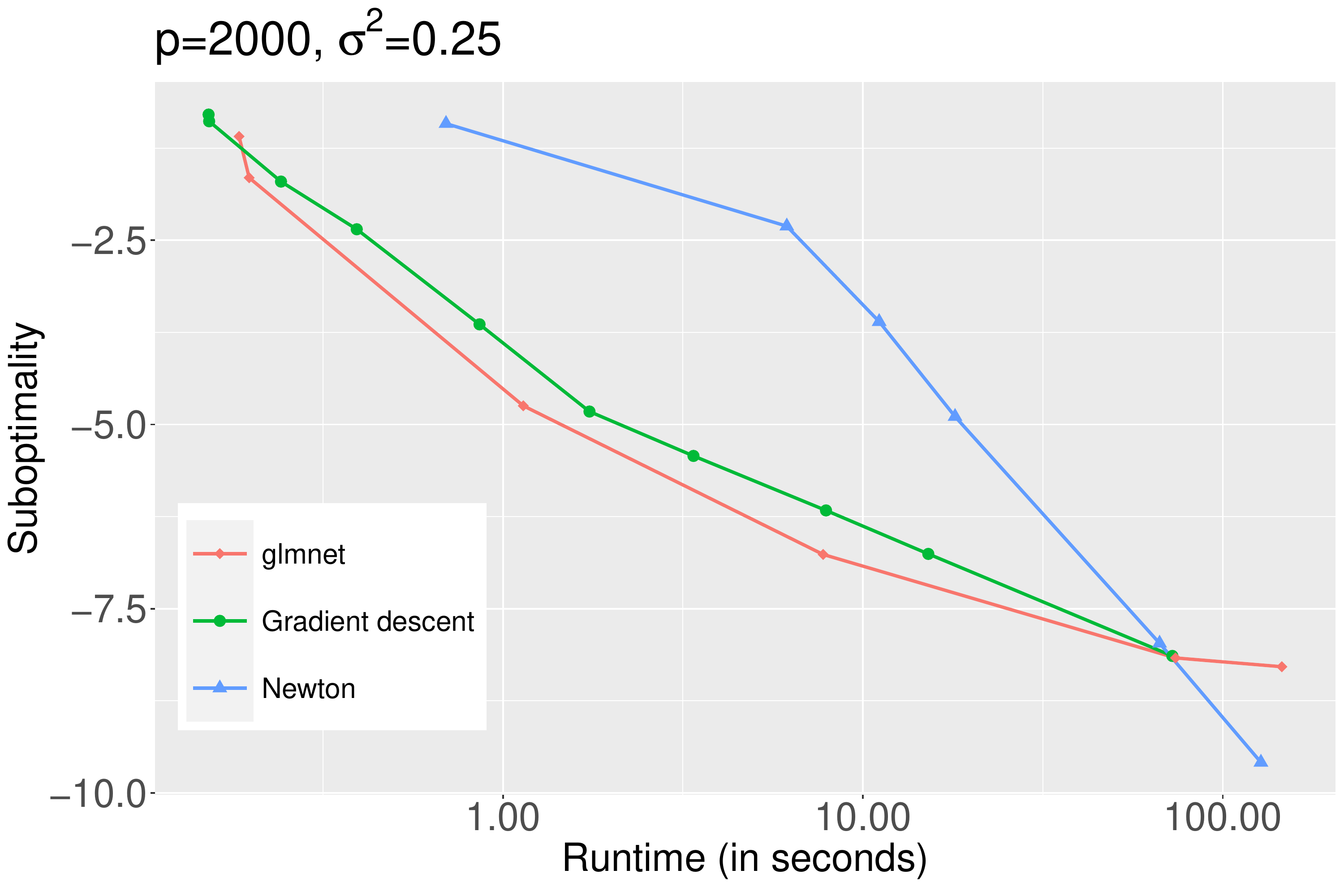}
\end{subfigure}
\hspace*{\fill}
\begin{subfigure}{0.5\textwidth}
\includegraphics[width=\linewidth]{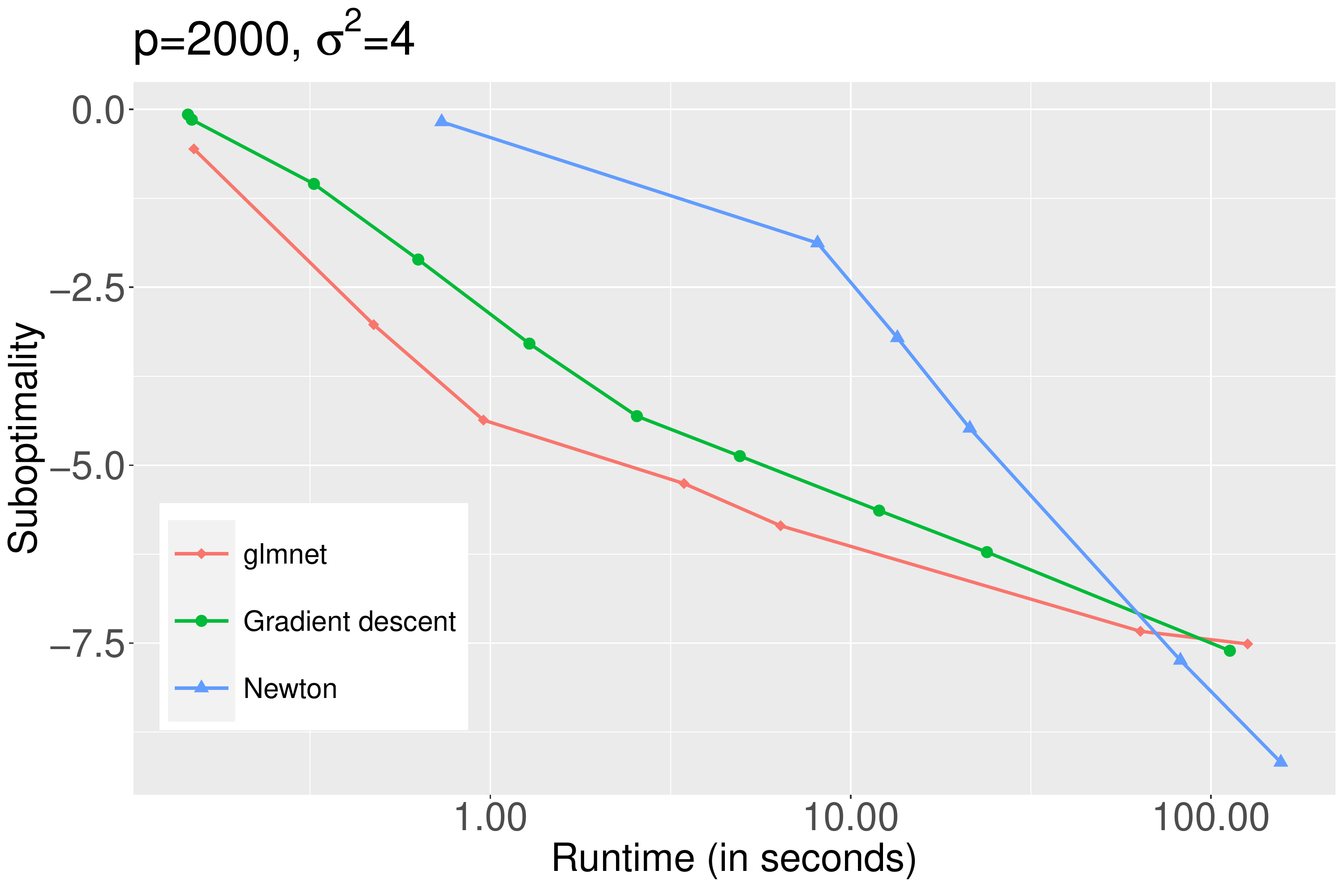}
\end{subfigure}
\caption{Runtime v.s. suboptimality
for the proposed Newton method, gradient descent method, and glmnet under six different scenarios,
when applied to ridge regression.
}
\label{figure:time_ridge}
\end{figure}

Figure \ref{figure:time_ridge} plots runtime versus
approximation error based on $100$ simulations.
Similar to Figure \ref{figure:time_nonsep},
we can see from Figure \ref{figure:time_ridge} that
in all scenarios
the proposed Newton method runs the fastest when
the required accuracy is high (small suboptimality).
Moreover,
glmnet is no better than Newton method for smaller problems ($p=500$ and $1000$);
while glmnet outperforms both Newton method and the gradient method when low accuracy solution is
sufficient and problem dimension is large ($p = 2000$).
Lastly, in all cases the gradient method runs faster than Newton method when the desired accuracy is low.

\newpage

\newpage
\bibliography{bibliography}

\end{document}